\pdfoutput=1

\documentclass[11pt]{article}

\usepackage[final]{acl}

\usepackage{times}
\usepackage{latexsym}

\usepackage[T1]{fontenc}

\usepackage[utf8]{inputenc}

\usepackage{microtype}

\usepackage{inconsolata}

\usepackage{graphicx}

\usepackage{enumitem}
\usepackage{hyperref}
\usepackage{longtable}
\usepackage{caption}
\usepackage{url}
\usepackage{todonotes} 
\usepackage{subcaption}
\usepackage{amssymb}
\usepackage{enumitem}
\usepackage{array}
\usepackage{longtable}
\usepackage{makecell}
\usepackage{xspace}
\usepackage{xcolor,colortbl}
\usepackage{tcolorbox}
\usepackage{arydshln}
\usepackage{adjustbox}

\usepackage{pgfplots}
\pgfplotsset{width=1.0\columnwidth}
\usepackage{multirow}
\usepackage{makecell}
\usepackage{pifont}
\usepackage{bbm}
\usepackage{rotating}
\usepackage{amsmath}
\usepackage{tablefootnote}
\usepackage{soul}
\usepackage{booktabs}
\usepackage{lipsum}
\usepackage{tikz}
\usepackage[tikz]{bclogo}
\usepackage{pgfplotstable}
\usepackage{mdframed}
\usepackage{graphicx}
\usepackage{amsmath}
\usepackage{algorithm}
\usepackage{algpseudocode}
\usepackage{wrapfig}
\usepackage[fixed]{fontawesome5}
\newcommand*{\affmark}[1][*]{\textsuperscript{#1}}

\DeclareRobustCommand{\hlpink}[1]{{\sethlcolor{pink}\hl{#1}}}

\definecolor{beaublue}{rgb}{0.74, 0.83, 0.9}

\title{Guiding Through Complexity: What Makes Good Supervision for Hard Math Reasoning Tasks?}

\author{Xuan He\thanks{\ \  These two authors contributed equally. The order of authorship decided by the flip of a coin.}\affmark[$\clubsuit$] 
\quad Da Yin\affmark[$*\spadesuit$] 
\quad Nanyun Peng$^\spadesuit$ \\
$^\clubsuit$ Tsinghua University \\
$^\spadesuit$ University of California, Los Angeles  \\
  {\tt hexuan21@mails.tsinghua.edu.cn, \{da.yin,violetpeng\}@cs.ucla.edu} \\
}

\begin{document}
\maketitle
\begin{abstract}

How can ``weak teacher models''~\cite{bowman2022measuring} such as average human annotators or existing AI systems, effectively supervise LLMs to improve performance on hard reasoning tasks, especially those that challenge and requires expertise or daily practice from the teacher models? In this paper, we seek for empirical answers to this question by investigating various data-driven strategies that offer supervision data at different quality levels upon tasks of varying complexity. 
Two intuitive strategies emerge for teacher models to provide supervision during alignment training: 1) using lower-quality supervision from hard full tasks that match the difficulty of the target reasoning tasks, and 2) leveraging higher-quality supervision from easier subtasks that are less challenging. Interestingly, we find that even when the outcome error rate for hard task supervision is high (e.g., 90\%), training on such data can outperform perfectly correct supervision of easier subtasks on multiple hard math benchmarks.
We further identify a more critical factor influencing training performance: step-wise error rates, which indicate the severity of errors in solutions. Specifically, training on hard task supervision with the same outcome error rates but disparate step-wise error rates can lead to a 30\% accuracy gap on MATH benchmark. Our results also reveal that supplementing hard task supervision with the corresponding subtask supervision can yield notable performance improvements than simply combining rephrased hard full task supervision, suggesting new avenues for data augmentation. Data and code will be released at \url{https://github.com/hexuan21/Weak-to-Strong}.

\end{abstract}

\begin{figure*}[h]
    \centering
    \vspace{-6pt}
    \includegraphics[width=0.83\textwidth]
    {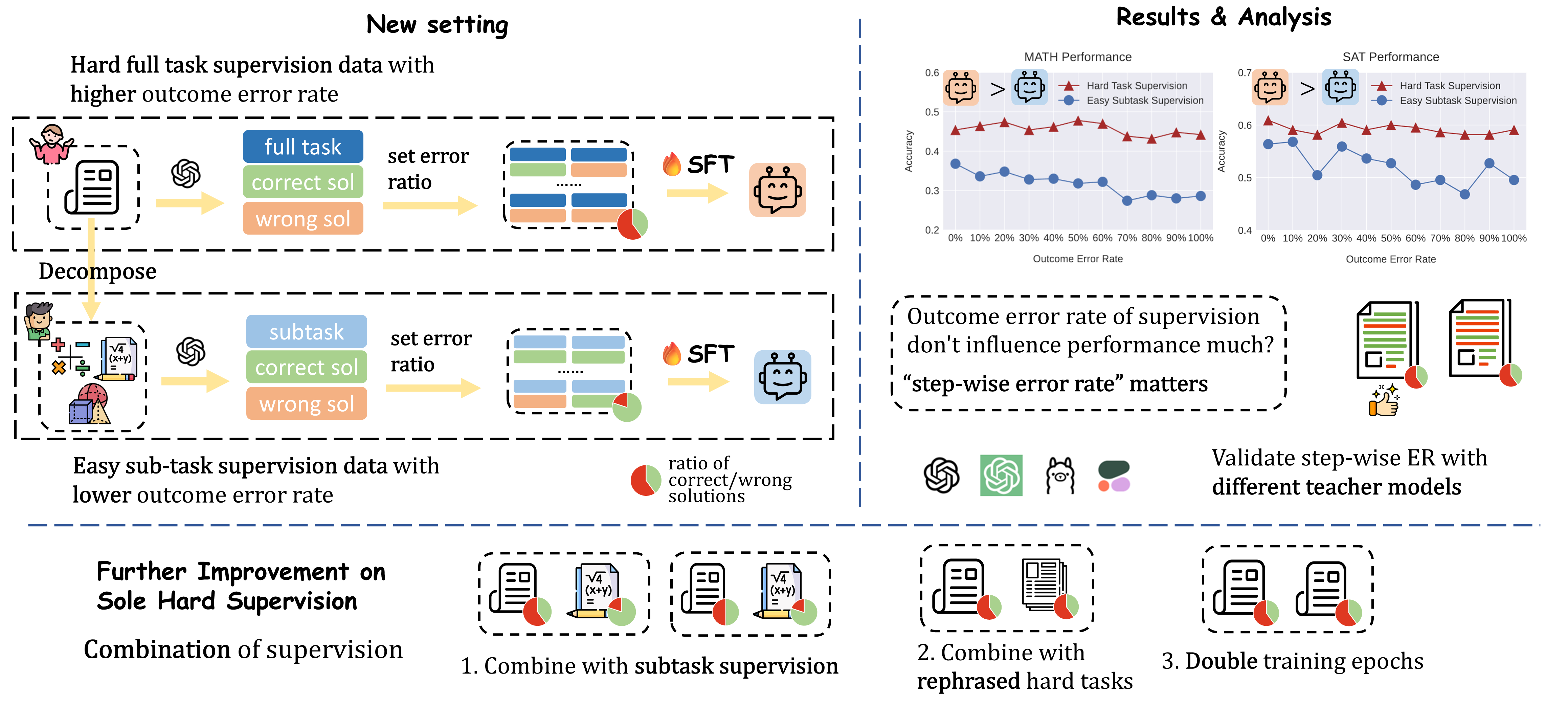}
    \caption{Overview of our empirical study on two contrasting supervision strategies and further analysis.} 
    \label{fig:teaser}
    \vspace{-9pt}
\end{figure*}

\section{Introduction}
\label{sec:1}


The rapid advancements in Large Language Models (LLMs) \citep{gpt4o_paper,claude_blog,reid2024gemini,dubey2024llama} indicate that AI systems will soon be capable of progressively solving hard reasoning tasks that challenge both humans and current LLMs. To enhance the performance of today's AI systems \citep{bowman2022measuring,burns2023weak} on such hard reasoning tasks, especially those that exceed the reasoning abilities of the average human or current LLMs, what strategies should we adopt to consistently and effectively supervise the AI systems?

\textbf{We are concerned what supervision data those ``weak teacher models'', humans and current LLMs, should annotate or synthesize for better hard downstream task performance.} We mainly consider two possible data-driven supervision strategies -- \textbf{Strategy 1}: Sampling or annotating solutions directly on the hard tasks that both human annotators and AI models typically struggle with; \textbf{Strategy 2:} Sampling or annotating solutions on hard tasks' corresponding subtasks, which are more manageable and where humans or AI models are more likely to succeed \citep{gulccehre2016knowledge,glasmachers2017limits,wiessub,jothimurugan2023robust}. For Strategy 1, while the tasks used for supervision match the reasoning difficulty of our studied hard downstream tasks, the quality of supervision may be compromised due to the difficulty that challenges the human and AI teacher models. In contrast, Strategy 2 involves easier tasks that do not match the difficulty level of the downstream tasks, but the supervision quality tends to be higher because these tasks are simpler for annotators or models to handle. Should we obtain low-quality supervision for hard tasks, or employ high-quality supervision for decomposed subtasks for fine-tuning?

To \emph{fairly} compare the two supervision strategies, which involve task difficulty and supervision quality, we introduce a simulated setting: 
The supervision quality is controlled by introducing varying outcome error rates, which represent the percentage of noisy supervision where solutions with wrong final answers replace the ground-truth solutions. It is designed for simulating the varying supervision quality from ``weak'' teacher models with different capacity.
After acquiring full task and subtask supervision with varying outcome error rates, and following \citet{bansal2024smallerweakerbettertraining}, we apply supervised fine-tuning (SFT) using the acquired supervision to LLMs for solving hard reasoning tasks. 

Across 5 hard math reasoning tasks, including college entrance exams, high-school competitions and college-level math tests, we find hard task supervision consistently outperforms subtask supervision, regardless of outcome error rates. Specifically, we observe that: 1) even if outcome error rate of hard supervision is 90\%, it can still outperform perfectly correct subtask supervision; 2) for easy subtask and hard full task supervision, as the outcome error rate increases from 0\% to 100\%, the performance does not decrease monotonically.

\emph{Rather than the outcome error rate, what factors highly impact the hard test task performance?} 
One key factor we identify is the \emph{step-wise error rate}, i.e., average ratio of erroneous steps (\S\ref{sec:6}) to the total steps in the supervision set. 
Specifically, we observe that when different teacher models exhibit similar outcome error rates, an increase in step-wise error rate leads to lower performance on hard reasoning tasks. We test a batch of LLMs, including GPT-4o-mini \citep{gpt-4o-mini_blog}, GPT-3.5-turbo \citep{chatgpt_paper}, Llama-3-70B-Instruct, and Command-R~\citep{command-r_blog}, as teacher models for providing hard task supervision. We then compare the performance of the rest models' supervision against that of GPT-4o-mini at a similar outcome error rate. Notably, with a lower step-wise error rate, GPT-4o-mini's sampling supervision always leads to higher results. When both GPT-4o-mini and Command-R have around 90\% outcome error rate in their generated supervision, Command-R underperforms by 12\% accuracy on MATH~\citep{hendrycks2measuring} and is even worse than training with subtask supervision. 

\emph{Can we further improve the hard downstream task performance without relying on more hard task annotations?} 
We find that supplementing hard task supervision with subtask supervision, i.e., the supervision of hard tasks' corresponding subtasks, can lead to more performance gains. 
Specifically, we combine the hard task supervision with an outcome error rate of $A$ and the subtask supervision with an outcome error rate of $B$ 
with three distinct combinations of $A$ and $B$. We observe that adding subtask supervision improves performance across most of the 5 downstream tasks. When $A=50\%$ and $B=10\%$, we achieve the best results on benchmarks like MATH and SAT-Math. We also find that this simple supervision combination method is more effective than simply doubling the training epochs on the mere hard task supervision or mixing rephrased hard task supervision.

Overall, we summarize our main contributions for enhancing LLM's hard reasoning task performance as follows:
\begin{itemize}[leftmargin=*]
    \setlength{\itemsep}{3pt}
    \item We offer new insights into how complexity and quality of weak supervision collectively impact hard task performance, distinct from traditional weak supervision and weak-to-strong generalization literature.
    \item We conduct one of the most comprehensive evaluations across a range of hard reasoning tasks in weak-to-strong generalization field, surpassing in both scope and task difficulty. We also introduce a fairer comparison by ensuring that the supervision from easy subtasks and hard full tasks is kept within a similar range.
    \item We provide comprehensive and novel insights which contribute data-driven guidance for researchers and practitioners aiming to improve LLMs in future reasoning-heavy applications.
\end{itemize}

\section{Formulation of Supervision Strategy Analysis}
\label{sec:3}
In this section, we present the framework for our analysis of supervision strategies aimed at enhancing performance on hard downstream reasoning tasks. We outline the motivation behind two supervision strategies involving both easy and hard tasks, introduce the key notations, and discuss the design principles of the analysis formulation for fair and comprehensive evaluation.

\subsection{Overview of Supervision Strategies} 
\label{subsec:3.1}
Future AI systems need to solve hard reasoning tasks that ``weak teacher models'' (e.g., average humans or current AI systems) currently struggle with. To achieve better performance on such hard tasks, what types of training supervision should these teacher offer for training LLMs? We propose the two following supervision strategies, each with distinct characteristics that could contribute to this improvement:


\vspace{3pt}
\noindent \textbf{Strategy 1}: Use teacher models to sample or annotate solutions for \textbf{hard full tasks} that human annotators or AI models typically struggle with. Although this supervision aligns with the reasoning difficulty of the target downstream tasks, the supervision may be \textbf{lower-quality} due to the task complexity for obtaining correct final answer.
\vspace{3pt}

\noindent \textbf{Strategy 2}: Use teacher models to sample or annotate solutions for \textbf{easy subtasks}. These tasks, while not matching the complexity of the downstream tasks, offer \textbf{higher-quality} supervision because they are simpler to handle.
\vspace{3pt}



We denote the two types of supervision as $\mathcal{D}_{\mathrm{Hard}}$ and $\mathcal{D}_{\mathrm{Subtask}}$. We also define supervision quality of $\mathcal{D}_{\mathrm{Hard}}$ and $\mathcal{D}_{\mathrm{Subtask}}$ in terms of outcome error rates, i.e., the proportion of incorrect solutions with wrong final answers relative to the total supervision instances. Higher outcome error rates indicate lower supervision quality.
Following \citet{bansal2024smallerweakerbettertraining}, we employ supervised fine-tuning (SFT) with the supervision from the above strategies to train LLMs for solving hard tasks and determine the more effective strategy.


In summary, we formally examine \textbf{whether hard task supervision $\mathcal{D}_{\mathrm{Hard}}$ with outcome error rate $\epsilon_{\mathrm{Hard}}$ yields better performance on hard reasoning tasks compared to the easy subtask supervision $\mathcal{D}_{\mathrm{Subtask}}$ with error rate $\epsilon_{\mathrm{Subtask}}$.}


\subsection{Design Principles of Analysis Formulation}
\label{subsec:3.3}

To ensure the fairness and comprehensiveness of our analysis, we address the following aspects:

\noindent \textbf{Comparison Fairness.} \emph{We ensure that subtasks must be less difficult than hard full tasks and avoid introducing any extra tasks that go beyond the scope of hard task supervision and might help solve the test tasks} (\S\ref{subsec:4.2}). 
It was neglected in prior related work which involves preparing easy and hard task supervision~\citep{sun2024easy,hase2024unreasonableeffectivenesseasytraining,bansal2024smallerweakerbettertraining}.

\noindent \textbf{Evaluation with Varying Supervision Quality.} \textit{Even if some teacher models like human annotators are weak, they still have varying capabilities.} For instance, some may fail 60\% of hard full tasks, while others might fail up to 90\%. To simulate varying supervision quality from those teacher models with different capacity, we first collect a substantial set of hard full tasks and subtasks, each containing both ground-truth and incorrect solutions (\S\ref{subsec:4.3}). To create supervision with $A\%$ outcome error rate, we can simply replace $A\%$ of the original ground-truth solutions with incorrect ones.


\section{Hard Full Task and Easy Subtask Supervision Synthesis}
\label{sec:4}

In this section, we detail the process of collecting easy subtask and hard full task supervision, $\mathcal{D}_{\mathrm{Subtask}}$ and $\mathcal{D}_{\mathrm{Hard}}$, based on the design principles of our supervision strategy analysis framework.

\subsection{Overview}
\label{subsec:4.1}
As outlined in \S\ref{subsec:3.3}, we need to decompose hard tasks into subtasks, ensuring that each decomposed subtask remains within the scope of the original hard task's solutions. Additionally, we introduce an outcome error rate metric as an indicator of supervision quality, allowing us to simulate varying levels of quality based on the capabilities of different teacher models. This enables us to analyze performance trends in relation to changes in supervision quality and facilitates direct comparisons between hard full task and subtask supervision across different quality levels. To this end, our supervision synthesis pipeline comprises two main stages: \textbf{Stage 1: Hard Task Decomposition} and \textbf{Stage 2: Outcome Error Rate Adjustment}. 

Our synthesis pipeline utilizes large-scale high-quality math training annotations (See \S\ref{subsec:4.4}) as the foundations. These resources provide many hard reasoning tasks with chain-of-thought solutions, which are fundamental for task decomposition, and supervision quality adjustment and validation.

\subsection{Stage 1: Hard Task Decomposition}
\label{subsec:4.2}
Concretely, we leverage the hard full tasks and their solutions from the large-scale high-quality math resource, such as NuminaMath-CoT, to construct $\mathcal{D}_{\mathrm{Hard}}$ (as detailed \S\ref{subsec:4.4}). Then, with the hard task decomposition prompt shown in Appendix \ref{sec:prompt_templates}, we employ GPT-4o-mini to break down each hard task and its complete solution into 2-3 subtasks along with their corresponding solutions derived directly from the hard full task solutions. 

To maintain the quality of subtasks, we filter out unqualified tasks with ill-defined problems and overly-simple solutions (Appendix \ref{subsec:filtering_eg}). After filtering, the product of Stage 1 is a collection of hard full tasks and their subtasks, each paired with a ground-truth solution and final answer. They serve as the foundation for generating supervision $\mathcal{D}_{\mathrm{Hard}}$ and $\mathcal{D}_{\mathrm{Subtask}}$ with varying outcome error rates and the filtering process also bridges the distribution gap between the text style of hard full and easy subtask instructions (we explore further about the potential discrepancies between $\mathcal{D}_{\mathrm{Hard}}$ and $\mathcal{D}_{\mathrm{Subtask}}$ in Appendix \ref{sec:gap_between_full_sub}). In a small sample set, the solutions to hard tasks have an average of $13.01$ steps while for subtasks, it's $5.96$.

\subsection{Stage 2: Outcome Error Rate Adjustment}
\label{subsec:4.3}

To introduce supervision quality change and conduct comparison at different quality levels, we swap a percentage of ground-truth solutions with the incorrect ones to adjust supervision quality. For example, to create hard task supervision with a 20\% outcome error rate, 20\% of ground-truth solutions are replaced with incorrect ones.

However, Stage 1 only offers ground-truth solutions and this process requires incorrect task solutions. How can we obtain the incorrect solutions for error-introducing replacement? Next we outline the process to ensure each task has both ground-truth and incorrect solutions.

\vspace{3pt}

\noindent \textbf{Sampling Multiple Solutions.}
We first use LLMs to perform multi-sampling, generating \textbf{three solutions} for each hard full task and subtask. To increase the likelihood of obtaining incorrect solutions for the tasks with different complexities, we leverage LLMs with different reasoning capabilities. Concretely, we employ GPT-4o for hard tasks and GPT-4o-mini for subtasks. After generating the three candidate solutions, we assess their correctness by comparing each one against the ground-truth final answers.

\vspace{3pt}

\noindent \textbf{Gathering and Filtering Tasks with Ground-Truth and Incorrect Solutions.}
With three sampled solutions and the ground-truth, we attempt to formulate solution pairs for each task, with one correct and one incorrect. Details and examples are shown in Appendix \ref{subsec:sampling_eg}. 
It may occur that all decomposed subtasks corresponding are filtered out, or vice versa, where a subtask’s corresponding hard full task is filtered out. To ensure that the hard full tasks in $\mathcal{D}_{\mathrm{Hard}}$ encompass the scope of the subtasks in $\mathcal{D}_{\mathrm{Subtask}}$ for comparison fairness discussed in \S\ref{subsec:3.3}, we enforce that every hard full task in $\mathcal{D}_{\mathrm{Hard}}$ must have at least one corresponding subtask in $\mathcal{D}_{\mathrm{Subtask}}$, and each subtask in $\mathcal{D}_{\mathrm{Subtask}}$ must correspond to a hard task in $\mathcal{D}_{\mathrm{Hard}}$. 

\vspace{3pt}
\noindent \textbf{Controlling Supervision Error Rate by Adjusting the Proportion of Noisy Supervision.}
Once we ensure that all hard full tasks and easy subtasks have both ground-truth and incorrect solutions, we adjust the supervision quality by substituting a specified percentage of the original ground-truth solutions with incorrect ones. For example, to create hard task supervision with a 20\% error rate, we replace 20\% of the ground-truth solutions with their corresponding incorrect versions.

Overall, after Stage 2, we obtain the final supervision datasets $\mathcal{D}_{\mathrm{Hard}}$ and $\mathcal{D}_{\mathrm{Subtask}}$ with outcome error rates ranging from 0\% to 100\% in 10\% increments, simulating varying supervision quality from different teacher models. As outlined in \S\ref{subsec:3.1}, following \cite{bansal2024smallerweakerbettertraining}, we apply SFT using the two types of supervision to train LLMs on hard reasoning tasks, followed by a comprehensive analysis of their performance.

\subsection{Supervision Sources and Quality Check}
\label{subsec:4.4}

We focus on the math domain and apply the pipeline to various math sources, including three subsets from NuminaMath-CoT \citep{numina_math_datasets}, PRM800K \citep{lightman2023let}, along with the difficulty levels 4-5 from MATH, all featuring \textbf{high school competition, college entrance exam and college-level exam difficulties} (See details in Appendix \ref{sec:sft_data}).
Table \ref{tab:sft_data_details} summarizes the quantity of hard full task and the subtask supervision from each data source. In total, $\mathcal{D}_{\mathrm{Hard}}$ contains 42,274 ground-truth hard task solutions, while $\mathcal{D}_{\mathrm{Subtask}}$ includes 67,328 ground-truth subtask solutions. We then introduce varying error rates, $\epsilon_{\mathrm{Hard}}$ and $\epsilon_{\mathrm{Subtask}}$, for both $\mathcal{D}_{\mathrm{Hard}}$ and $\mathcal{D}_{\mathrm{Subtask}}$. We present a thorough analysis and detailed case studies on the supervision synthesis process, spanning Appendix \ref{subsec:decompose_eg} to \ref{subsec:sampling_eg}, to illustrate the effectiveness of our carefully simulated settings for fair comparison. We conduct human evaluation on 60 samples to further verify the data quality, as described in Appendix \ref{sec:human_eval_sft_data}.


\section{Which Supervision Strategy is Better?}
\label{sec:5}

\subsection{Setup}
\label{subsec:5.1}
We compare the hard math task performance after training with final supervision $\mathcal{D}_{\mathrm{Hard}}$ and $\mathcal{D}_{\mathrm{Subtask}}$ obtained from the two supervision strategies. Our test data of hard tasks encompass a range of difficulty levels, spanning from high school competition to college-level exams. This includes challenges from the Olympic-Arena~\cite{huang2024olympicarena}, MATH, and JEE-Bench~\cite{arora2023have} (as detailed in Appendix~\ref{sec:test_data}). As outlined in \S\ref{subsec:4.3}, we introduce varying error rates from 0\% to 100\% in 10\% increments to simulate different quality levels for both hard and subtask supervision.

Inspired by \citet{sun2024easy}, we first establish a base model with fundamental instruction-following and mathematical reasoning abilities. To this end, we fine-tune Llama-3-8B using elementary school and middle school-level math data from Orca-Math \cite{mitra2024orca} subset of NuminaMath-CoT, as well as the MATH dataset (level 1-3). It serves as the base model that we will train upon using $\mathcal{D}_{\mathrm{Hard}}$ and $\mathcal{D}_{\mathrm{Subtask}}$. Additional training and inference hyper-parameters are provided in Appendix \ref{sec:hyper_param}.

\subsection{Observations}
\label{subsec:5.2}
\begin{figure*}[t]
    \centering
    \begin{subfigure}{0.58\columnwidth}
        \centering
        \scalebox{0.9}{
        \includegraphics[width=\textwidth]{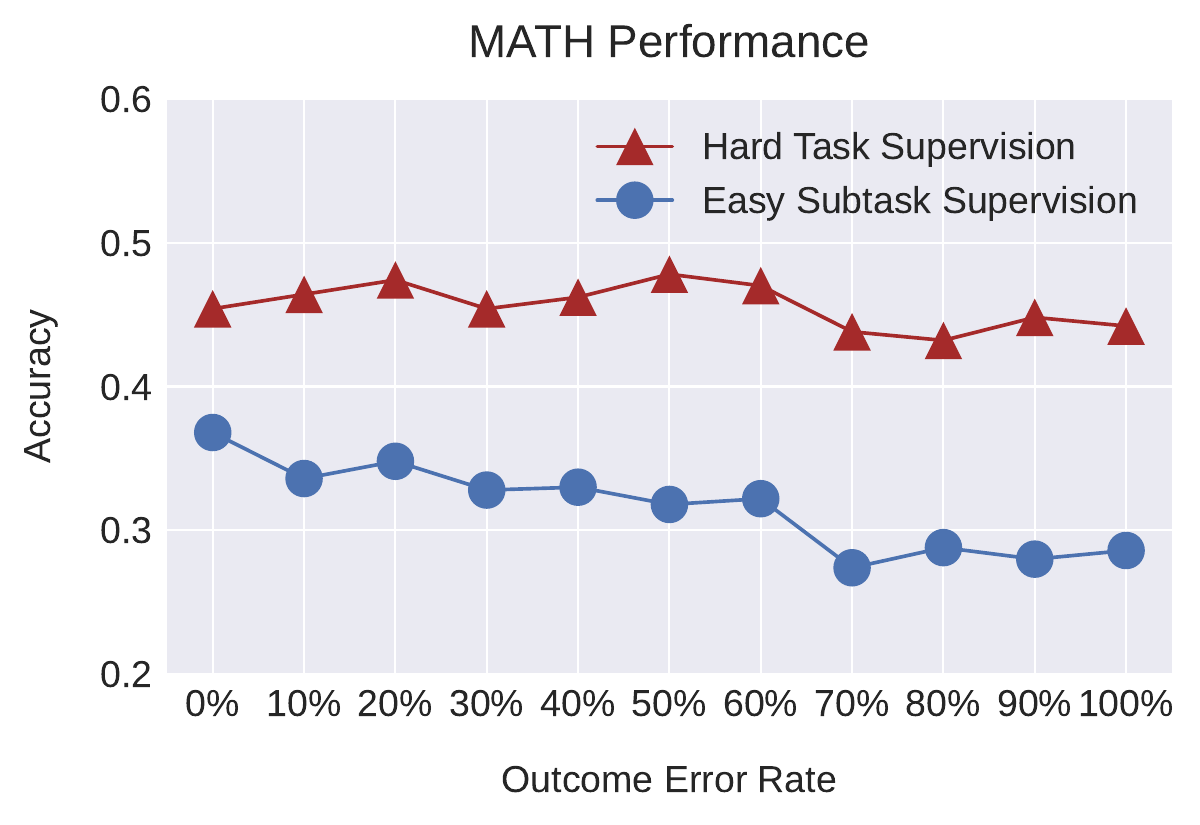}
        } 
        \caption{Accuracy on MATH.}
        \label{fig:math-test500-acc}
    \end{subfigure}
    \hfill
    \begin{subfigure}{0.58\columnwidth}
        \centering
        \scalebox{0.9}{
        \includegraphics[width=\textwidth]{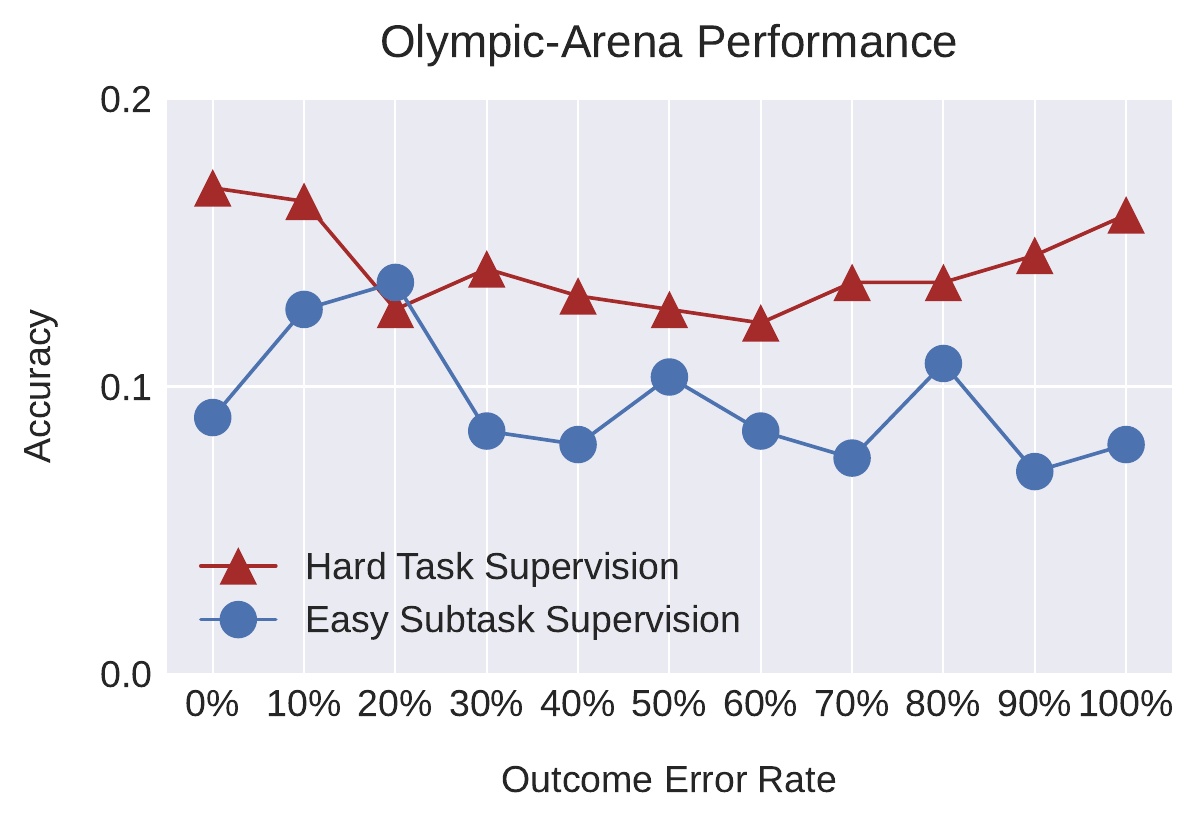}
        } 
        \caption{Accuracy on Olympic-Arena.}
        \label{fig:olympic-arena-acc}
    \end{subfigure}
    \hfill
    \begin{subfigure}{0.58\columnwidth}
        \centering
        \scalebox{0.9}{
        \includegraphics[width=\textwidth]{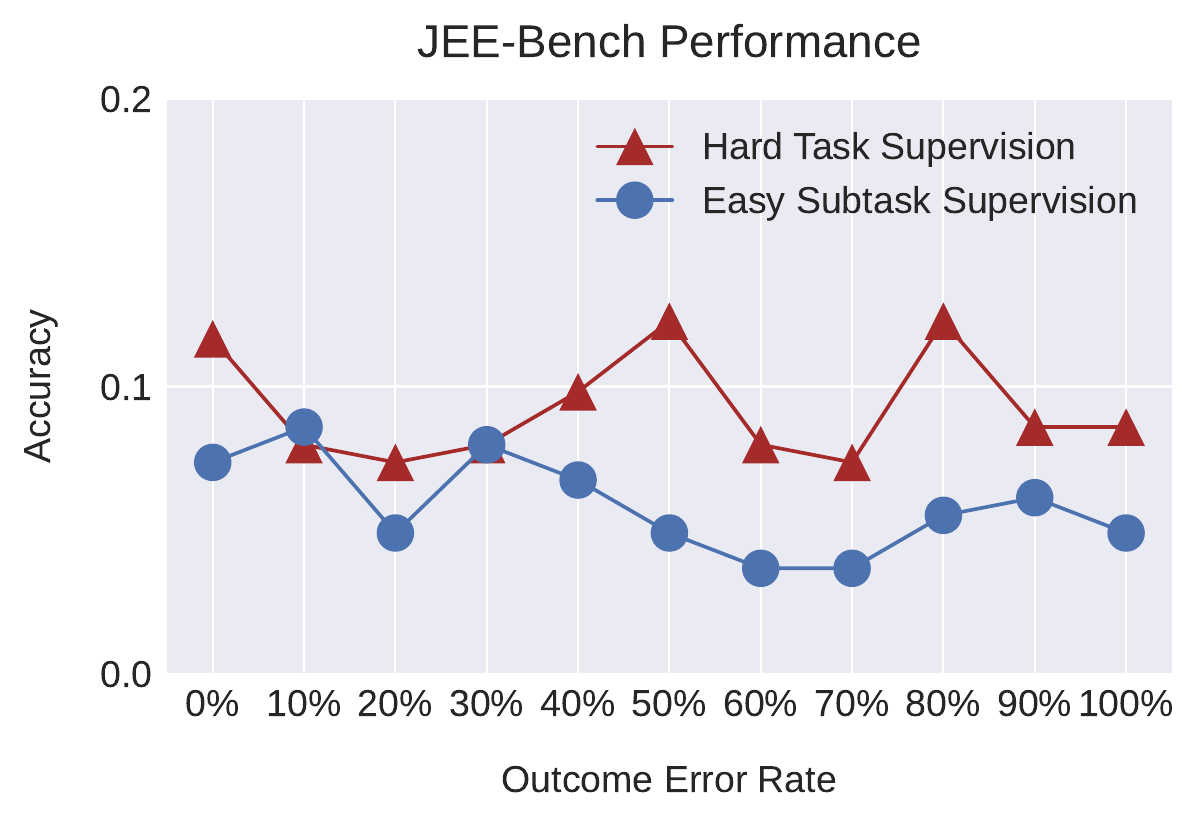}
        } 
        \caption{Accuracy on JEE-Bench.}
        \label{fig:jee-bench-acc}
    \end{subfigure}

    \vspace{0.5em} 

    \begin{subfigure}{0.58\columnwidth}
        \centering
        \scalebox{0.9}{
        \includegraphics[width=\textwidth]{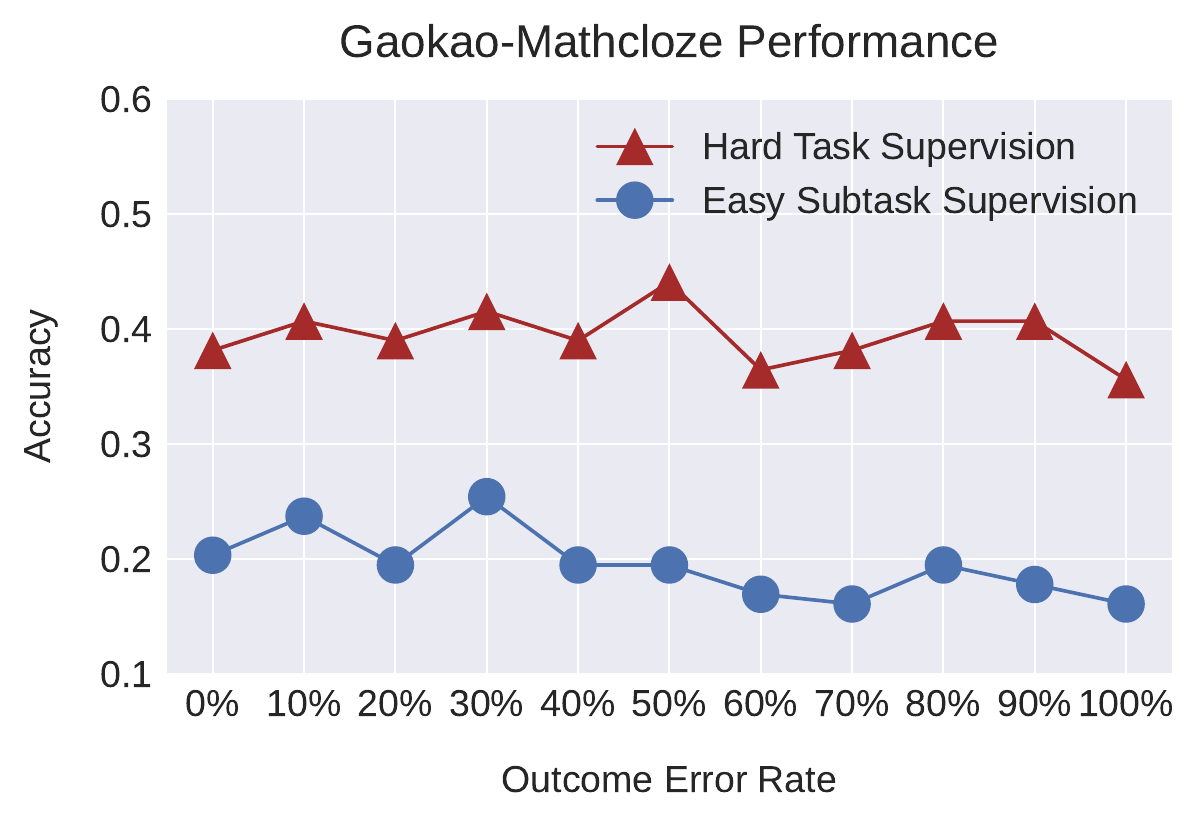}
        } 
        \caption{Accuracy on Gaokao-Mathcloze.}
        \label{fig:gaokao-mathcloze-acc}
    \end{subfigure}
    \hspace{12pt}
    \begin{subfigure}{0.58\columnwidth}
        \centering
        \scalebox{0.9}{
        \includegraphics[width=\textwidth]{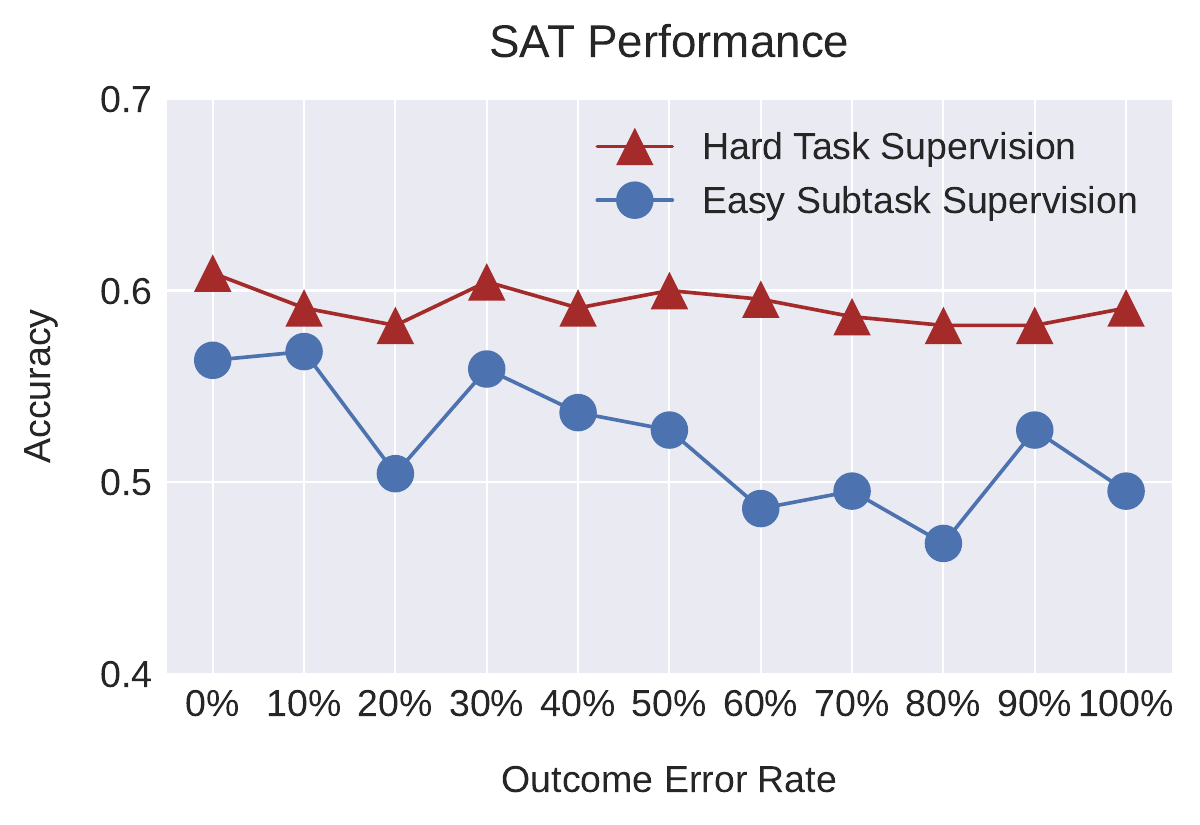}
        } 
        \caption{Accuracy on SAT-Math.}
        \label{fig:sat-acc}
    \end{subfigure}

    \caption{Comparison of easy and hard task supervision with varying outcome ER on 5 hard reasoning benchmarks.}
    \vspace{-3pt}
    \label{fig:hard-downstream-task-acc}
\end{figure*}

Figure \ref{fig:hard-downstream-task-acc} reveals several intriguing insights regarding the comparison between hard task supervision and subtask supervision, as well as the resulting trends in performance changes.

\noindent \textbf{Hard task supervision consistently outperforms subtask supervision, even with higher outcome error rates.} Despite the higher error rate of hard task supervision $\mathcal{D}_{\mathrm{Hard}}$, models trained on $\mathcal{D}_{\mathrm{Hard}}$ still demonstrate superior performance across all five test sets. Remarkably, $\mathcal{D}_{\mathrm{Hard}}$ with a 90\% error rate outperforms $\mathcal{D}_{\mathrm{Subtask}}$ with a 10\% error rate by approximately 11\% on the MATH and 16.9\% on Gaokao-Mathcloze. This suggests that, even with a high error rate, adopting hard task supervision is \emph{seemingly} a more beneficial choice than leveraging more accurate subtask supervision.

\noindent \textbf{Performance does not consistently degrade with increasing outcome error rates.} Across 5 test benchmarks, accuracy remains stable even as the error rate increases. For example, on JEE-Bench, accuracy fluctuates with rising error rates, while on SAT and MATH, performance remains steady, showing marginal changes with variations in outcome error rates. Interestingly, hard task supervision with a 100\% error rate is more effective than with a 10\% error rate on JEE-Bench. This finding is different from previous studies in weak supervision literature \citep{zhang2017understanding}, which generally suggest that supervision quality significantly impacts training performance.


\noindent \textbf{Changes in outcome error rates have a greater impact on subtask supervision than on hard task supervision.}
As shown in Figure \ref{fig:hard-downstream-task-acc}, increasing the outcome error rates in subtask supervision leads to a more significant drop in performance compared to hard task supervision. This is evident in the performance gap between the highest and lowest results achieved among varying error rates. For instance, on MATH, the performance drop for subtask supervision is 9.4\%, whereas for hard full task supervision, it is only 4.6\%. This suggests that the SFT process is more robust to outcome error rate variations when using hard full task supervision compared to subtask supervision.

In summary, it appears that irrespective of the quality of hard full task supervision, it consistently outperforms subtask supervision, even if the latter is more precise in practice. Also, training seems to be more robust across varying quality levels when using hard full task supervision. Why does this occur? Should we always rely on hard full task supervision with high outcome error rates to achieve strong performance on hard tasks? What other factors should we take into account? We explore these questions further in \S\ref{sec:6}.

\section{Severity of Erroneous Solutions Matters}
\label{sec:6}


Observations in \S\ref{sec:5} suggest that in domains requiring long-form reasoning, such as math, \textbf{outcome error rate MAY NOT be a reliable indicator of quality}. Earlier analysis define ``supervision quality'' based on a binary match between the final predicted result and ground truth, ignoring the correctness of intermediate steps. Two incorrect solutions may differ greatly in quality: one with a minor computational error but coherent, and another with fundamentally flawed logic. The latter represents a more severe error, yet this distinction is missed by a simple binary evaluation.


To evaluate error severity in solutions, we introduce the concept of \textbf{erroneous step}, defined as a step that is logically incorrect or irrelevant to preceding steps. Importantly, we do not classify a step as erroneous if it is logically correct but only relies on incorrect intermediate values from prior mistakes. Given similar outcome error rates, it is intuitive to consider that the proportion of erroneous steps affects the quality of supervision. 
To validate this on a larger scale, we first examine the impact of using different teacher models with similar outcome error rates and then analyze whether the severity of step-wise errors correlates with the observed performance gap.






\subsection{Setup}
\label{subsec:6.1}

\begin{figure*}[t]
    \centering
    \begin{subfigure}{0.88\columnwidth}
        \centering
        \scalebox{0.8}{
        \includegraphics[width=\textwidth]{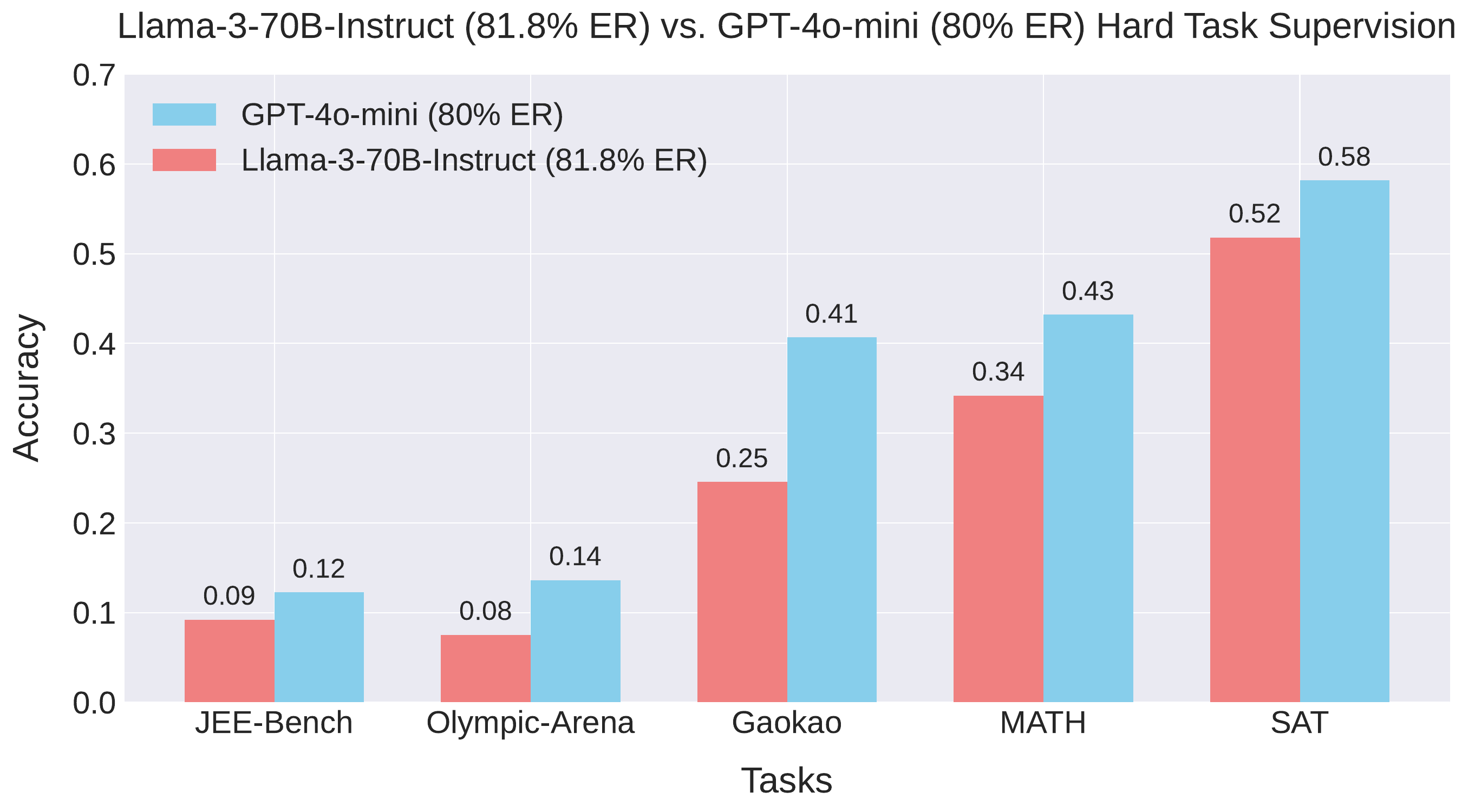}
        } 
        \caption{Llama-3-70B-Instruct (81.7\% ER) vs. GPT-4o-mini (80\% ER) hard full task supervision.}
        \label{fig:llama-3-70b-vs-gpt-4o-mini}
    \end{subfigure}
    \hspace{6pt}
    \begin{subfigure}{0.88\columnwidth}
        \centering
        \scalebox{0.8}{
        \includegraphics[width=\textwidth]{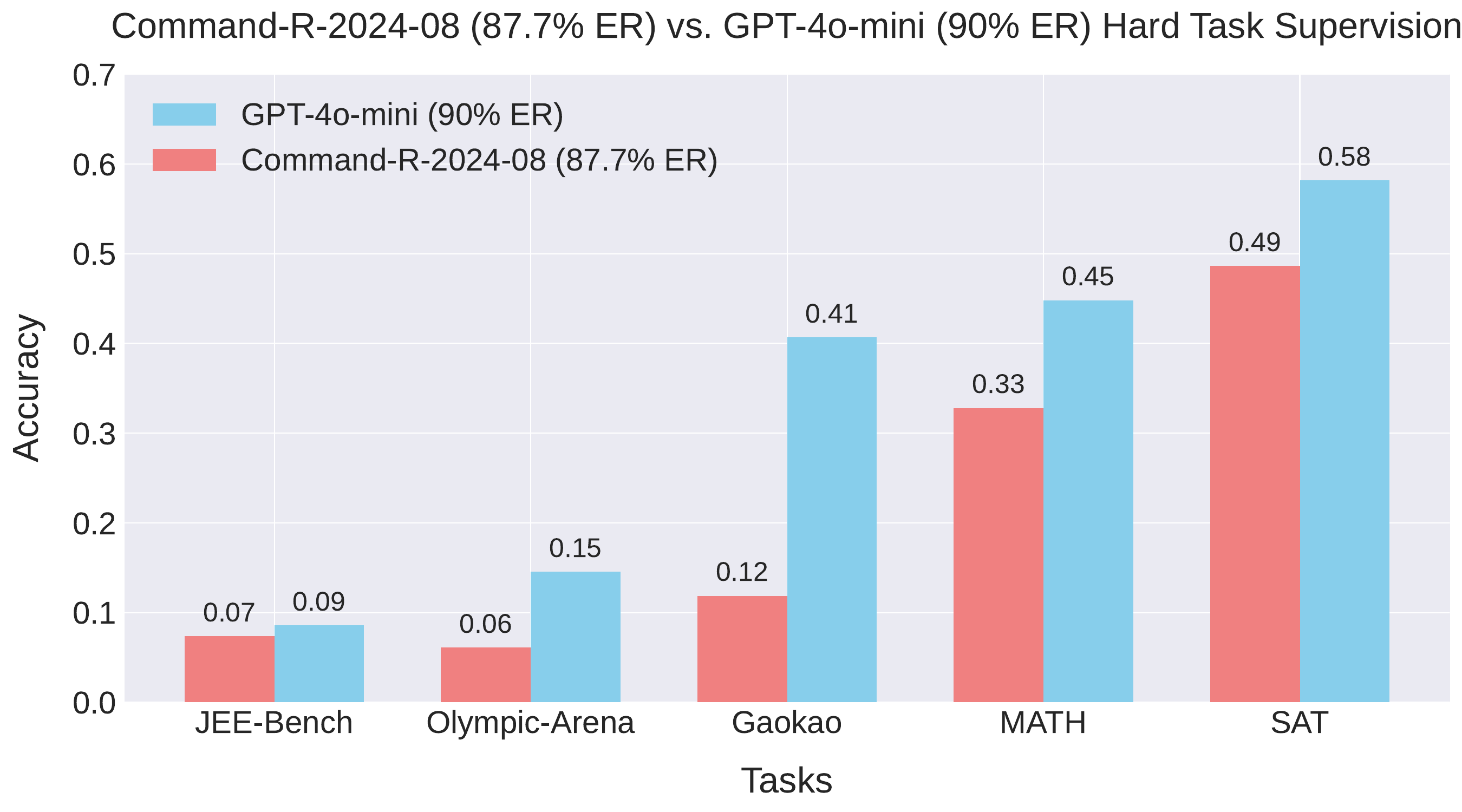}
        } 
        \caption{Command-R-2024-08 (87.7\% ER) vs. GPT-4o-mini (90\% ER) hard full task supervision.}
        \label{fig:command-r-08-vs-gpt-4o-mini}
    \end{subfigure}

    \vspace{0.5em} 

    \begin{subfigure}{0.88\columnwidth}
        \centering
        \scalebox{0.8}{
        \includegraphics[width=\textwidth]{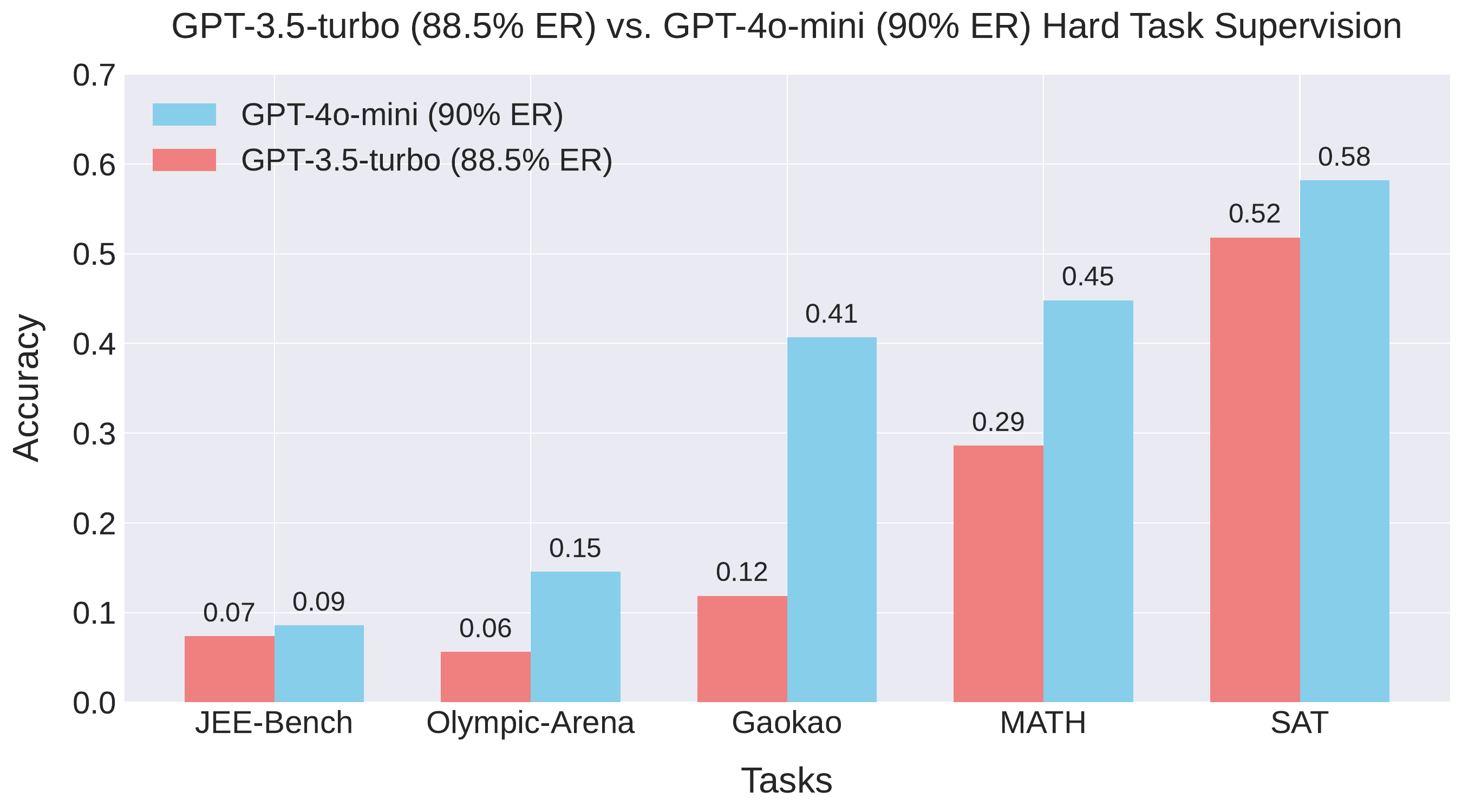}
        } 
        \caption{GPT-3.5-turbo (88.5\% ER) vs. GPT-4o-mini (90\% ER) hard full task supervision.}
        \label{fig:gpt-3.5-vs-gpt-4o-mini}
    \end{subfigure}
    \hspace{3pt}
    \begin{subfigure}{0.88\columnwidth}
        \centering
        \scalebox{0.8}{
        \includegraphics[width=\textwidth]{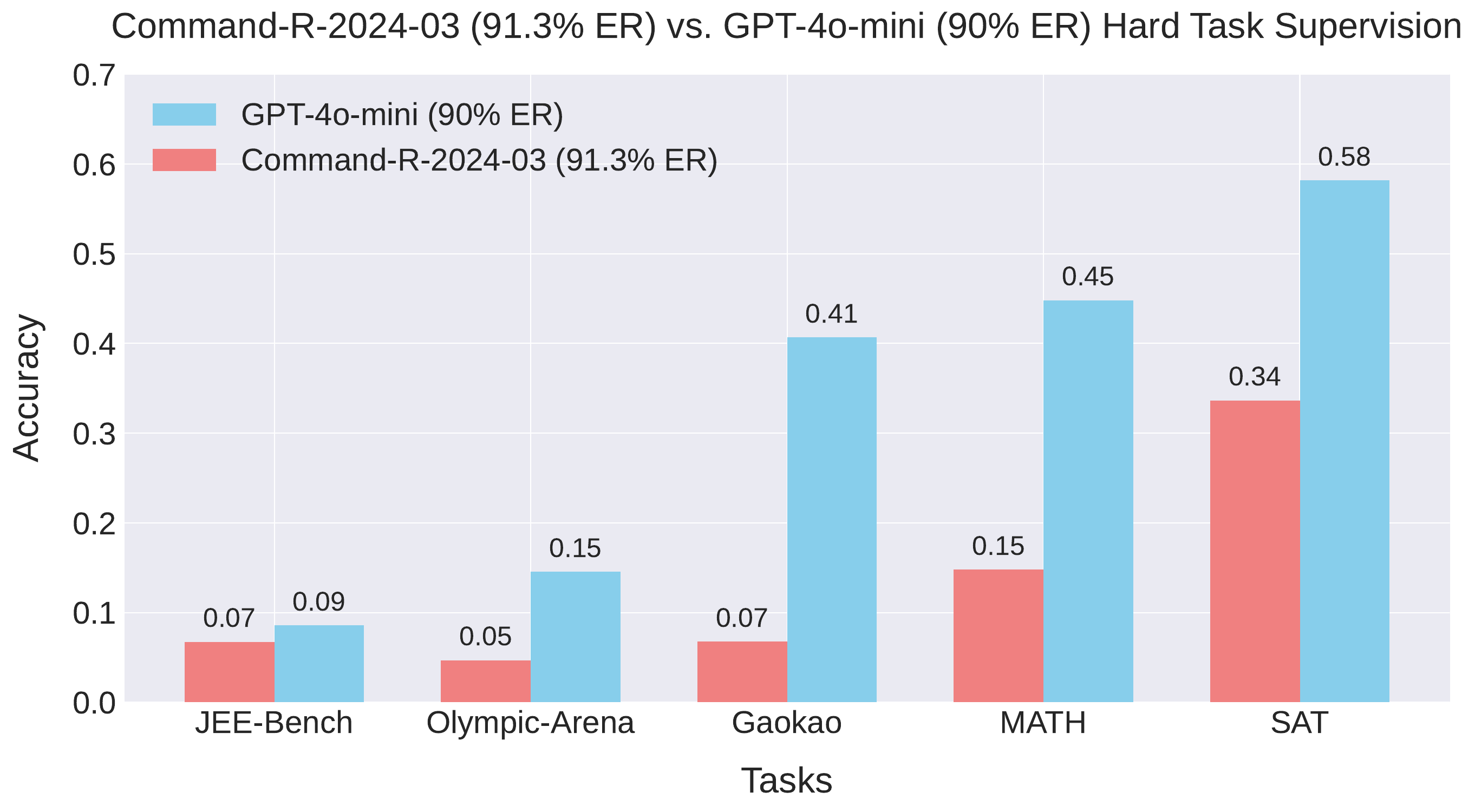}
        } 
        \caption{Command-R-2024-03 (91.3\% ER) vs. GPT-4o-mini (90\% ER) hard full task supervision.}
        \label{fig:command-r-03-vs-gpt-4o-mini}
    \end{subfigure}

    \caption{Performance of the supervision synthesized by different teacher models under similar outcome error rates (a.k.a., ER).}
    \vspace{-3pt}
    \label{fig:different-model-supervision-comparison}
\end{figure*}

In addition to using the solutions previously sampled from GPT-4o-mini, we include GPT-3.5-turbo, Llama-3-70B-Instruct, Command-R-2024-08, and Command-R-2024-03 as \textbf{teacher models with varying capabilities} to generate solutions for constructing new hard full task supervision, denoted as $\mathcal{D'}_{\mathrm{Hard}}$. Concretely, we sample one solution per hard full task in $\mathcal{D}_{\mathrm{Hard}}$ and use these to create $\mathcal{D'}_{\mathrm{Hard}}$ for SFT. For each teacher model, we assess the outcome error rates of $\mathcal{D'}_{\mathrm{Hard}}$ by comparing the predicted final results against the ground-truth answers of all the hard training tasks.

After obtaining the outcome error rates of $\mathcal{D'}_{\mathrm{Hard}}$ generated by different teacher models, we identify the model variant $\mathcal{M}_{\mathrm{Closest}}$ trained with GPT-4o-mini supervision with an outcome error rate closest to the new teacher model for comparison. We then compare $\mathcal{M}_{\mathrm{Closest}}$ with the model trained on $\mathcal{D'}_{\mathrm{Hard}}$ across downstream benchmarks. For example, if the supervision sampled from Llama-3-70B-Instruct has an outcome error rate of 81.8\% on the hard tasks, we compare Llama-3-70B-Instruct sampled supervision with the corresponding model in \S\ref{subsec:5.2} trained by GPT-4o-mini sampled supervision with an 80\% outcome error rate. \looseness=-1 

\subsection{Performance of Supervision Under Similar Outcome Error Rates}
\label{subsec:6.2}

As shown in Figure \ref{fig:different-model-supervision-comparison}, even with similar outcome error rates, models trained with supervision from diverse teacher models behave differently. For instance, a model trained on solutions from Command-R-2024-08, which has an outcome error rate of 87.7\%, achieves only 32.8\% accuracy on MATH. In contrast, the model trained on data from GPT-4o-mini, despite a higher error rate of 90\%, achieves significantly better accuracy at 44.8\%. Additionally, the model trained with supervision from Llama-3-70B-Instruct (81.8\% error rate) performs worse than one trained with 100\% incorrect data from GPT-4o-mini. Notably, Command-R-2024-03's sampled hard task supervision is less effective than using the subtask supervision from GPT-4o-mini in 4 out of 5 test hard tasks.
This suggests the key factor influencing performance may lie in the quality of the solution steps embedded within the supervision from different teacher models.

\subsection{Correlation Between Performance Disparity and Step-Wise Error Rates}
\label{subsec:6.3}

\begin{table}
    \centering
    \vspace{-3pt}
    \small
    \begin{tabular}{lc}\toprule
     \textbf{Teacher models}                 & \textbf{Step-wise Error Ratio (\%)} \\ \midrule
GPT-4o-mini       &       \textbf{11.7}                \\ \midrule
Llama-3-70B-Instruct       &       30.2                \\
Command-R-2024-08 &       40.0                \\
Command-R-2024-03 &       62.6                \\
GPT-3.5-turbo     &      34.7                 \\
    \bottomrule 
    \end{tabular}
    \caption{Step-wise error rates (\%) for teacher models.}
    \vspace{-15pt}
    \label{tab:step-wise-error-rate}
\end{table}

To further validate our hypothesis about the severity of incorrect solutions generated by different teacher model models, human evaluation is required. We first collect 60 tasks along with their incorrect solutions from the supervision sampled by each of the four studied teacher models discussed in \S\ref{subsec:6.1}, including Command-R-2024-08 and Llama-3-70B-Instruct. We then conduct a manual review of these incorrect solutions. Specifically, we count the total steps and erroneous steps in each solution and calculate \textbf{step-wise error rates} by averaging the ratio of erroneous steps to total steps across all the solutions of sampled 60 examples (See Appendix \ref{sec:case_study} for comprehensive case studies). As shown in Table \ref{tab:step-wise-error-rate}, the step-wise error rates vary among these teacher models, with Command-R-2024-03 exhibiting the highest ratio at 62.6\% and GPT-4o-mini having the lowest at 11.7\%. 
From Figure \ref{fig:different-model-supervision-comparison}, we observe a strong correlation between the step-wise error rates and downstream performance: lower rates correspond to higher downstream performance, aligning more closely with results from supervision synthesized by GPT-4o-mini. This indicates that the step-wise error rate is a strong indicator of supervision quality and its impact on hard task performance. Besides the above observations, we attempt to borrow step-wise error rate metric to explain more phenomenons, detailed in Appendix \ref{sec:other_phenom}.

\section{Further Improvement Over Sole Hard Task Supervision}
\label{sec:7}





As outlined previously, hard task supervision with lower step-wise error rates is highly beneficial. 
But can performance be further improved using existing hard task supervision without introducing new ones? 
A natural approach is to integrate easy subtask supervision into existing supervision sources. 
Since these subtasks are decomposed from the hard ones, this strategy boosts task diversity without adding new information, offering a potential avenue for further performance gains.

What happens after we combine them? Due to the substantial resources and time required to explore all possible combinations (a total of $11 \times 11$ combinations) of different outcome error rates for hard and easy data,
we select several combinations: (20\%, 10\%), (20\%, 0\%), (50\%, 10\%), (50\%, 0\%), (80\%, 40\%), (80\%, 10\%) and (80\%, 0\%), 
where the two values represent the outcome error rates for hard full task and easy subtask supervision. In each combination, the outcome error rate for hard data is higher, reflecting the reality that annotations for hard tasks are often less accurate. 

From another perspective, since the information in hard tasks essentially covers that of the subtasks, one might argue that combining easy subtask and hard full task supervision merely makes LLMs learn the same information roughly twice. We design a baseline denoted as ``doubled epochs'', where we train the model on hard task supervision for 4 epochs, which double the training epochs of all previous experiments. We also merge the existing hard task supervision along with its rephrased version which increases more diversity.

\paragraph{Combined Supervision vs. Doubled Epochs vs. Merge Rephrased.} Table \ref{tab:combination-and-other-baselines} shows that in most cases, combining easy subtask and hard task supervision enhances performance. For instance, combining hard task supervision with a 50\% outcome error rate and subtask supervision with a 10\% outcome error rate leads to the highest accuracy on the MATH, Gaokao-Mathcloze and SAT-Math among all previously trained baselines. It also outperforms doubling the training epochs or mixing rephrased data with original SFT supervision, highlighting that integrating subtask supervision can push performance limits beyond repetitive supervision.


\begin{table*}[t]
    \centering
    \vspace{-1.5em}
    \begin{subtable}[ht]{0.95\textwidth}
    \centering
    \small
    \setlength\tabcolsep{.5mm}
    \begin{tabular}{lcccccccccc}
    \toprule
    \multirow{2}{*}{\textbf{Tasks}} & \multicolumn{3}{c}{\textbf{Combination 1}}   & \multicolumn{3}{c}{\textbf{Combination 2}}      & \multicolumn{4}{c}{\textbf{Combination 3}}                         \\
    \cmidrule(lr){2-4}
    \cmidrule(lr){5-7}
    \cmidrule(lr){8-11}
                           & (20\%, 10\%) & (20\%,0\%) & 20\% & (50\%, 10\%) & (50\%,0\%) & 50\% & (80\%, 40\%) & (80\%, 10\%) & (80\%, 0\%) & 80\% \\
    \midrule

MATH             & 43.2                    & \textbf{48.0}                & 47.4             & \textbf{48.6}           & 45.8                       & 47.8             & 43.8                    & 46.8                        & \textbf{47.2}              & 43.2             \\
Olympic-Arena    & 13.6                    & \textbf{16.4}              & 12.7             & 12.7                    & \textbf{18.8}              & 12.7             & 14.1                    & \textbf{19.2}               & 17.8                       & 13.6             \\
JEE-Bench        & \textbf{12.3}           & 10.4                       & 7.4              & 11.7                    & 9.2                        & \textbf{12.3}    & \textbf{12.3}           & 9.8                         & 7.4                        & \textbf{12.3}    \\
Gaokao-Mathcloze & \textbf{40.7}           & \textbf{40.7}              & 39               & \textbf{45.8}           & 43.2                       & 44.1             & 42.4                    & 37.3                        & \textbf{43.2}              & 40.7             \\
SAT-Math         & 58.2                    & \textbf{60.5}              & 58.2             & 61.0                      & \textbf{63.2}              & 60.0               & 60.0                      & \textbf{66.8}               & 62.3                       & 58.2             \\
\midrule
Average          & 33.6                    & \textbf{35.2}              & 32.9            & 36.0                   & \textbf{36.0}             & 35.4            & 34.5                   & \textbf{36.0}              & 35.6             & 33.6            \\

    \bottomrule
    \end{tabular}
    \caption{Combination of easy and hard task supervision vs. Sole hard task supervision.}
    \label{tab:full-sub-combination}
    \end{subtable}

    \hfill

    \begin{subtable}[ht]{0.95\textwidth}
    \centering
    \small
    \begin{tabular}{lccc}
    \toprule
    \multirow{2}{*}{\textbf{Tasks}} & \multicolumn{3}{c}{\textbf{Combination}}                                     \\
    \cmidrule(lr){2-4}
                           & (50\%, 10\%) & 50\% (Doubled Epochs) & 50\% (Merge Rephrased.) \\
    \midrule
    MATH           & \hlpink{48.6}                                    & 45.0       &   48.4                \\
    Olympic-Arena                       &  12.7                                       &   12.7           &     12.7                   \\
    JEE-Bench                       &  \hlpink{11.7}                                       &  9.2         &   10.4                        \\
    Gaokao-Mathcloze                       &  \hlpink{45.8}                                       &  42.4            &   44.1                     \\
    SAT-Math                       &  \hlpink{61.0}                                       &  55.5          &  57.3          \\ 
    \bottomrule
    \end{tabular}
    \caption{Combined vs. Doubled Epochs vs. Merge Rephrased.}
    \vspace{-3pt}
    \label{tab:compare-with-double-epochs}
    \end{subtable}

\caption{Accuracy (\%) of different combination methods. (a) combination of hard and easy task supervision with varying error rates, (b) other combinations: "Doubled Epochs" refers to doubling original training epochs, while "Merge Rephrased" denotes the training data that combines original hard tasks with the rephrased versions.}
\label{tab:combination-and-other-baselines}
\end{table*}

\paragraph{Effects of Step-Wise Error Rates of Hard Task Supervision after Merging.} We also combined hard task supervision from other different teacher models (Llama-3-70B-Instruct and Command-R-2024-03) and the original easy subtask supervision. We then compared these new combinations with the original ones to further assess the effect of step-wise error rate in hard task supervision, as discussed in \S\ref{sec:6}. The results align with our earlier findings: when a teacher model exhibits a lower step-wise error rate, the performance of the combined hard and easy supervision improves. As shown in Table \ref{tab:step-wise-error-rate}, Command-R-2024-03, which has the highest step-wise error rate, yields the lowest accuracy on the test data. This comparison reinforces the importance of step-wise error rate in determining the quality of math reasoning supervision data for hard downstream tasks.

\begin{table*}[ht]
    \centering
    \small
    \begin{tabular}{lccc}
    \toprule
     & \multicolumn{3}{c}{\textbf{Combination}}                                     \\

    \cmidrule(lr){1-4}
        \textbf{Outcome ER Combinations}   & \makecell{GPT-4o-mini \\(80\%, 10\%)} &  \makecell{Llama-3-70B-Instruct \\(81.7\%, 10\%)} & \makecell{Command-R-2024-03 \\(87.7\%, 10\%)} \\
    \cmidrule(lr){1-4}
        \textbf{Step-Wise ER of Hard Task Supervision}    & 11.7\% &  30.2\% & 62.6\% \\
    \midrule
    MATH           & \hlpink{46.8}                                    & 31.2       &   24.6                \\
    Olympic-Arena                       &  \hlpink{19.2}                                       &   12.2           &     5.2                   \\
    JEE-Bench                       &  \hlpink{8.7}                                       &  10.4         &   5.5                        \\
    Gaokao-Mathcloze                       &  \hlpink{37.3}                                       &  23.7            &   5.1                     \\
    SAT-Math                       &  \hlpink{61.0}                                       &  54.5          &  47.7          \\ 
    \bottomrule
    \end{tabular}
    \caption{Combination performance of hard full task supervision from different teacher models and original easy subtask supervision, teacher models include GPT-4o-mini, Llama-3-70B-Instruct and Command-R-2024-03.}
    \vspace{-6pt}
    \label{tab:combine-other-teachers}
\end{table*}


\section{Concluded Empirical Advice}
\label{sec:8}

Based on the findings above, we offer some practical advice for optimizing annotation efforts of weak teacher models to enhance on hard tasks:

\noindent \textbf{Prioritize hard task supervision for better performance on challenging reasoning tasks.} We reveal in \S\ref{sec:5} that the best performance is consistently achieved with hard full task supervision. It significantly surpasses the results obtained using only subtask supervision. 

\noindent \textbf{Step-wise error rates of teacher models is more critical than outcome error rates on hard reasoning task performance.} In \S\ref{sec:6}, we find that, even with identical outcome error rates, supervision with higher step-wise error rates leads to more evident performance drop than what we observe when the outcome error rates change.

\noindent \textbf{To further enhance performance, consider sampling or annotating hard tasks and their associated subtasks, then use the combined supervision  to train LLMs.} As shown in \S\ref{sec:7}, given a fixed task scope covered in the hard task supervision, integrating hard task supervision and their corresponding subtask supervision boosts or even achieves the best performance across benchmarks. This combination is also more effective than simply increasing the training epochs or further merging the rephrased version of the original supervision.

\section{Related Works}
\label{sec:2}


\noindent \textbf{Weak Supervision.}
Weak supervision has traditionally been applied to text classification tasks, typically following an iterative framework that generates pseudo-labels using heuristics and expands the training set 
\citep{agichtein2000snowball,meng2018weakly}. Recent works have employed LLM-based teacher models to generate training data for fine-tuning smaller models \citep{bosselut-etal-2019-comet,west-etal-2022-symbolic,wang-etal-2023-self-instruct,hsieh-etal-2023-distilling,brahman2023plasma}, enabling student models to follow instructions and perform reasoning. This approach assumes that a strong teacher LLM offers high-quality supervision. However, our focus shifts to challenging reasoning tasks that even a current strong LLM struggles to solve. We analyze how supervision complexity and quality jointly impacts performance when a teacher model provides noisy weak supervision for hard tasks. 

\vspace{3pt}
\noindent \textbf{Easy-to-Hard Generalization.}
This area \citep{zhou2022least,burns2023weak} involves training on easy task supervision and then evaluating on more difficult examples. \citet{hase2024unreasonableeffectivenesseasytraining} demonstrate that in-context learning or fine-tuning on easy tasks offers a strong baseline for multiple-choice tasks such as ARC \citep{clark2018think} and MMLU \citep{hendrycksmeasuring}, while \citet{sun2024easy} show that an easy-to-hard evaluator enhances a generator’s generalization on hard task benchmark MATH, based on the fact that evaluation is easier to learn than generation. In contrast to prior works, our study uniquely focuses on how two extra correlated aspects, supervision task difficulty and supervision quality, jointly influence final performance on hard tasks. Our experimental setup provides two key advantages: 1) one of the most comprehensive evaluations of hard reasoning tasks, far exceeding prior studies; and 2) a fairer comparison by controlling task scope to keep easy subtasks and hard full task supervision within a similar range.

\section{Conclusion}
\label{sec:9}
In this paper, we explore two possible supervision strategies that weak teacher models can employ to enhance performance on hard reasoning tasks. We first find that even when the outcome error rate for hard task supervision is high, training on this data can surpass the performance of perfectly correct supervision from easier subtasks across multiple hard math tasks. Then, we pinpoint step-wise error rates as a more critical factor influencing hard task performance, as its variation can result in a more evident drop on the hard tasks. We further observe that if no more hard task supervision is allowed to be added, supplementing existing hard task supervision with corresponding subtask supervision yields greater performance than merely merging rephrased hard task supervision. These insights contribute valuable data-driven guidance for researchers and practitioners aiming to improve LLMs in future reasoning-heavy applications.

\section*{Limitations}
\label{sec:limitations}

In this section, we list several limitations of our current work and consider to address them to enhance the paper in the future.

\vspace{3pt}
\noindent \textbf{Effects of Step-Wise Error Rates.} In \S\ref{sec:6}, we focused on examining the effects of step-wise error rates on the quality of supervision for hard full tasks, primarily due to cost considerations. Future work will include an investigation into the impact of step-wise error rates on subtask supervision. Additionally, we can perturb the supervision produced by the same teacher models to introduce more step-wise erroneous steps, further reinforcing the significance of step-wise error rates.

\vspace{3pt}
\noindent \textbf{Supervision Combination Strategy.} The experiments presented in \S\ref{sec:7} are conducted with a limited range of error rate combinations for hard full task and easy subtask supervision provided by GPT-4o-mini. We plan to test a broader array of combinations and extend our analysis to the supervision synthesized by other teacher models, such as GPT-3.5-turbo and Llama-3-70B-Instruct, if our training budget allows. 

\vspace{3pt}
\noindent \textbf{Exploration on More Reasoning Domains.} Our experiments primarily target the math domain; however, we intend to explore other reasoning domains, such as scientific reasoning, to further bolster the arguments presented in our study.

\vspace{3pt}
\noindent \textbf{Integrating More Alignment Methods into Our Supervision Strategy Analysis Framework.} Following \citet{bansal2024smallerweakerbettertraining,hase2024unreasonableeffectivenesseasytraining}, we employ supervised fine-tuning (SFT) to investigate the effects of various types of supervision. Additionally, we plan to integrate more alignment training methods, such as DPO~\cite{rafailov2024direct}, to determine whether our observations remain consistent in the future work.

\section*{Acknowledgement}
This research is based upon work supported by NSF CAREER \#2339766, and an Amazon AGI Research Award. We thank anonymous reviewers and UCLA NLP group members for their valuable feedback.


\bibliography{acl_latex}

\begin{thebibliography}{37}
\providecommand{\natexlab}[1]{#1}

\bibitem[{Agichtein and Gravano(2000)}]{agichtein2000snowball}
Eugene Agichtein and Luis Gravano. 2000.
\newblock Snowball: Extracting relations from large plain-text collections.
\newblock In \emph{Proceedings of the fifth ACM conference on Digital libraries}, pages 85--94.

\bibitem[{Anthropic(2024)}]{claude_blog}
Anthropic. 2024.
\newblock \href {https://www.anthropic.com/news/claude-3-5-sonnet} {{Claude 3.5 Sonnet}}.

\bibitem[{Arora et~al.(2023{\natexlab{a}})Arora, Singh et~al.}]{arora2023have}
Daman Arora, Himanshu Singh, et~al. 2023{\natexlab{a}}.
\newblock Have llms advanced enough? a challenging problem solving benchmark for large language models.
\newblock In \emph{Proceedings of the 2023 Conference on Empirical Methods in Natural Language Processing}, pages 7527--7543.

\bibitem[{Arora et~al.(2023{\natexlab{b}})Arora, Singh, and Mausam}]{Arora2023HaveLA}
Daman Arora, Himanshu~Gaurav Singh, and Mausam. 2023{\natexlab{b}}.
\newblock \href {https://api.semanticscholar.org/CorpusID:258866000} {Have llms advanced enough? a challenging problem solving benchmark for large language models}.
\newblock \emph{ArXiv}, abs/2305.15074.

\bibitem[{Bansal et~al.(2024)Bansal, Hosseini, Agarwal, Tran, and Kazemi}]{bansal2024smallerweakerbettertraining}
Hritik Bansal, Arian Hosseini, Rishabh Agarwal, Vinh~Q. Tran, and Mehran Kazemi. 2024.
\newblock \href {https://arxiv.org/abs/2408.16737} {Smaller, weaker, yet better: Training llm reasoners via compute-optimal sampling}.
\newblock \emph{Preprint}, arXiv:2408.16737.

\bibitem[{Bosselut et~al.(2019)Bosselut, Rashkin, Sap, Malaviya, Celikyilmaz, and Choi}]{bosselut-etal-2019-comet}
Antoine Bosselut, Hannah Rashkin, Maarten Sap, Chaitanya Malaviya, Asli Celikyilmaz, and Yejin Choi. 2019.
\newblock \href {https://doi.org/10.18653/v1/P19-1470} {{COMET}: Commonsense transformers for automatic knowledge graph construction}.
\newblock In \emph{Proceedings of the 57th Annual Meeting of the Association for Computational Linguistics}, pages 4762--4779, Florence, Italy. Association for Computational Linguistics.

\bibitem[{Bowman et~al.(2022)Bowman, Hyun, Perez, Chen, Pettit, Heiner, Luko{\v{s}}i{\=u}t{\.e}, Askell, Jones, Chen et~al.}]{bowman2022measuring}
Samuel~R Bowman, Jeeyoon Hyun, Ethan Perez, Edwin Chen, Craig Pettit, Scott Heiner, Kamil{\.e} Luko{\v{s}}i{\=u}t{\.e}, Amanda Askell, Andy Jones, Anna Chen, et~al. 2022.
\newblock Measuring progress on scalable oversight for large language models.
\newblock \emph{arXiv preprint arXiv:2211.03540}.

\bibitem[{Brahman et~al.(2023)Brahman, Bhagavatula, Pyatkin, Hwang, Li, Arai, Sanyal, Sakaguchi, Ren, and Choi}]{brahman2023plasma}
Faeze Brahman, Chandra Bhagavatula, Valentina Pyatkin, Jena~D Hwang, Xiang~Lorraine Li, Hirona~J Arai, Soumya Sanyal, Keisuke Sakaguchi, Xiang Ren, and Yejin Choi. 2023.
\newblock Plasma: Making small language models better procedural knowledge models for (counterfactual) planning.
\newblock \emph{arXiv preprint arXiv:2305.19472}.

\bibitem[{Burns et~al.(2023)Burns, Izmailov, Kirchner, Baker, Gao, Aschenbrenner, Chen, Ecoffet, Joglekar, Leike et~al.}]{burns2023weak}
Collin Burns, Pavel Izmailov, Jan~Hendrik Kirchner, Bowen Baker, Leo Gao, Leopold Aschenbrenner, Yining Chen, Adrien Ecoffet, Manas Joglekar, Jan Leike, et~al. 2023.
\newblock Weak-to-strong generalization: Eliciting strong capabilities with weak supervision.
\newblock \emph{arXiv preprint arXiv:2312.09390}.

\bibitem[{Clark et~al.(2018)Clark, Cowhey, Etzioni, Khot, Sabharwal, Schoenick, and Tafjord}]{clark2018think}
Peter Clark, Isaac Cowhey, Oren Etzioni, Tushar Khot, Ashish Sabharwal, Carissa Schoenick, and Oyvind Tafjord. 2018.
\newblock Think you have solved question answering? try arc, the ai2 reasoning challenge.
\newblock \emph{arXiv preprint arXiv:1803.05457}.

\bibitem[{Cohere(2024)}]{command-r_blog}
Cohere. 2024.
\newblock \href {https://cohere.com/command} {{Command-R}}.

\bibitem[{Dubey et~al.(2024)Dubey, Jauhri, Pandey, Kadian, Al-Dahle, Letman, Mathur, Schelten, Yang, Fan et~al.}]{dubey2024llama}
Abhimanyu Dubey, Abhinav Jauhri, Abhinav Pandey, Abhishek Kadian, Ahmad Al-Dahle, Aiesha Letman, Akhil Mathur, Alan Schelten, Amy Yang, Angela Fan, et~al. 2024.
\newblock The llama 3 herd of models.
\newblock \emph{arXiv preprint arXiv:2407.21783}.

\bibitem[{Glasmachers(2017)}]{glasmachers2017limits}
Tobias Glasmachers. 2017.
\newblock Limits of end-to-end learning.
\newblock In \emph{Asian conference on machine learning}, pages 17--32. PMLR.

\bibitem[{G{\"u}l{\c{c}}ehre and Bengio(2016)}]{gulccehre2016knowledge}
{\c{C}}aǧlar G{\"u}l{\c{c}}ehre and Yoshua Bengio. 2016.
\newblock Knowledge matters: Importance of prior information for optimization.
\newblock \emph{The Journal of Machine Learning Research}, 17(1):226--257.

\bibitem[{Hase et~al.(2024)Hase, Bansal, Clark, and Wiegreffe}]{hase2024unreasonableeffectivenesseasytraining}
Peter Hase, Mohit Bansal, Peter Clark, and Sarah Wiegreffe. 2024.
\newblock \href {https://arxiv.org/abs/2401.06751} {The unreasonable effectiveness of easy training data for hard tasks}.
\newblock \emph{Preprint}, arXiv:2401.06751.

\bibitem[{Hendrycks et~al.(2021{\natexlab{a}})Hendrycks, Burns, Basart, Zou, Mazeika, Song, and Steinhardt}]{hendrycksmeasuring}
Dan Hendrycks, Collin Burns, Steven Basart, Andy Zou, Mantas Mazeika, Dawn Song, and Jacob Steinhardt. 2021{\natexlab{a}}.
\newblock Measuring massive multitask language understanding.
\newblock In \emph{International Conference on Learning Representations}.

\bibitem[{Hendrycks et~al.(2021{\natexlab{b}})Hendrycks, Burns, Kadavath, Arora, Basart, Tang, Song, and Steinhardt}]{hendrycks2measuring}
Dan Hendrycks, Collin Burns, Saurav Kadavath, Akul Arora, Steven Basart, Eric Tang, Dawn Song, and Jacob Steinhardt. 2021{\natexlab{b}}.
\newblock Measuring mathematical problem solving with the math dataset.
\newblock In \emph{Thirty-fifth Conference on Neural Information Processing Systems Datasets and Benchmarks Track (Round 2)}.

\bibitem[{Hsieh et~al.(2023)Hsieh, Li, Yeh, Nakhost, Fujii, Ratner, Krishna, Lee, and Pfister}]{hsieh-etal-2023-distilling}
Cheng-Yu Hsieh, Chun-Liang Li, Chih-kuan Yeh, Hootan Nakhost, Yasuhisa Fujii, Alex Ratner, Ranjay Krishna, Chen-Yu Lee, and Tomas Pfister. 2023.
\newblock \href {https://doi.org/10.18653/v1/2023.findings-acl.507} {Distilling step-by-step! outperforming larger language models with less training data and smaller model sizes}.
\newblock In \emph{Findings of the Association for Computational Linguistics: ACL 2023}, pages 8003--8017, Toronto, Canada. Association for Computational Linguistics.

\bibitem[{Huang et~al.(2024)Huang, Wang, Xia, Li, Zou, Xu, Fan, Ye, Chern, Ye et~al.}]{huang2024olympicarena}
Zhen Huang, Zengzhi Wang, Shijie Xia, Xuefeng Li, Haoyang Zou, Ruijie Xu, Run-Ze Fan, Lyumanshan Ye, Ethan Chern, Yixin Ye, et~al. 2024.
\newblock Olympicarena: Benchmarking multi-discipline cognitive reasoning for superintelligent ai.
\newblock \emph{arXiv preprint arXiv:2406.12753}.

\bibitem[{Jothimurugan et~al.(2023)Jothimurugan, Hsu, Bastani, and Alur}]{jothimurugan2023robust}
Kishor Jothimurugan, Steve Hsu, Osbert Bastani, and Rajeev Alur. 2023.
\newblock Robust subtask learning for compositional generalization.
\newblock In \emph{International Conference on Machine Learning}, pages 15371--15387. PMLR.

\bibitem[{Li et~al.(2024{\natexlab{a}})Li, Wang, Hu, Wei, Zheng, Hu, Zhang, and Peng}]{Li2024Common7L}
Chen Li, Weiqi Wang, Jingcheng Hu, Yixuan Wei, Nanning Zheng, Han Hu, Zheng Zhang, and Houwen Peng. 2024{\natexlab{a}}.
\newblock \href {https://api.semanticscholar.org/CorpusID:268264074} {Common 7b language models already possess strong math capabilities}.
\newblock \emph{ArXiv}, abs/2403.04706.

\bibitem[{Li et~al.(2024{\natexlab{b}})Li, Beeching, Tunstall, Lipkin, Soletskyi, Huang, Rasul, Yu, Jiang, Shen, Qin, Dong, Zhou, Fleureau, Lample, and Polu}]{numina_math_datasets}
Jia Li, Edward Beeching, Lewis Tunstall, Ben Lipkin, Roman Soletskyi, Shengyi~Costa Huang, Kashif Rasul, Longhui Yu, Albert Jiang, Ziju Shen, Zihan Qin, Bin Dong, Li~Zhou, Yann Fleureau, Guillaume Lample, and Stanislas Polu. 2024{\natexlab{b}}.
\newblock Numinamath.
\newblock \url{[https://huggingface.co/AI-MO/NuminaMath-CoT](https://github.com/project-numina/aimo-progress-prize/blob/main/report/numina_dataset.pdf)}.

\bibitem[{Lightman et~al.(2023)Lightman, Kosaraju, Burda, Edwards, Baker, Lee, Leike, Schulman, Sutskever, and Cobbe}]{lightman2023let}
Hunter Lightman, Vineet Kosaraju, Yura Burda, Harri Edwards, Bowen Baker, Teddy Lee, Jan Leike, John Schulman, Ilya Sutskever, and Karl Cobbe. 2023.
\newblock Let's verify step by step.
\newblock \emph{arXiv preprint arXiv:2305.20050}.

\bibitem[{Meng et~al.(2018)Meng, Shen, Zhang, and Han}]{meng2018weakly}
Yu~Meng, Jiaming Shen, Chao Zhang, and Jiawei Han. 2018.
\newblock Weakly-supervised neural text classification.
\newblock In \emph{proceedings of the 27th ACM International Conference on information and knowledge management}, pages 983--992.

\bibitem[{Mitra et~al.(2024)Mitra, Khanpour, Rosset, and Awadallah}]{mitra2024orca}
Arindam Mitra, Hamed Khanpour, Corby Rosset, and Ahmed Awadallah. 2024.
\newblock Orca-math: Unlocking the potential of slms in grade school math.
\newblock \emph{arXiv preprint arXiv:2402.14830}.

\bibitem[{OpenAI(2022)}]{chatgpt_paper}
OpenAI. 2022.
\newblock \href {https://openai.com/blog/chatgpt} {{ChatGPT}}.

\bibitem[{OpenAI(2024{\natexlab{a}})}]{gpt-4o-mini_blog}
OpenAI. 2024{\natexlab{a}}.
\newblock \href {https://openai.com/index/gpt-4o-mini-advancing-cost-efficient-intelligence/} {{GPT-4o mini: advancing cost-efficient intelligence}}.

\bibitem[{OpenAI(2024{\natexlab{b}})}]{gpt4o_paper}
OpenAI. 2024{\natexlab{b}}.
\newblock \href {https://openai.com/index/gpt-4o-system-card/} {{GPT-4o System Card}}.

\bibitem[{Rafailov et~al.(2024)Rafailov, Sharma, Mitchell, Manning, Ermon, and Finn}]{rafailov2024direct}
Rafael Rafailov, Archit Sharma, Eric Mitchell, Christopher~D Manning, Stefano Ermon, and Chelsea Finn. 2024.
\newblock Direct preference optimization: Your language model is secretly a reward model.
\newblock \emph{Advances in Neural Information Processing Systems}, 36.

\bibitem[{Reid et~al.(2024)Reid, Savinov, Teplyashin, Lepikhin, Lillicrap, Alayrac, Soricut, Lazaridou, Firat, Schrittwieser et~al.}]{reid2024gemini}
Machel Reid, Nikolay Savinov, Denis Teplyashin, Dmitry Lepikhin, Timothy Lillicrap, Jean-baptiste Alayrac, Radu Soricut, Angeliki Lazaridou, Orhan Firat, Julian Schrittwieser, et~al. 2024.
\newblock Gemini 1.5: Unlocking multimodal understanding across millions of tokens of context.
\newblock \emph{arXiv preprint arXiv:2403.05530}.

\bibitem[{Sun et~al.(2024)Sun, Yu, Shen, Liu, Yang, Welleck, and Gan}]{sun2024easy}
Zhiqing Sun, Longhui Yu, Yikang Shen, Weiyang Liu, Yiming Yang, Sean Welleck, and Chuang Gan. 2024.
\newblock Easy-to-hard generalization: Scalable alignment beyond human supervision.
\newblock \emph{arXiv preprint arXiv:2403.09472}.

\bibitem[{Wang et~al.(2023)Wang, Kordi, Mishra, Liu, Smith, Khashabi, and Hajishirzi}]{wang-etal-2023-self-instruct}
Yizhong Wang, Yeganeh Kordi, Swaroop Mishra, Alisa Liu, Noah~A. Smith, Daniel Khashabi, and Hannaneh Hajishirzi. 2023.
\newblock \href {https://doi.org/10.18653/v1/2023.acl-long.754} {Self-instruct: Aligning language models with self-generated instructions}.
\newblock In \emph{Proceedings of the 61st Annual Meeting of the Association for Computational Linguistics (Volume 1: Long Papers)}, pages 13484--13508, Toronto, Canada. Association for Computational Linguistics.

\bibitem[{West et~al.(2022)West, Bhagavatula, Hessel, Hwang, Jiang, Le~Bras, Lu, Welleck, and Choi}]{west-etal-2022-symbolic}
Peter West, Chandra Bhagavatula, Jack Hessel, Jena Hwang, Liwei Jiang, Ronan Le~Bras, Ximing Lu, Sean Welleck, and Yejin Choi. 2022.
\newblock \href {https://doi.org/10.18653/v1/2022.naacl-main.341} {Symbolic knowledge distillation: from general language models to commonsense models}.
\newblock In \emph{Proceedings of the 2022 Conference of the North American Chapter of the Association for Computational Linguistics: Human Language Technologies}, pages 4602--4625, Seattle, United States. Association for Computational Linguistics.

\bibitem[{Wies et~al.(2023)Wies, Levine, and Shashua}]{wiessub}
Noam Wies, Yoav Levine, and Amnon Shashua. 2023.
\newblock Sub-task decomposition enables learning in sequence to sequence tasks.
\newblock In \emph{The Eleventh International Conference on Learning Representations}.

\bibitem[{Zhang et~al.(2017)Zhang, Bengio, Hardt, Recht, and Vinyals}]{zhang2017understanding}
Chiyuan Zhang, Samy Bengio, Moritz Hardt, Benjamin Recht, and Oriol Vinyals. 2017.
\newblock Understanding deep learning requires rethinking generalization.
\newblock In \emph{International Conference on Learning Representations}.

\bibitem[{Zhong et~al.(2023)Zhong, Cui, Guo, Liang, Lu, Wang, Saied, Chen, and Duan}]{Zhong2023AGIEvalAH}
Wanjun Zhong, Ruixiang Cui, Yiduo Guo, Yaobo Liang, Shuai Lu, Yanlin Wang, Amin Saied~Sanosi Saied, Weizhu Chen, and Nan Duan. 2023.
\newblock \href {https://api.semanticscholar.org/CorpusID:258108259} {Agieval: A human-centric benchmark for evaluating foundation models}.
\newblock \emph{ArXiv}, abs/2304.06364.

\bibitem[{Zhou et~al.(2022)Zhou, Sch{\"a}rli, Hou, Wei, Scales, Wang, Schuurmans, Cui, Bousquet, Le et~al.}]{zhou2022least}
Denny Zhou, Nathanael Sch{\"a}rli, Le~Hou, Jason Wei, Nathan Scales, Xuezhi Wang, Dale Schuurmans, Claire Cui, Olivier Bousquet, Quoc Le, et~al. 2022.
\newblock Least-to-most prompting enables complex reasoning in large language models.
\newblock \emph{arXiv preprint arXiv:2205.10625}.

\end{thebibliography}

\clearpage

\appendix

\appendix
\label{sec:appendix}










\section{Hyperparameters for Training and Inference}
\label{sec:hyper_param}
As outlined in \S\ref{subsec:5.1}, we fine-tune Llama-3-8B base model using the elementary school and middle school-level math task annotations from the Orca-Math subset of NuminaMath-CoT, and MATH (level 1-3) to establish a base model equipped with foundational instruction-following and mathematical reasoning capabilities. In total, there are 164,162 training instances. Subsequently, we fine-tune this base model with our synthesized easy subtask and hard full task supervision. We set the learning rate to $2e-5$ and employed the cosine learning rate scheduler. 
Each model variant is trained on A100 and H100 GPUs for 2 epochs, taking around 1 hour. During inference on downstream tasks, we set temperature as 0.6, top-p as 0.9, top-k as 50 and maximal output token numbers as 1680.

\section{Supervision Source Data Details}
\label{sec:sft_data}

Table \ref{tab:sft_data_details} shows the original size, the number of tasks generated by the synthesis pipeline and the difficulty level of the five data sources for synthesizing hard task and easy subtask supervision. For the three subsets of NuminaMath-CoT, ``cn\_k12'' consists of problems collected and translated from Chinese K-12 education math exercises. ``synthetic\_math'' is synthesized from the MATH dataset following the methods in \cite{Li2024Common7L} and ``synthetic\_amc'' subset is similarly synthesized based on the training split of the AMC-AIME dataset. Regarding the remaining two sources, PRM800K, initially a preference dataset, is utilized here by taking positive answers from the preference pairs as solutions to the problems. For MATH, we exclusively used its training split with difficulty level 4-5.

\begin{table*}[!ht]
\small
\centering
\scalebox{0.85}{
\begin{tabular}{lcccc}
\toprule 
\textbf{Sources} & \textbf{Original Sizes} & \textbf{\# Hard Sup.} & \textbf{\# Easy Sup.} & \textbf{Difficulty}\\
\midrule 

NuminaMath-CoT (cn\_k12) & 105,785 & 22,397 & 35,210 & high-school \& college entrance exam\\

NuminaMath-CoT (synthetic\_math)  & 40,580 & 11,172 & 18,107 & high school\\
 
NuminaMath-CoT (synthetic\_amc)   & 31,752 & 7,547 & 12,190 & high school competition\\

PRM800k & 2,041 & 340 &  523 & high school\\

MATH (level 4-5) & 1,040  & 150 & 261 & high school competition\\

\bottomrule
\end{tabular}
}
\caption{Data sources used for constructing hard and easy task supervision, including their original sizes, final sizes after filtering (as described in \S\ref{subsec:4.3}) that the data sources contribute, and corresponding difficulty levels.}
\label{tab:sft_data_details}
\end{table*}

\section{Test Data Details}
\label{sec:test_data}

Our test data contains five subset and Table \ref{tab:test_data_details} shows the size and level of difficulty of downstream tasks in these five subsets. 
Our MATH test set is a small set split of size 500 from the original MATH test set, as mentioned in \cite{lightman2023let}. 

JEE-Bench \citep{Arora2023HaveLA} is derived from the IIT JEE-Advanced Exam, with high-school and college level problems. Some are multiple choice or fill-in-the-blank questions, we convert them to open-ended questions using GPT-4o-mini and filter out some unqualified converted problems like ill-defined ones or the ones with incorrect solutions.

Olympic-Arena \citep{huang2024olympicarena} includes 11,163 text-only and interleaved text-image problems, covering multiple subjects like mathematics, physics, computer science, etc. We select the text-only math problems as our test data. Some problems are multiple choice or fill-in-the-blank questions, we also convert them to open-ended questions and manually filter out the unqualified ones, such as questions not well-defined or those with wrong solutions after conversion.

The other two subsets are sourced from the benchmark AGI-Eval \citep{Zhong2023AGIEvalAH}, SAT contains math problems from SAT exam in U.S. while Gaokao-Mathcloze is sourced from Chinese College Entrance Exam. For these two subsets, we also do format conversion and automatic filtering as described above.  

\begin{table}[ht]
\small
\centering
\scalebox{0.85}{
\begin{tabular}{lcccc}
\toprule 
\textbf{Test Tasks} & \textbf{Sizes} &  \textbf{Difficulty} \\
\midrule 
MATH & 500 & high school competition\\

JEE-Bench & 163 & high school \& college\\
 
Olympic-Arena & 213 & Olympiad competition\\

SAT-Math & 220 & college entrance exam\\

Gaokao-Mathcloze & 118 & college entrance exam\\
\bottomrule 
\end{tabular}
}
\caption{Detailed statistics of 5 downstream hard math reasoning tasks and their corresponding difficulty levels.}
\label{tab:test_data_details}
\end{table}

\section{Details of Data Synthesis Pipeline and Examples for Demonstration}
\label{sec:pipeline_eg}
This section offers an in-depth demonstration of our pipeline through multiple examples, highlighting the effectiveness and comprehensiveness of our supervision synthesis approach discussed in \S\ref{sec:4}. 

\subsection{Decomposing Hard Full Task into Subtasks}
\label{subsec:decompose_eg}
At Stage 1 of our pipeline (\S\ref{subsec:4.2}), we decompose the original hard full task into 2 or 3 subtasks. Here we show some examples of the original hard full tasks and the decomposed subtasks.

\noindent \textbf{Decomposition Example 1}: 
Table \ref{tab:demo_decompose_eg1_full}, \ref{tab:demo_decompose_eg1_sub1} and  \ref{tab:demo_decompose_eg1_sub2} show the original hard full task, decomposed subtask 1 and decomposed subtask 2, respectively, as our first example.

\noindent \textbf{Decomposition Example 2}: 
Table \ref{tab:demo_decompose_eg2_full}, \ref{tab:demo_decompose_eg2_sub1} and \ref{tab:demo_decompose_eg2_sub2} show the original hard full task, decomposed subtask 1 and decomposed subtask 2, respectively, as our second example.

\subsection{Filtering Unqualified Subtasks}
\label{subsec:filtering_eg}
As mentioned in \S\ref{subsec:4.2}, we use GPT-4o-mini to filter out the unqualified subtasks. Our filtering criteria are shown below:

\begin{itemize}[leftmargin=*]
     \item \textbf{Incomplete and Ill-defined Subtasks.} We filter out the decomposed subtasks that lack necessary conditions for solving problem, especially those where their solutions mention the information not present in the task. For example, if a solution mentions ``$f(x)$ is an odd function'' or uses the condition ``Anna has three apples'' but these conditions do not exist in the subtask, the subtask would not be utilized in the final supervision.

    \item \textbf{Overly Simple Subtasks.} We prompt GPT-4o-mini to count the computation and reasoning steps in each solution, filtering out easy subtasks that involve fewer than a threshold of 3 steps. These overly simple tasks, which require minimal reasoning, can already be handled by current LLMs and don't need to be retaught. For instance, the discarded subtasks include ``Calculate sine of 30°'', ``Solve for $ 2x+6=18 $'' or ``What's the area of semicircle if its radius is 3?'', which are too simple and can be solved within 3 steps.

    \item \textbf{Subtasks That Are Too Similar to the Original Hard Full Tasks.} We prompt GPT-4o-mini to evaluate whether a decomposed subtask is merely a variation of the original full hard task, such as only changes in numerical values or conditions, and filter out such unqualified tasks to guarantee the lower difficulty of the subtasks. For example, if the original task is ``Calculate the derivative of $f(x)$ and determine its monotonicity for $a = 1$,'' and the subtask is ``Determine whether $f(x)$ is increasing or decreasing for $a = 3$ by calculating its derivative,'' the subtask is simply a variant of the original task and thus excluded. 
\end{itemize}

\noindent \textbf{Type 1 (Ill-defined) - Example 1.} 
Table \ref{tab:demo_filtering1_eg1} shows the first example for Type 1 (ill-defined) of unqualified subtask.

\noindent \textbf{Type 1 (Ill-defined) - Example 2.} 
Table \ref{tab:demo_filtering1_eg2} shows the second example for Type 1 (ill-defined) of unqualified subtask.

\noindent \textbf{Type 2 (Overly simple) - Example 1.} 
Table \ref{tab:demo_filtering2_eg1} shows the first example for Type 2 (overly simple) of unqualified subtask.

\noindent \textbf{Type 2 (Overly simple) - Example 2.} 
Table \ref{tab:demo_filtering2_eg2} shows the second example for Type 2 (overly simple) of unqualified subtask.

\noindent \textbf{Type 3 (Too similar to the original) - Example 1.} 
Table \ref{tab:demo_filtering3_eg1} shows the first example for Type 3 (too similar to original hard task) of unqualified subtask.

\noindent \textbf{Type 3 (Too similar to the original) - Example 2.} 
Table \ref{tab:demo_filtering3_eg2} shows the second example for the type 3 (too similar to the original hard full task) of unqualified subtask.

\subsection{Sampling Multiple Solutions for Both Hard and Subtasks}
\label{subsec:sampling_eg}
To adjust the outcome error rates in the supervision set during Stage 2 of the supervision synthesis process (\S\ref{subsec:4.3}), each task is paired with solutions that include both a ground-truth and an incorrect option. To obtain the incorrect solution, we sample three potential solutions for each hard full task and easy subtask in the supervision sets, then employ GPT-4o-mini to validate their correctness against the ground-truth final answer. 

With regard to the correctness of the three sampled solutions for each hard full task and decomposed subtask, there are three situations: \textbf{Situation 1: All Correct} - All the three sampled solutions are correct; \textbf{Situation 2: Mixed} - The three sampled solutions contain both correct and incorrect solutions; \textbf{Situation 3: All Incorrect} - All the three sampled solutions are incorrect.
In Situation 1, tasks without incorrect solutions are excluded from the final supervision set. In Situation 2, a correct sampled solution is used as the ground-truth, and an incorrect one is used for error rate adjustment. In Situation 3, an incorrect solution is selected and the ground truth is reformatted to match the style of incorrect solution. We ensure consistency between ground-truth and incorrect solutions to prevent LLMs from learning divergent styles during fine-tuning.

Once we construct the pairs of correct solution and incorrect solution, the incorrect solution will then be used to replace the ground-truth answer when we intentionally introduce more outcome errors to manage quality. Below, we provide examples that demonstrate the multi-sampling process and the creation of these solution pairs.

\noindent \textbf{Sampling and Construction Example 1.}
Table \ref{tab:demo_multi_sampling_eg1} presents the first example of our process for sampling three solutions from LLMs and constructing solution pairs that include these three solutions and the ground-truth.

\noindent \textbf{Sampling and Construction Example 2.}
Table \ref{tab:demo_multi_sampling_eg2} presents the second example of our process for sampling three solutions from LLMs and constructing solution pairs that include these three solutions and the ground-truth.

\noindent \textbf{Sampling and Construction Example 3.}
Table \ref{tab:demo_multi_sampling_eg3} presents the third example of our process for sampling three solutions from LLMs and constructing solution pairs that include these three solutions and the ground-truth.

\section{Human Evaluation for SFT Supervision Quality}
\label{sec:human_eval_sft_data}
To further assess the quality of the synthesized data, we randomly sample a small set with 60 hard full tasks and their corresponding 87 easy subtasks, and conduct a human evaluation for these samples to examine data quality. Our evaluation suite includes the following dimensions: 
1) checking correct solutions for both hard full tasks and subtasks; 
2) checking incorrect solutions for both hard full tasks and subtasks;  
3) checking the quality and appropriateness of the subtasks. 

\paragraph{Correct Solution Checking.}
The correct solutions are required to be complete, correct and detailed, showing each essential computing and reasoning step. For hard full task, \textbf{59/60} are qualified and for easy subtask, \textbf{85/87} is qualified.

\paragraph{Incorrect Solution Checking.}
The incorrect solutions should reach incorrect final result and be detailed in solving process. \textbf{54/60} in hard full tasks is qualified while \textbf{83/87} in subtasks is qualified. Most of the unqualified solutions exhibit correct process and results, while very few are incomplete. 

\paragraph{Decomposed Subtask Checking.}
As mentioned in \S\ref{subsec:3.3}, tasks in subtask supervision is decomposed from the tasks in hard full task supervision and we have filtered out three types of unqualified subtasks, which corresponds to three essential features we expect in the subtask:

\begin{itemize}[leftmargin=*]
\item \textit{Criteria 1: Subtask should be well-defined, with no missing conditions.} In our review, \textbf{84/87} subtasks are qualified. Those unqualified are all missing conditions or information in the problem while the solution mentions and utilizes them. 

\item \textit{Criteria 2: Subtask should not be overly simple, featuring less than 3 steps in solution.} In our review, \textbf{all the easy subtasks} are qualified.

\item \textit{Criteria 3: Subtask should not be too similar to the hard full task, with only some numerical conditions changed.} In this dimension, \textbf{86/87} easy subtasks are qualified. 

\end{itemize}

With human evaluation, we further verify the overall quality of synthesized hard and easy data, which can be regarded as reliable foundations of our controlled supervision strategy analysis.

\section{Measure Possible Discrepancy between Hard Full tasks and Decomposed Subtasks}
\label{sec:gap_between_full_sub}
To investigate the potential discrepancies between hard full task questions and their decomposed subtask counterparts, which may be challenging for humans to detect, we measure the average cosine similarity of embeddings between these two task types on a small sample set. Additionally, we employ GPT-4o-mini to perform a style transfer on the question components of the sampled subtasks, aligning them with the full task questions in terms of descriptive manner, contextual setup, and overall style. The prompt used is shown in Table \ref{tab:prompt_subtask_style_transfer}.
Then we compare the similarity metric before and after this style transfer to further verify if there are some style differences between full task and subtask, which may potentially contribute to the performance gap between these two types of supervision data.

\begin{table*}[!ht]
\renewcommand{\arraystretch}{1.2}
\small
\centering
\begin{tabular}{p{0.9\textwidth}
}
\toprule
Below we provide some mathematical tasks, including several examples and one task T to be refined and modified in style. \\

Please modify the task T to make its style is as similar as possible to that of the examples, making the question type (if example tasks are not MCQ, then task T MUST NOT be MCQ), description manner, contextual setup, and the overall style are aligned closely. **BUT DO NOT CHANGE THE ORIGINAL MEANING OF THE TASK T**. \\

Input format: \\
Example Task 1: <example task 1> \\
Example Task 2: <example task 2> \\
Example Task 3: <example task 3> \\
Task to be refined: <task T to be modified> \\
Here we provide the examples and the task T to be modified, please ONLY output the modified version of task T, NO OTHER TEXTS!\\

\bottomrule
    \end{tabular}
\caption{Prompt template for transferring style of the questioin part of subtasks to the style of full task in our measure of the style differences between two types of tasks.}
\label{tab:prompt_subtask_style_transfer}
\end{table*}

We sampled 200 full tasks, each paired with a subtask, to compute the embedding similarity using popular models such as \texttt{bert-base-v2} from the BERT family and \texttt{paraphrase-mpnet-base-v2} from the Sentence Transformer family. The results are presented in Table \ref{tab:full_sub_embed_cos_sim}. The findings show that the cosine similarity between the embeddings of full tasks and their corresponding original subtasks remains consistently high, ranging from $0.7$ to $0.85$ across different embedding models. Additionally, after style transfer, the subtasks maintain the high similarity scores with the full tasks, with no significant differences compared to the original subtasks. These results further confirm that the gap between the two types of supervision data is marginal.

\begin{table*}[]
\centering
\small
\begin{tabular}{cccc}
\toprule
\textbf{Embedding Model}          & \multicolumn{1}{c}{$\mathrm{Sim}$(Full Task, Original Subtask)} & \multicolumn{1}{c}{$\mathrm{Sim}$(Full Task, Style Transferred Subtask)} & $|\delta|$ \\
\midrule
\texttt{albert-base}              & 0.8311                                            & 0.8327   & 0.0016                                          \\
\texttt{bert-base-v2}             & 0.8549                                            & 0.8535  & 0.0014                                           \\
\texttt{paraphrase-MiniLM-L12-v2} & 0.6970                                            & 0.6993  & 0.0023                                           \\
\texttt{paraphrase-mpnet-base-v2} & 0.7498                & 0.7493      & 0.0005           \\

\bottomrule
\end{tabular}
\caption{Cosine similarity between the embeddings of full task and subtask text instructions, and the change in similarity after subtask style transfer. The marginal variation suggests that the subtask and full task instructions are already highly similar before applying any post-hoc methods to reduce the style disparity between them.}
\label{tab:full_sub_embed_cos_sim}
\end{table*}

\section{Other Phenomenons Explained by Step-Wise Error Rates}
\label{sec:other_phenom}

In addition to the observations and conclusions regarding the correlation between step-wise error rates and downstream performance, several phenomena can be explained by this new metric:

\vspace{3pt}
\noindent \textbf{To be Explained 1: When outcome error rates in hard full task supervision vary, performance remains relatively stable.}
As discussed in \S\ref{subsec:5.2}, GPT-4o-mini serves as the teacher model for hard full tasks and human evaluation in \S\ref{subsec:6.3} shows GPT-4o-mini has a low step-wise error rates, with most errors involving just a single incorrect step, while the rest are accurate. This observation suggests that, despite the varying outcome error rates, the proportion of noisy supervision was minimal, which explains why models fine-tuned with hard full task supervision are robust to varying outcome error rates, as seen in \S\ref{subsec:5.2}, where performance remains stable across different outcome error rates on both MATH and SAT-Math.




\vspace{3pt}
\noindent \textbf{To be Explained 2: Hard full task supervision consistently outperforms subtask supervision.} 
We can imagine that, for example, where a complex task requires 10 steps, while a simpler task only needs 5 steps. If both tasks contain one erroneous step, the step-wise error rates for the subtask becomes proportionally higher.
In line with this reasoning, our sample set shows that, for hard full tasks, there is an average of 1.36 erroneous steps out of an average of 13.08 total steps, while subtasks have 1.14 erroneous steps out of 5.96 total steps. The number of erroneous steps is similar between hard full task and easy subtask supervision. 
This step-wise rate difference may become one of the possible factors about why easy subtask supervision tends to be less effective than hard full task supervision under identical outcome error rates, resulting in consistently lower performance.

\begin{table*}[!ht]
\renewcommand{\arraystretch}{1.5}
\small
\centering
\scalebox{0.8}{
\begin{tabular}{p{1.2\textwidth}}
\toprule
\textbf{Hard task (full task) of Example 1:} 
\\
In an arithmetic sequence $\{a_{n}\}$ where all terms are distinct, $a_{1}=1$, and $a_{1}$, $a_{2}$, $a_{5}$ form a geometric sequence. The sum of the first $n$ terms of sequence $\{b_{n}\}$ is $S_{n}=2^{n+1}-2$.  
(1) Find the general formulas for sequences $\{a_{n}\}$ and $\{b_{n}\}$.  
(2) Let ${c_n}={2^{{a_n}}}+{\log_2}{b_n}$, find the sum of the first $n$ terms of sequence $\{c_{n}\}$ denoted as $T_{n}$.
\\
\midrule
\textbf{Solution:}
\\

Part (1): Finding General Formulas

- Sequence $\{a_{n}\}$:

Given $a_{1}=1$, let the common difference be $d$. Thus, we have:
- $a_{2}=a_{1}+d=1+d$
- $a_{5}=a_{1}+4d=1+4d$

Since $a_{1}$, $a_{2}$, and $a_{5}$ form a geometric sequence, we apply the property that the square of the middle term equals the product of the first and third terms:
$$
(a_{1}+d)^{2} = a_{1}(a_{1}+4d)
$$
Substituting $a_{1}=1$ gives:
$$
(1+d)^{2} = 1(1+4d)
$$
Simplifying, we find:
$$
1+2d+d^{2} = 1+4d
$$
$$
d^{2}-2d = 0
$$
$$
d(d-2) = 0
$$
Thus, $d=0$ or $d=2$. Since $d=0$ would imply no progression (which contradicts the distinct terms condition), we discard it and accept $d=2$. Therefore, the general formula for $\{a_{n}\}$ is:
$$
a_{n} = 1 + (n-1)d = 1 + 2(n-1) = 2n-1
$$

- Sequence $\{b_{n}\}$:

Given $S_{n}=2^{n+1}-2$, the $n$th term $b_{n}$ can be found by subtracting the sum of the first $n-1$ terms from the sum of the first $n$ terms:
$$
b_{n} = S_{n} - S_{n-1} = (2^{n+1}-2) - (2^{n}-2) = 2 \times 2^{n} - 2^{n} = 2^{n}
$$
For $n=1$, we directly verify that $b_{1}=2$ fits the formula. Thus, the general formula for $\{b_{n}\}$ is:
$$
b_{n} = 2^{n}
$$

\#\#\#\# Part (2): Sum of the First $n$ Terms of Sequence $\{c_{n}\}$

Given $c_{n} = 2^{a_{n}} + \log_{2}{b_{n}}$ and using the formulas derived in part (1), we have:
$$
c_{n} = 2^{2n-1} + n
$$
To find the sum $T_{n}$ of the first $n$ terms of $\{c_{n}\}$:
$$
T_{n} = \sum_{k=1}^{n} (2^{2k-1} + k)
$$
Breaking it down:
$$
T_{n} = (2 + 2^{3} + 2^{5} + \ldots + 2^{2n-1}) + (1 + 2 + 3 + \ldots + n)
$$
Using the formula for the sum of a geometric series and the sum of the first $n$ natural numbers:
$$
T_{n} = \frac{2(1-4^{n})}{1-4} + \frac{n(n+1)}{2}
$$
Simplifying:
$$
T_{n} = \frac{2^{2n+1}-2}{3} + \frac{n^{2}+n}{2}
$$

Therefore, the sum of the first $n$ terms of sequence $\{c_{n}\}$ is:
$$
\boxed{T_{n} = \frac{2^{2n+1}-2}{3} + \frac{n^{2}+n}{2}}
$$.
\\
\bottomrule
\end{tabular}
}
\caption{Decomposition - Example 1: Hard task.}
\label{tab:demo_decompose_eg1_full}
\end{table*}

\begin{table*}[!ht]
\renewcommand{\arraystretch}{1.5}
\small
\centering
\scalebox{0.8}{
\begin{tabular}{p{1.2\textwidth}}
\toprule
\textbf{Easy task 1 of Example 1:} 
\\
Given an arithmetic sequence $\{a_{n}\}$ where $a_{1} = 1$ and the common difference $d = 2$. Find the explicit formula for the $n$-th term $a_{n}$ of the sequence.
\\
\midrule
\textbf{Solution:}
\\
The general formula for an arithmetic sequence is given by:
\[
a_{n} = a_{1} + (n-1)d
\]
Substituting the known values, we get:
\[
a_{n} = 1 + (n-1) \cdot 2
\]
Simplifying:
\[
a_{n} = 1 + 2n - 2 = 2n - 1
\]
Thus, the explicit formula for the $n$-th term of the sequence $\{a_{n}\}$ is:
\[
\boxed{a_{n} = 2n - 1}
\]
\\
\bottomrule
\end{tabular}
}
\caption{Decomposition - Example 1: Easy task 1.}
\label{tab:demo_decompose_eg1_sub1}
\end{table*}

\begin{table*}[!ht]
\renewcommand{\arraystretch}{1.5}
\small
\centering
\scalebox{0.8}{
\begin{tabular}{p{1.2\textwidth}}
\toprule
\textbf{Easy task 2 of Example 1:} 
\\
Given an arithmetic sequence $\{a_{n}\}$ where $a_{1} = 1$ and the common difference $d = 2$. Find the explicit formula for the $n$-th term $a_{n}$ of the sequence.
\\
\midrule
\textbf{Solution:}
\\
    To find the 5th term \( b_{5} \), we use the relationship between the sum of the first \( n \) terms and the \( n \)-th term: \\
    \[
    b_{n} = S_{n} - S_{n-1}
    \] 
    Calculating \( S_{5} \) and \( S_{4} \): \\
    \[
    S_{5} = 2^{5+1} - 2 = 2^{6} - 2 = 64 - 2 = 62
    \] 
    \[
    S_{4} = 2^{4+1} - 2 = 2^{5} - 2 = 32 - 2 = 30
    \] 
    Now, substituting into the equation for \( b_{5} \): \\
    \[
    b_{5} = S_{5} - S_{4} = 62 - 30 = 32
    \] 
    Therefore, the 5th term of the sequence \( \{b_{n}\} \) is: \\
    \[
    \boxed{b_{5} = 32}
    \]
\\
\bottomrule
\end{tabular}
}
\caption{Decomposition - Example 1: Easy task 2.}
\label{tab:demo_decompose_eg1_sub2}
\end{table*}

\begin{table*}[!ht]
\renewcommand{\arraystretch}{1.5}
\small
\centering
\scalebox{0.8}{
\begin{tabular}{p{1.2\textwidth}}
\toprule
\textbf{Hard task (full task) of Example 2:} \\
Find the range of real numbers $m$ for which the quadratic equation $x^2 + (m-1)x + 1 = 0$ has solutions in the interval $[0,2]$.\\
\midrule
\textbf{Solution:}\\
For the quadratic equation $x^2 + (m-1)x + 1 = 0$ to have solutions in the interval $[0,2]$, it must satisfy certain conditions related to its discriminant and the values of $x$ at the endpoints of the interval. \\

First, let's consider the discriminant of the equation, $\Delta = (m-1)^2 - 4*1*1 = m^2 - 2m - 3$. \\
For the equation to have real solutions, the discriminant must be non-negative, $\Delta \geq 0$. \\
This gives us $m^2 - 2m - 3 \geq 0$, which simplifies to $(m-3)(m+1) \geq 0$. This inequality holds for $m \leq -1$ or $m \geq 3$. \\
However, we also need to ensure that the solutions are within the interval $[0,2]$. By evaluating the quadratic equation at the endpoints of the interval, we get: \\

1. For $x=0$: $0^2 + (m-1)0 + 1 = 1$, which is always positive. \\
2. For $x=2$: $2^2 + (m-1)2 + 1 = 4 + 2m - 2 + 1 = 2m + 3$, which needs to be non-negative for the solutions to lie within the interval. This gives us $2m + 3 \geq 0$, simplifying to $m \geq -\frac{3}{2}$. \\

Combining these conditions, we find that the range of $m$ for which the quadratic equation has solutions in the interval $[0,2]$ is $\boxed{(-\infty, -1)}$.\\
\bottomrule
\end{tabular}
}
\caption{Decomposition - Example 2: Hard task.}
\label{tab:demo_decompose_eg2_full}
\end{table*}

\begin{table*}[!ht]
\renewcommand{\arraystretch}{1.5}
\small
\centering
\scalebox{0.8}{
\begin{tabular}{p{1.2\textwidth}}
\toprule
\textbf{Easy task 1 of Example 2:} \\
Determine the values of \( m \) for which the discriminant of the quadratic equation \( x^2 + (m-1)x + 1 = 0 \) is non-negative.\\
\midrule
\textbf{Solution:} \\
The discriminant \( \Delta \) of the quadratic equation \( x^2 + (m-1)x + 1 = 0 \) is given by:\\
\[
\Delta = (m-1)^2 - 4 \cdot 1 \cdot 1 = m^2 - 2m - 3.
\]

To find the values of \( m \) for which the discriminant is non-negative, we need to solve the inequality:

\[
m^2 - 2m - 3 \geq 0.
\]

Factoring the quadratic expression yields:

\[
(m-3)(m+1) \geq 0.
\]

To solve this inequality, we can find the critical points where the expression equals zero:

1. \( m - 3 = 0 \) gives \( m = 3 \).
2. \( m + 1 = 0 \) gives \( m = -1 \).

Now we analyze the sign of the product \( (m-3)(m+1) \):

- For \( m < -1 \), both factors are negative, hence the product is positive. \\
- For \( -1 < m < 3 \), one factor is negative and the other is positive, hence the product is negative.\\
- For \( m > 3 \), both factors are positive, hence the product is positive. \\

Thus, the solution to the inequality is:

\[
m \leq -1 \quad \text{or} \quad m \geq 3.
\]

The values of \( m \) for which the discriminant is non-negative are therefore:

\(\boxed{(-\infty, -1) \cup (3, \infty)}\).\\
\bottomrule
\end{tabular}
}
\caption{Decomposition - Example 2: Easy task 1.}
\label{tab:demo_decompose_eg2_sub1}
\end{table*}

\begin{table*}[!ht]
\renewcommand{\arraystretch}{1.5}
\small
\centering
\scalebox{0.8}{
\begin{tabular}{p{1.2\textwidth}}
\toprule
\textbf{Easy task 2 of Example 2:} \\
Find the values of \( m \) such that the value of the quadratic equation \( x^2 + (m-1)x + 1 = 0 \) evaluated at \( x = 2 \) is non-negative.\\
\midrule
\textbf{Solution:} \\
Evaluating the quadratic equation at \( x = 2 \):

\[
2^2 + (m-1) \cdot 2 + 1 = 4 + 2(m-1) + 1 = 2m + 3.
\]

We want this expression to be non-negative:

\[
2m + 3 \geq 0.
\]

Solving for \( m \):

\[
2m \geq -3 \quad \Rightarrow \quad m \geq -\frac{3}{2}.
\]

Thus, the values of \( m \) for which the quadratic evaluated at \( x = 2 \) is non-negative are:

\(\boxed{[-\frac{3}{2}, \infty)}\).\\
\bottomrule
\end{tabular}
}
\caption{Decomposition - Example 2: Easy task 2.}
\label{tab:demo_decompose_eg2_sub2}
\end{table*}

\begin{table*}[!ht]
\renewcommand{\arraystretch}{1.5}
\small
\centering
\scalebox{0.8}{
\begin{tabular}{p{1.2\textwidth}}
\toprule
\textbf{Unqualified easy task - Type 1 (Ill-defined) - Example 1} 
\\
\midrule
\textbf{Task: } 
\\
Let \( b_n = L^{(2)} \left( \frac{25}{n} \right) \), where \( L^{(2)}(x) \) denotes the second iterate of \( L(x) \). Determine \( b_n \) as \( n \) approaches infinity
\\
\textbf{Solution:}
\\
First, we need to compute \( L^{(2)}(x) \):

\[
L^{(1)}(x) = \textcolor{red}{\boxed{L(x) = x - \frac{x^2}{2}}}.
\]

\textcolor{red}{Comment: The definition of $L(x)$ does not exist in the problem.}

Now we apply \( L \) again to \( L(x) \):

\[
\textcolor{red}{\boxed{L^{(2)}(x) = L(L(x))}} = L \left( x - \frac{x^2}{2} \right).
\]

\textcolor{red}{Comment: It's NOT mentioned in the problem what's ``iterate of $L(x)$'' and how to calculate it.}

Substituting \( x - \frac{x^2}{2} \) into \( L \):

\[
L \left( x - \frac{x^2}{2} \right) = \left( x - \frac{x^2}{2} \right) - \frac{1}{2} \left( x - \frac{x^2}{2} \right)^2.
\]

Calculating \( \left( x - \frac{x^2}{2} \right)^2 \):

\[
\left( x - \frac{x^2}{2} \right)^2 = x^2 - x^3 + \frac{x^4}{4}.
\]

Now, substituting back:

\[
L^{(2)}(x) = x - \frac{x^2}{2} - \frac{1}{2} \left( x^2 - x^3 + \frac{x^4}{4} \right),
\]

which simplifies to:

\[
L^{(2)}(x) = x - \frac{x^2}{2} - \frac{1}{2}x^2 + \frac{1}{2}x^3 - \frac{1}{8}x^4.
\]

Combining terms gives:

\[
L^{(2)}(x) = x - \frac{3}{4}x^2 + \frac{1}{2}x^3 - \frac{1}{8}x^4.
\]

Now, we evaluate \( b_n = L^{(2)} \left( \frac{25}{n} \right) \):

\[
b_n = L^{(2)} \left( \frac{25}{n} \right) = \frac{25}{n} - \frac{3}{4} \left( \frac{25}{n} \right)^2 + \frac{1}{2} \left( \frac{25}{n} \right)^3 - \frac{1}{8} \left( \frac{25}{n} \right)^4.
\]

Calculating each term:

1. \( \frac{25}{n} \)
2. \( -\frac{3}{4} \cdot \frac{625}{n^2} = -\frac{1875}{4n^2} \)
3. \( \frac{1}{2} \cdot \frac{15625}{n^3} = \frac{7812.5}{n^3} \)
4. \( -\frac{1}{8} \cdot \frac{390625}{n^4} = -\frac{48828.125}{n^4} \)

As \( n \) approaches infinity, the higher-order terms vanish. Therefore, the leading term dominates:

\[
b_n \approx \frac{25}{n}.
\]

Thus, as \( n \) approaches infinity,

\[
b_n \rightarrow 0.
\]

The value that \( b_n \) approaches is \( \boxed{0} \).
\\
\bottomrule
\end{tabular}
}
\caption{Unqualified easy task - Type 1 (Ill-defined) - Example 1.}
\label{tab:demo_filtering1_eg1}
\end{table*}

\begin{table*}[!ht]
\renewcommand{\arraystretch}{1.5}
\small
\centering
\scalebox{0.8}{
\begin{tabular}{p{1.2\textwidth}}
\toprule
\textbf{Unqualified easy task - Type 1 (Ill-defined) - Example 2} 
\\
\midrule
\textbf{Task: } 
\\
Given triangle \(AMC\) with \(AM = kx\), \(AC = 3\), and \(MC = \frac{x}{2}\), find the relationship between \(k\), \(x\), and \(\cos(180^\circ - \theta)\).
\\
\textbf{Solution:}
\\
Using the Law of Cosines for triangle \(AMC\):

\[
\textcolor{red}{
\boxed{
\left(\frac{x}{2}\right)^2 + 3^2 - 2 \cdot \frac{x}{2} \cdot 3 \cdot \cos(180^\circ - \theta) = (kx)^2}
}
\]
\textcolor{red}{Comment: Which angle of the triangle is $\cos(180^\circ - \theta)$ is not defined in the problem (we can infer from the formula above it's angle ACM).}

Substituting the cosine identity \(\cos(180^\circ - \theta) = -\cos\theta\):

\[
\frac{x^2}{4} + 9 + 3x\cos\theta = k^2x^2
\]

Rearranging gives:

\[
3x\cos\theta = k^2x^2 - 9 - \frac{x^2}{4}
\]

Thus, isolating \(\cos\theta\):

\[
\cos\theta = \frac{k^2x^2 - 9 - \frac{x^2}{4}}{3x}
\]

This provides a relationship between \(k\), \(x\), and \(\cos\theta\) in triangle \(AMC\).

Conclusion:

\[
\cos\theta = \frac{k^2x^2 - 9 - \frac{x^2}{4}}{3x}
\].
\\
\bottomrule
\end{tabular}
}
\caption{Unqualified easy task - Type 1 (Ill-defined) - Example 2.}
\label{tab:demo_filtering1_eg2}
\end{table*}

\begin{table*}[!ht]
\renewcommand{\arraystretch}{1.5}
\small
\centering
\scalebox{0.8}{
\begin{tabular}{p{1.2\textwidth}}
\toprule
\textbf{Unqualified easy task - Type 2 (Overly Simple) - Example 1} 
\\
\midrule
\textbf{Task: } 
\\
Given a quadratic function \( f(x) = ax^2 + bx + c \) with \( a > 0 \) and \( b > 0 \), find the expression for the value of \( f(1) \) in terms of \( a, b, c \)
\\
\textbf{Solution:}
\\
To find the expression for \( f(1) \), we start with the quadratic function defined as \( f(x) = ax^2 + bx + c \).

1. Substitute \( x = 1 \) into the function:
   \[
   f(1) = a(1^2) + b(1) + c = a + b + c.
   \]

2. Therefore, the expression for \( f(1) \) in terms of \( a, b, \) and \( c \) is:
   \[
   f(1) = a + b + c.
   \]

Thus, the value of \( f(1) \) is expressed as \( a + b + c \).

\textcolor{red}{Comment: the solution has only one step, which indicates this decomposed easy task is overly simple.}
\\
\bottomrule
\end{tabular}
}
\caption{Unqualified easy task - Type 2 (Overly Simple) - Example 1.}
\label{tab:demo_filtering2_eg1}
\end{table*}

\begin{table*}[!ht]
\renewcommand{\arraystretch}{1.5}
\small
\centering
\scalebox{0.8}{
\begin{tabular}{p{1.2\textwidth}}
\toprule
\textbf{Unqualified easy task - Type 2 (Overly Simple) - Example 2} \\
\midrule
\textbf{Task: } 
\\
Find the expression for the first derivative of the function \( f_0(x) = \frac{1}{x} \)
\\
\textbf{Solution:}
\\
To find the first derivative \( f_1(x) \) of the function \( f_0(x) = \frac{1}{x} \), we can use the power rule for differentiation.

1. Rewrite the function: \( f_0(x) = x^{-1} \).
2. Apply the power rule: The derivative of \( x^n \) is \( n \cdot x^{n-1} \). Therefore, for \( f_0(x) \):
   \[
   f_1(x) = \frac{d}{dx}(x^{-1}) = -1 \cdot x^{-2} = -\frac{1}{x^2}.
   \]

Thus, the expression for the first derivative is:
\[
\boxed{f_1(x) = -\frac{1}{x^2}}.
\]
\textcolor{red}{Comment: the solution has only two steps, which indicates this decomposed easy task is overly simple.}\\
\bottomrule
\end{tabular}
}
\caption{Unqualified easy task - Type 2 (Overly Simple) - Example 2.}
\label{tab:demo_filtering2_eg2}
\end{table*}

\begin{table*}[!ht]
\renewcommand{\arraystretch}{1.5}
\small
\centering
\scalebox{0.8}{
\begin{tabular}{p{1.2\textwidth}}
\toprule
\textbf{Unqualified easy task - Type 3 (too similar to the original task) - Example 1} 
\\
\midrule

\textbf{Hard Task (full task): } 

Given the function \( h(x) = 3\ln(x) - x^3 + 1 \) defined on the interval \(\left[\frac{1}{e}, e\right]\), find the critical points of \( h(x) \) and determine whether \( h(x) \) has a maximum or minimum at these points.\\
\midrule

\textbf{Easy task 2 (unqualified, too similar): } 

Evaluate the values of the function \( h(x) = 3\ln(x) - x^3 + 1 \) at the endpoints of the interval \( \left[\frac{1}{e}, e\right] \) and determine the minimum and maximum values of \( h(x) \) over this interval.

\textcolor{red}{Comment: decomposed easy task is almost the same as hard task since they solve for the same things.}\\
\bottomrule
\end{tabular}
}
\caption{Unqualified easy task - Type 3 (too similar to the original task) - Example 1.}
\label{tab:demo_filtering3_eg1}
\end{table*}

\begin{table*}[!ht]
\renewcommand{\arraystretch}{1.5}
\small
\centering
\scalebox{0.8}{
\begin{tabular}{p{1.2\textwidth}}
\toprule
\textbf{Unqualified easy task - Type 3 (too similar to the original task) - Example 2} \\
\midrule
\textbf{Hard Task (full task): } \\
A conical hill stands with its base on level ground and has a total height of 10,000 feet. Given that the top one-tenth of the volume of the hill is above the cloud layer, what is the height of the cloud layer above the ground in feet?\\
\midrule
\textbf{Easy task 2 (unqualified, too similar): } \\
Given a conical hill with a total height of 10,000 feet, if the portion of the cone above the cloud layer represents one-tenth of the volume, what is the height of that portion above the cloud layer in feet?\\
\textcolor{red}{Comment: the decomposed easy task has the same difficulty of hard task, even though they solve for different things, their solutions are comparable.}\\
\bottomrule
\end{tabular}
}
\caption{Unqualified easy task - Type 3 (too similar to the original task) - Example 2.}
\label{tab:demo_filtering3_eg2}
\end{table*}

\section{Case Study for Step-Wise Error Rate}
\label{sec:case_study}
As discussed in \S\ref{subsec:5.2}, models consistently perform better when trained with supervision from hard full tasks. They also demonstrate robustness against varying outcome error rates, even when these rates differ significantly, such as between 0\% and 100\%. This leads us to explore how other factors, such as the severity of erroneous solutions, impact supervision quality and downstream performance. To measure severity, we consider the metric, step-wise error rate. Defined in \S\ref{sec:6}, the step-wise error rate represents the proportion of erroneous steps relative to the total number of steps. Importantly, we exclude steps that, while based on incorrect numerical values from prior errors, exhibit correct logic and reasoning. In this context, we focus solely on the logical correctness of each step.

We present case studies of sampled solutions from five different teacher models with varying reasoning capabilities: Command-R-2024-03, Command-R-2024-08, GPT-3.5-turbo, Llama-3-70B-Instruct, and GPT-4o-mini. Each case study includes annotations of the correctness of individual steps, along with explanations for any errors identified and step-wise error rate calculation.

\noindent \textbf{Example 1 for Step-Wise Error Rate.}
Table 
\ref{tab:case_study_1_model1}, \ref{tab:case_study_1_model2}, \ref{tab:case_study_1_model3},  
\ref{tab:case_study_1_model4},
\ref{tab:case_study_1_model5} 
shows the first example of step-wise examination for the solutions to the same problem sampled from five supervision teacher models.

\noindent \textbf{Example 2 for Step-Wise Error Rate.}
Table 
\ref{tab:case_study_2_model1}, \ref{tab:case_study_2_model2}, \ref{tab:case_study_2_model3},  
\ref{tab:case_study_2_model4},
\ref{tab:case_study_2_model5} 
shows the second example of step-wise examination for the solutions to the same problem sampled from five supervision teacher models.

\noindent \textbf{Example 3 for Step-Wise Error Rate.}
Table 
\ref{tab:case_study_3_model1}, \ref{tab:case_study_3_model2}, \ref{tab:case_study_3_model3},  
\ref{tab:case_study_3_model4},
\ref{tab:case_study_3_model5} 
shows the third example of step-wise examination for the solutions to the same problem sampled from five supervision teacher models.





\section{Prompt Templates}
\label{sec:prompt_templates}

We utilize multiple advanced LLMs in both supervision synthesis and quality controls. Here we show all prompt templates used in our experiments. 

In Stage 1 of our supervision synthesis pipeline, we utilize GPT-4o-mini to decompose hard full tasks into 2 or 3 easy subtasks, as outlined in Table \ref{tab:prompt_s1_decompose}. Following this decomposition, we apply filtering process using GPT-4o-mini to eliminate unqualified subtasks, with the prompt in Table \ref{tab:prompt_s1_filtering}.

In Stage 2, we sample three solutions for each full hard full task and easy subtask, as shown in Table \ref{tab:prompt_s2_sampling}, and verify their correctness using the prompt as shown in Table \ref{tab:prompt_s2_check_sample}. When constructing pairs of correct and incorrect solutions, as discussed in \S\ref{subsec:4.3}, we employ style transfer to the ground-truth solution when all sampled solutions are incorrect, with the prompt for style transfer provided in Table \ref{tab:prompt_s2_style_transfer}, then we take the transferred version as the correct solution in the pair.

In \S\ref{sec:6}, we study the severity of erroneous answers and utilize five different teacher models to sample solutions of hard full tasks. Here we use the same prompt templates as the data synthesis pipeline in \S\ref{sec:4} for both sampling (Table \ref{tab:prompt_s2_sampling}) and checking solutions (Table \ref{tab:prompt_s2_check_sample}). 

In \S\ref{sec:7}, we explore the strategy of combining hard full tasks and easy subtasks, with two baselines of 1) training with doubled epochs and 2) training with rephrased tasks merged with original data. The prompt template used for rephrasing in the second baseline is shown in Table \ref{tab:prompt_rephrase}.

\begin{table*}[!ht]
\renewcommand{\arraystretch}{1.5}
\small
\centering
\scalebox{0.8}{
\begin{tabular}{p{1.2\textwidth}}
\toprule
\textbf{Sampling multiple solutions and constructing pairs - Example 1} \\
\midrule
\textbf{Task (this example is from hard tasks): }\\
Two circles intersect at points A(1, 3) and B(m, n), and both of their centers lie on the line $x - y - 2 = 0$. What is the value of $m + n$?\\
\midrule
\textbf{Ground Truth: }\\
Given that the circles intersect at points A(1, 3) and B(m, n) and that the centers of both circles lie on the line $x - y - 2 = 0$, we can derive the slope of line segment AB as follows:
\begin{center}
\textbf{...... (process partly omitted here)}\\
\end{center}
Substituting $n = -1$ back into equation (3) gives us:

$$ m - (-1) = 6 $$
$$ m + 1 = 6 $$
$$ m = 5 $$

Thus, we have found the coordinates $B(m, n)$ to be $B(5, -1)$. Now we calculate $m + n$:

$$ m + n = 5 + (-1) $$

$$ \boxed{m + n = 4} $$\\

\midrule
\textbf{Sampled solution 1: \textcolor{green}{(Correct)}}\\
To solve the problem, we start by noting that the centers of both circles lie on the line given by the equation \(x - y - 2 = 0\). We can rewrite this line in slope-intercept form as \(y = x - 2\). 

Let the centers of the two circles be denoted as \(C_1(x_1, y_1)\) and \(C_2(x_2, y_2)\). Since both centers lie on the line, we have:
\begin{center}
\textbf{...... (process partly omitted here)}\\
\end{center}
Rearranging terms, we find:

\[
n + m = 4
\]

Thus, the value of \(m + n\) is:

\[
\boxed{4}
\]\\
\midrule
\textbf{Sampled solution 2: \textcolor{red}{(Incorrect)}}\\
Let the centers of the two circles be \( C_1(h_1, k_1) \) and \( C_2(h_2, k_2) \). Since both centers lie on the line \( x - y - 2 = 0 \), we can express the coordinates of the centers in terms of a single variable. The equation can be rewritten as \( y = x - 2 \). Therefore, we can say:

- For \( C_1(h_1, k_1) \): \( k_1 = h_1 - 2 \)
- For \( C_2(h_2, k_2) \): \( k_2 = h_2 - 2 \)
\begin{center}
\textbf{...... (process partly omitted here)}\\
\end{center}
Thus, 

\[
m + n = 2 + 0 = 2.
\]

Therefore, the value of \( m + n \) is 

\[
\boxed{2}.
\] \\

\midrule
\textbf{Sampled solution 3: \textcolor{red}{(Incorrect)}}\\ 
Let the centers of the two circles be \( C_1(h_1, k_1) \) and \( C_2(h_2, k_2) \). Since both centers lie on the line \( x - y - 2 = 0 \), we can express their coordinates in terms of a single variable \( t \):
\begin{center}
\textbf{...... (process partly omitted here)}\\
\end{center}
\[
n = 5 - 2 = 3.
\]

Thus:

\[
m + n = 5 + 3 = 8.
\]

Therefore, the final result is:

\[
\boxed{8}.
\]  \\

\midrule

\textbf{Comment:} There are both correct and incorrect solutions in the samples, thus we choose sample 1 as correct solution in the pair and choose sample 3 (randomly chosed) as incorrect solution in the pair.\\
\bottomrule
\end{tabular}
}
\caption{Sampling multiple solutions and constructing pairs - Example 1.}
\label{tab:demo_multi_sampling_eg1}
\end{table*}

\begin{table*}[!ht]
\renewcommand{\arraystretch}{1.5}
\small
\centering
\scalebox{0.8}{
\begin{tabular}{p{1.2\textwidth}}
\toprule
\textbf{Sampling multiple solutions and constructing pairs - Example 2} \\
\midrule
\textbf{Task (this example is from easy tasks): }\\
Solve the inequality \( \frac{1-x}{x-5} \geq 1 \) with respect to \( x \) \\

\midrule

\textbf{Ground Truth: }\\
To solve the inequality \( \frac{1-x}{x-5} \geq 1 \), we follow these steps: \\
1. **Given Inequality**: \( \frac{1-x}{x-5} \geq 1 \). \\
2. **Transforming the inequality**: Rewrite as \( \frac{1-x}{x-5} - 1 \geq 0 \), which gives \( \frac{1-x - (x-5)}{x-5} \geq 0 \). \\
\begin{center}
\textbf{...... (process partly omitted here)}\\
\end{center}
7. **Conclusion**: The solution is in the interval \( [3, 5) \), excluding \( x = 5 \). \\
Thus, the solution set is: 
$$\boxed{\{ x \, | \, 3 \leqslant x < 5 \}}$$ \\

\midrule

\textbf{Sampled solution 1: \textcolor{red}{(Inorrect)}}\\
To solve the inequality \( \frac{1-x}{x-5} \geq 1 \), find a common denominator: \\
\( \frac{1-x}{x-5} \geq 1 \) \\
Combine fractions: 
\begin{center}
\textbf{...... (process partly omitted here)}\\
\end{center}
3. \( x > 5 \): Choose \( x = 6 \) yields \( < 0 \). Not a solution. \\
Therefore, solution is \( \boxed{3 < x < 5}\) \\

\midrule

\textbf{Sampled solution 2: \textcolor{red}{(Incorrect)}}\\
To solve the inequality \( \frac{1-x}{x-5} \geq 1 \), find a common denominator: \\
Multiplying both sides by \( (x-5) \): \\
\( 1-x \geq x-5 \) \\
Simplifying gives: \\
\( 1 + 5 \geq 2x \) \\
Thus, solution is: \\
\( \boxed{x \leq 3} \)\\
\midrule

\textbf{Sampled solution 3: \textcolor{red}{(Incorrect)}}\\ 
To solve the inequality \( \frac{1-x}{x-5} \geq 1 \), find common denominator: \\
Multiplying both sides by \( x-5 \): \\
\( 1-x \geq x-5 \) \\
Simplifying yields: \\
\( 1 \geq 0 \), true for all \( x \). \\
Thus, solution is: \\
\( \boxed{x \in \mathbb{R}} \)\\
\midrule

\textbf{Ground truth with style transferred: \textcolor{green}{(Reference)}}\\ 
To solve the inequality \( \frac{1-x}{x-5} \geq 1 \), we start by reformulating it: 
1. Start with the inequality: 
   \[ 
   \frac{1-x}{x-5} \geq 1.
   \] 
\begin{center}
\textbf{...... (process partly omitted here)}\\
\end{center}
6. We will now test the sign of \( (2x-6)(x-5) \) across the intervals created by these points: 
   - For \( x < 3 \): both factors \( 2x-6 < 0 \) and \( x-5 < 0 \) yield a positive product. 
   - For \( 3 < x < 5 \): \( 2x-6 < 0 \) and \( x-5 < 0 \) yield a positive product. 
   - For \( x > 5 \): both factors \( 2x-6 > 0 \) and \( x-5 > 0 \) yield a positive product. 
   
7. Evaluating the overall expression, we find it is non-positive exactly in the interval \( [3, 5) \), since at \( x=3 \) it equals zero, and we exclude \( x=5 \) due to the denominator. 
Thus, the solution set is: 
$
\boxed{[3, 5)}.
$ \\
\midrule

\textbf{Comment:} All the samples are incorrect, thus we convert the style of ground truth to the style of sampled solution and take transferred ground truth as correct solution and choose sample 2 (randomly chosed) as incorrect solution in the pair.\\
\bottomrule
\end{tabular}
}
\caption{Sampling multiple solutions and constructing pairs - Example 2.}
\label{tab:demo_multi_sampling_eg2}
\end{table*}

\begin{table*}[!ht]
\renewcommand{\arraystretch}{1.5}
\small
\centering
\scalebox{0.8}{
\begin{tabular}{p{1.2\textwidth}}
\toprule
\textbf{Sampling multiple solutions and constructing pairs - Example 3} \\
\midrule
\textbf{Task (this example is from easy tasks): }\\
Given an arithmetic sequence $\{a_n\}$ where $a_1 = 1$, $a_2 = b_2$, and $a_4 + 2 = b_3$, if the common difference of the sequence is $d$, find the value of $d$ if it is known that $a_4 = 7$. \\
\midrule
\textbf{Ground Truth: }\\
Since we know that $a_1 = 1$ and the common difference is $d$, we can express the terms of the arithmetic sequence as follows: 
- $a_2 = a_1 + d = 1 + d$ 
- $a_4 = a_1 + 3d = 1 + 3d$ 
Given $a_4 = 7$, we can set up the equation: 
$$
1 + 3d = 7
$$ 

Subtracting 1 from both sides gives: 
$$
3d = 6
$$ 

Dividing both sides by 3 yields: 
$$
d = 2
$$ \\

Thus, the value of the common difference $d$ is $\boxed{2}$ \\

\midrule
\textbf{Sampled solution 1: \textcolor{green}{(Correct)}}\\

The general formula for an arithmetic sequence is $a_n = a_1 + (n-1)d$, where $a_n$ is the nth term, $a_1$ is the first term, $d$ is the common difference, and $n$ is the term number. 

Given $a_1 = 1$, $a_2 = b_2$, and $a_4 = 7$, we can write the following equations: 
1. $a_2 = a_1 + d = 1 + d = b_2$ 
2. $a_4 = a_1 + 3d = 1 + 3d = 7$ .

From equation 2, we can solve for $d$: 
$1 + 3d = 7$, $3d = 6$, $d = 2$ \\

Therefore, the common difference of the arithmetic sequence is $\boxed{2}$ \\

\midrule
\textbf{Sampled solution 2: \textcolor{green}{(Correct)}}\\
Since the sequence is arithmetic, we know that $a_n = a_1 + (n-1)d$. 
We are given that $a_4 = 7$, so we can substitute $n=4$ into the formula: 
$a_4 = a_1 + (4-1)d = 1 + 3d = 7$. \\

Solving for $d$: 
$3d = 6$, $d = 2$ \\

Therefore, the common difference of the sequence is $\boxed{2}$ \\

\midrule
\textbf{Sampled solution 3: \textcolor{green}{(Correct)}}\\ 
The general formula for an arithmetic sequence is $a_n = a_1 + (n-1)d$, where $a_n$ is the nth term, $a_1$ is the first term, and $d$ is the common difference. 

Given that $a_1 = 1$ and $a_4 = 7$, we can write: 
$a_4 = a_1 + 3d = 7$ 

Substitute $a_1 = 1$ into the equation: \\
$1 + 3d = 7$ 
Solving for $d$: 
$3d = 6$, $d = 2$ \\

Therefore, the common difference $d$ is $\boxed{2}$ \\

\midrule

\textbf{Comment:} All the sampled solutions are correct, in this case we skip and discard this example.\\
\bottomrule
\end{tabular}
}
\caption{Sampling multiple solutions and constructing pairs - Example 3.}
\label{tab:demo_multi_sampling_eg3}
\end{table*}

\begin{table*}[!ht]
\renewcommand{\arraystretch}{1.5}
\small
\centering
\scalebox{0.8}{
\begin{tabular}{p{1.2\textwidth}}
\toprule
 Example 1 for step-wise error ratio - ``Trigonometric identity transformations'': Supervision sampled by \textbf{Command-R-2024-03} \\
\midrule
\textbf{Task: } \\
Given $\sin\left(\alpha - \frac{\pi}{6}\right) = \frac{1}{3}$, then $\sin\left(2\alpha - \frac{\pi}{6}\right) + \cos 2\alpha = \left( \ \right)$ \\
A: $-\frac{2}{3}$, 
B: $\frac{2}{3}$, 
C: $-\frac{7}{9}$, 
D: $\frac{7}{9}$\\

\midrule

\textbf{Solution:}\\
We can write: 
$\sin\left(2\alpha - \frac{\pi}{6}\right) = \sin\left(\alpha - \frac{\pi}{3} + \alpha\right)$ \textcolor{red}{(Step 1: wrong, it should be ``$\sin\left(\alpha - \frac{\pi}{6} + \alpha\right)$'')}\\

$= \sin\left(\alpha - \frac{\pi}{3}\right) \cdot \cos\alpha + \cos\left(\alpha - \frac{\pi}{3}\right) \cdot \sin\alpha$ \textcolor{green}{(Step 2: correct)}\\
$= \sin\alpha \cdot \cos\alpha - \frac{1}{3} \cdot \sin\alpha $ \textcolor{red}{(Step 3: wrong, it incorrectly take $\sin\left(\alpha - \frac{\pi}{3}\right)$ as $\sin\alpha$, and take $\cos\left(\alpha - \frac{\pi}{3}\right)$ as $-\frac{1}{3}$)}\\
$= \boxed{\frac{2}{3}}$. \textcolor{red}{(Step 4: wrong, the value of above equation can not be calculated since $\sin\alpha$ and $\cos\alpha$ are not known.)}\\

Therefore, the final answer is B. \\
\midrule

\textbf{Comment: }\\
There are 4 steps totally while 3 of them are wrong, step-wise error ratio is \textbf{75.0\%}.
\\
\bottomrule
\end{tabular}
}
\caption{ Example 1 for step-wise error ratio - ``Trigonometric identity transformations'': Supervision sampled by \textbf{Command-R-2024-03}.}
\label{tab:case_study_1_model1}
\end{table*}

\begin{table*}[!ht]
\renewcommand{\arraystretch}{1.5}
\small
\centering
\scalebox{0.8}{
\begin{tabular}{p{1.2\textwidth}}
\toprule
 Example 1 for step-wise error ratio - ``Trigonometric identity transformations'': Supervision sampled by \textbf{Command-R-2024-08} \\
\midrule
\textbf{Task: } \\
Given $\sin\left(\alpha - \frac{\pi}{6}\right) = \frac{1}{3}$, then $\sin\left(2\alpha - \frac{\pi}{6}\right) + \cos 2\alpha = \left( \ \right)$ \\
A: $-\frac{2}{3}$, 
B: $\frac{2}{3}$, 
C: $-\frac{7}{9}$, 
D: $\frac{7}{9}$\\

\midrule

\textbf{Solution:}\\
Let's think step by step to solve this trigonometry problem. \\

We are given that $\sin(\alpha - \frac{\pi}{6}) = \frac{1}{3}$. \textcolor{green}{(Step 1: correct)}\\

We can use the double-angle formula for sine to rewrite $\sin(2\alpha - \frac{\pi}{6})$ as: \\
$\sin(2\alpha - \frac{\pi}{6}) = \sin(2\alpha)\cos(\frac{\pi}{6}) - \cos(2\alpha)\sin(\frac{\pi}{6})$ \textcolor{green}{(Step 2: correct)}\\

Now, we can substitute $\sin(\alpha - \frac{\pi}{6}) = \frac{1}{3}$: 

$\sin(\alpha)\cos(\alpha) = \frac{1}{3}\cos(\frac{\pi}{6}) = \frac{1}{3}\cdot\frac{1}{2} = \frac{1}{6} $ \textcolor{red}{(Step 3: wrong, $\cos(\frac{\pi}{6})$ is $\frac{\sqrt{3}}{2}$, not $\frac{1}{2}$})\\

$\cos^2(\alpha) = 1 - \sin^2(\alpha) = 1 - \frac{1}{9} = \frac{8}{9} $  \textcolor{green}{(Step 4: correct)}\\

$\sin^2(\alpha) = \frac{1}{9}$ \textcolor{green}{(Step 5: correct)}\\

Substituting these values back into our equation: 

$\sin(2\alpha - \frac{\pi}{6}) = \frac{1}{6} - \frac{1}{2}\cdot\frac{8}{9} + \frac{1}{2}\cdot\frac{1}{9}$ \textcolor{green}{(Step 6: correct, this step substitutes values into equation of Step 2, $\sin(2\alpha)\cos(\frac{\pi}{6}) =2\sin(\alpha)\cos(\alpha)\cos(\frac{\pi}{6})$, since Step 3 mistakenly take $\cos(\frac{\pi}{6})$ as $\cos(\frac{1}{2})$, so here $\sin(2\alpha)\cos(\frac{\pi}{6}) = 2\sin(\alpha)\cos(\alpha)\frac{1}{2}=\sin(\alpha)\cos(\alpha)=\frac{1}{6}$. The error is due to previous wrong numerical values, not this step.)}\\

$= \frac{1}{6} - \frac{4}{9} + \frac{1}{18}$ 
\textcolor{green}{(Step 7: correct)}\\
$= \frac{1}{6} - \frac{11}{18}$ 
\textcolor{red}{(Step 8: wrong, $- \frac{4}{9} + \frac{1}{18} = -\frac{7}{18}$, not $-\frac{11}{18}$)}\\
$= \frac{3 - 11}{18}$ 
\textcolor{green}{(Step 9: correct, Step 9 carries wrong values from Step 8 but its calculation is correct, thus this step is not included in ``erroneous steps'')}\\
$= -\frac{8}{18}$ 
\textcolor{green}{(Step 10: correct)}\\
$= -\frac{2}{3}$ 
\textcolor{red}{(Step 11: wrong, $-\frac{8}{18} = -\frac{4}{9}$, not $-\frac{2}{3}$)}\\

So, the value of $\sin(2\alpha - \frac{\pi}{6}) + \cos(2\alpha)$ is $-\frac{2}{3}$. 

Therefore, the correct answer is **A: $-\frac{2}{3}$**. \\ 

\midrule

\textbf{Comment: }\\
There are 11 steps totally while 3 of them are wrong, step-wise error ratio is \textbf{27.3\%}.
\\

\bottomrule
\end{tabular}
}
\caption{Example 1 for step-wise error ratio - ``Trigonometric identity transformations'': Supervision sampled by \textbf{Command-R-2024-08}.}
\label{tab:case_study_1_model2}
\end{table*}

\begin{table*}[!ht]
\renewcommand{\arraystretch}{1.5}
\small
\centering
\scalebox{0.8}{
\begin{tabular}{p{1.2\textwidth}}
\toprule
 Example 1 for step-wise error ratio - ``Trigonometric identity transformations'': Supervision sampled by \textbf{GPT-3.5-turbo} \\
\midrule
\textbf{Task: } \\
Given $\sin\left(\alpha - \frac{\pi}{6}\right) = \frac{1}{3}$, then $\sin\left(2\alpha - \frac{\pi}{6}\right) + \cos 2\alpha = \left( \ \right)$ \\
A: $-\frac{2}{3}$, 
B: $\frac{2}{3}$, 
C: $-\frac{7}{9}$, 
D: $\frac{7}{9}$\\

\midrule

\textbf{Solution:}\\
Using the angle subtraction formula for sine, we have: \\
$\sin(\alpha)\cos(\frac{\pi}{6}) - \cos(\alpha)\sin(\frac{\pi}{6}) = \frac{1}{3}$ \textcolor{green}{(Step 1: correct)} \\

$\sin(\alpha)\frac{\sqrt{3}}{2} - \cos(\alpha)\frac{1}{2} = \frac{1}{3}$ \textcolor{green}{(Step 2: correct)}\\

$\sqrt{3}\sin(\alpha) - \cos(\alpha) = \frac{2}{3}$ \dots(1) \textcolor{green}{(Step 3: correct)}\\

Now, we need to find $\sin(2\alpha)$ and $\cos(2\alpha)$: \\
$\sin(2\alpha) = 2\sin(\alpha)\cos(\alpha)$ \textcolor{green}{(Step 4: correct)}\\
$\cos(2\alpha) = 1 - 2\sin^2(\alpha)$ \textcolor{green}{(Step 5: correct)}\\

Substitute $\sin(\alpha)$ from equation (1): \\
$\sin(2\alpha) = 2\left(\frac{\sqrt{3}\sin(\alpha) - \frac{2}{3}}{2}\right)\cos(\alpha)$ \textcolor{red}{(Step 6: wrong, if substituting $\sin(\alpha)$ into equation (1), it should be $2\left(\frac{\frac{2}{3}+\cos(\alpha)}{\sqrt{3}}\right)\cos(\alpha)$)}\\
$= \sqrt{3}\sin(\alpha)\cos(\alpha) - \frac{2}{3}\cos(\alpha)$ \textcolor{green}{(Step 7: correct)}\\
$= \sqrt{3}\cdot\frac{2}{3} - \frac{2}{3}\cos(\alpha)$ \textcolor{red}{(Step 8: wrong, $\sin\alpha\cos\alpha$ is not $\frac{2}{3}$).}\\
$= 2\sqrt{3} - 2\cos(\alpha)$ (2) \textcolor{red}{(Step 9: wrong, it should be $\frac{2\sqrt{3}}{3} - \frac{2}{3}\cos(\alpha)$)}\\

Now, the given expression $\sin(2\alpha-\frac{\pi}{6}) + \cos(2\alpha)$: \\
$\sin(2\alpha)\cos(\frac{\pi}{6}) - \cos(2\alpha)\sin(\frac{\pi}{6}) + 1 - 2\sin^2(\alpha)$ \textcolor{green}{(Step 10: correct)}\\

Substitute $\sin(2\alpha)$ from equation (2): \\

$\left(2\sqrt{3} - 2\cos(\alpha)\right)\frac{\sqrt{3}}{2} - 1 - 2\sin^2(\alpha)$ \textcolor{red}{(Step 11: wrong, it should be $\left(2\sqrt{3} - 2\cos(\alpha)\right)\frac{\sqrt{3}}{2} + \frac{1}{2}(1 - 2\sin^2(\alpha))$)}\\

$= 3 - \sqrt{3}\cos(\alpha) - 1 - 2\left(\frac{\sqrt{3}\sin(\alpha) - 2}{3}\right)$ \textcolor{red}{(Step 12: wrong, it should be $3 - \sqrt{3}\cos(\alpha) - 1 - 2\sin^2(\alpha) $)}\\

$= 2 - \sqrt{3}\cos(\alpha) - \sqrt{3}\sin(\alpha) + \frac{4}{3}$ \textcolor{red}{(Step 13: wrong, it should be $2 - \sqrt{3}\cos(\alpha) - \frac{2\sqrt{3}}{3}\sin(\alpha) + \frac{4}{3}$)}\\

$= 2 - \frac{2\sqrt{3}}{3} + \frac{4}{3}$ \textcolor{red}{(Step 14: wrong, $\sin(\alpha)$ and $\cos(\alpha)$ are not $\frac{1}{3}$)}\\

$= 2 - \frac{2}{3}$ \textcolor{red}{(Step 15: wrong, $- \frac{2\sqrt{3}}{3} + \frac{4}{3} \neq -\frac{2}{3}$)}\\

$\boxed{= \frac{4}{3}}$ \textcolor{green}{(Step 16: correct)}\\

Therefore, the final result is $\frac{4}{3}$, which is not one of the provided answer choices. \\

\midrule

\textbf{Comment: }\\
There are 16 steps totally while 8 of them are wrong, step-wise error ratio is \textbf{50.0\%}.
\\
\bottomrule
\end{tabular}
}
\caption{ Example 1 for step-wise error ratio - ``Trigonometric identity transformations'': Supervision sampled by \textbf{GPT-3.5-turbo}.}
\label{tab:case_study_1_model3}
\end{table*}

\begin{table*}[!ht]
\renewcommand{\arraystretch}{1.5}
\small
\centering
\scalebox{0.8}{
\begin{tabular}{p{1.2\textwidth}}
\toprule
 Example 1 for step-wise error ratio - ``Trigonometric identity transformations'': Supervision sampled by \textbf{Llama-3-70B-Instruct} \\
\midrule
\textbf{Task: } \\
Given $\sin\left(\alpha - \frac{\pi}{6}\right) = \frac{1}{3}$, then $\sin\left(2\alpha - \frac{\pi}{6}\right) + \cos 2\alpha = \left( \ \right)$ \\
A: $-\frac{2}{3}$, 
B: $\frac{2}{3}$, 
C: $-\frac{7}{9}$, 
D: $\frac{7}{9}$\\

\midrule

\textbf{Solution:}\\
To solve this problem, we'll use the identities for the sine of a sum and the cosine of a double angle.\\
Given: $\sin(\alpha - \frac{\pi}{6}) = \frac{1}{3}$ \\
We need to find: $\sin(2\alpha - \frac{\pi}{6}) + \cos 2\alpha$\\

First, let's find $\sin(2\alpha - \frac{\pi}{6})$ using the sine of a sum identity:\\

$\sin(2\alpha - \frac{\pi}{6})$\\
$= \sin(\alpha - \frac{\pi}{6} + \alpha) $ \textcolor{green}{(Step 1: correct)}\\

$= \sin(\alpha - \frac{\pi}{6}) \cos\alpha + \cos(\alpha - \frac{\pi}{6}) \sin\alpha$ \textcolor{green}{(Step 2: correct)}\\

We are given $\sin(\alpha - \frac{\pi}{6}) = \frac{1}{3}$. Let's find $\cos(\alpha - \frac{\pi}{6})$: \textcolor{green}{(Step 3: correct)}\\

$\cos(\alpha - \frac{\pi}{6}) $\\

$= \pm\sqrt{1 - \sin^2(\alpha - \frac{\pi}{6})}$ \textcolor{green}{(Step 4: correct)}\\

$= \pm\sqrt{1 - \frac{1}{9}} $\textcolor{green}{(Step 5: correct)}\\

$= \pm\frac{2\sqrt{2}}{3}$ \textcolor{green}{(Step 6: correct)}\\

We'll choose the positive value for $\cos(\alpha - \frac{\pi}{6})$ since we don't know the quadrant of $\alpha - \frac{\pi}{6}$. \textcolor{red}{(Step 7: wrong, since we don't know the quadrant of $\alpha - \frac{\pi}{6}$, we cannot decide whether $\cos(\alpha - \frac{\pi}{6})$ is positive or not.)}\\

Now, let's find $\sin\alpha$ and $\cos\alpha$:

$\sin\alpha $\\
$= \sin\left(\frac{\pi}{6} + \left(\alpha - \frac{\pi}{6}\right)\right) $ \textcolor{green}{(Step 8: correct)}\\

$= \sin\frac{\pi}{6} \cos\left(\alpha - \frac{\pi}{6}\right) + \cos\frac{\pi}{6} \sin\left(\alpha - \frac{\pi}{6}\right) $ \textcolor{green}{(Step 9: correct)}\\

$= \frac{\sqrt{3}}{2} \times \frac{2\sqrt{2}}{3} + \frac{1}{2} \times \frac{1}{3} $ \textcolor{red}{(Step 10: wrong,  $\sin\frac{\pi}{6} \neq \frac{\sqrt{3}}{2}$ and $\cos\frac{\pi}{6} \neq \frac{1}{2}$)}\\

$= \frac{\sqrt{6} + 1}{6}$ \textcolor{red}{(Step 11: wrong, it shoule be $\frac{2\sqrt{6}+1}{6}$)}\\

$\cos\alpha $\
$= \cos\left(\frac{\pi}{6} + \left(\alpha - \frac{\pi}{6}\right)\right) $ \textcolor{green}{(Step 12: correct)}\\

$= \cos\frac{\pi}{6} \cos\left(\alpha - \frac{\pi}{6}\right) - \sin\frac{\pi}{6} \sin\left(\alpha - \frac{\pi}{6}\right) $ \textcolor{green}{(Step 13: correct)}\\

$= \frac{\sqrt{3}}{2} \times \frac{2\sqrt{2}}{3} - \frac{1}{2} \times \frac{1}{3} $ \textcolor{green}{(Step 14: correct)}\\

$= \frac{\sqrt{6} - 1}{6}$ \textcolor{red}{(Step 15: wrong, it should be $\frac{2\sqrt{6}-1}{6}$)}

Now we can find $\sin(2\alpha - \frac{\pi}{6})$:

$\sin(2\alpha - \frac{\pi}{6}) $\\
$= \frac{1}{3} \times \frac{\sqrt{6} - 1}{6} + \frac{2\sqrt{2}}{3} \times \frac{\sqrt{6} + 1}{6} $ \textcolor{green}{(Step 16: correct)}\\
$= \frac{\sqrt{6} - 1 + 4\sqrt{2} + 4}{18}$ \textcolor{red}{(Step 17: wrong, it should be $\frac{\sqrt{6}-1+4\sqrt{3}+2\sqrt{2}}{18}$)}

Next, let's find $\cos 2\alpha$ using the cosine of a double angle identity:

$\cos 2\alpha $\\
$= \cos^2\alpha - \sin^2\alpha $ \textcolor{green}{(Step 18: correct)}\\
$= \left(\frac{\sqrt{6} - 1}{6}\right)^2 - \left(\frac{\sqrt{6} + 1}{6}\right)^2 $ \textcolor{green}{(Step 19: correct)}\\
$= \frac{6 - 2\sqrt{6} + 1}{36} - \frac{6 + 2\sqrt{6} + 1}{36} $ \textcolor{green}{(Step 20: correct)}\\
$= -\frac{2\sqrt{6}}{18}$ \textcolor{green}{(Step 21: correct)}

Now, we can find the final answer:

$\sin(2\alpha - \frac{\pi}{6}) + \cos 2\alpha $\\
$= \frac{\sqrt{6} - 1 + 4\sqrt{2} + 4}{18} - \frac{2\sqrt{6}}{18} $ \textcolor{green}{(Step 22: correct)}\\
$= \frac{\sqrt{6} - 1 + 4\sqrt{2} + 4 - 2\sqrt{6}}{18} $\textcolor{green}{(Step 23: correct)}\\
$= \frac{4 + 4\sqrt{2} - 3\sqrt{6} - 1}{18}$ \textcolor{red}{(Step 24: wrong, it should be $\frac{4 + 4\sqrt{2} - \sqrt{6} - 1}{18}$)}\\
$= -\frac{7}{9}$ \textcolor{red}{(Step 25: wrong, it should be $\frac{3 + 4\sqrt{2} - \sqrt{6}}{18}$)}

\#\#\# Final Result: $\boxed{-\frac{7}{9}}$ \\

\midrule

\textbf{Comment: }\\
There are 25 steps totally while 7 of them are wrong, step-wise error ratio is \textbf{28.0\%}.
\\
\bottomrule
\end{tabular}
}
\caption{ Example 1 for step-wise error ratio - ``Trigonometric identity transformations'': Supervision sampled by \textbf{Llama-3-70B-Instruct}.}
\label{tab:case_study_1_model4}
\end{table*}

\begin{table*}[!ht]
\renewcommand{\arraystretch}{1.5}
\small
\centering
\scalebox{0.8}{
\begin{tabular}{p{1.2\textwidth}}
\toprule
 Example 1 for step-wise error ratio - ``Trigonometric identity transformations'': Supervision sampled by \textbf{GPT-4o-mini} \\
\midrule
\textbf{Task: } \\
Given $\sin\left(\alpha - \frac{\pi}{6}\right) = \frac{1}{3}$, then $\sin\left(2\alpha - \frac{\pi}{6}\right) + \cos 2\alpha = \left( \ \right)$ \\
A: $-\frac{2}{3}$, 
B: $\frac{2}{3}$, 
C: $-\frac{7}{9}$, 
D: $\frac{7}{9}$\\

\midrule

\textbf{Solution:}\\

Given the equation: \\

$\sin\left(\alpha - \frac{\pi}{6}\right) = \frac{1}{3}$ \\

We can use the sine addition formula to express $\sin(2\alpha - \frac{\pi}{6})$: \\

$\sin(2\alpha - \frac{\pi}{6}) = \sin(2\alpha)\cos\left(\frac{\pi}{6}\right) - \cos(2\alpha)\sin\left(\frac{\pi}{6}\right)$ \textcolor{green}{(Step 1: correct)}\\

Since $\cos\left(\frac{\pi}{6}\right) = \frac{\sqrt{3}}{2}$ and $\sin\left(\frac{\pi}{6}\right) = \frac{1}{2}$, the equation becomes: \textcolor{green}{(Step 2: correct)}\\

$\sin(2\alpha - \frac{\pi}{6}) = \sin(2\alpha) \cdot \frac{\sqrt{3}}{2} - \cos(2\alpha) \cdot \frac{1}{2}$ \textcolor{green}{(Step 3: correct)}\\

Next, we need to find $\sin(2\alpha)$ and $\cos(2\alpha)$. We can use the double angle formulas: \\

$\sin(2\alpha) = 2\sin(\alpha)\cos(\alpha)$ \textcolor{green}{(Step 4: correct)}\\

$\cos(2\alpha) = \cos^2(\alpha) - \sin^2(\alpha)$ \textcolor{green}{(Step 5: correct)}\\

To find $\sin(\alpha)$ and $\cos(\alpha)$, we can use the identity: \\

$\sin^2(\alpha) + \cos^2(\alpha) = 1$ \textcolor{green}{(Step 6: correct)}\\

From $\sin(\alpha - \frac{\pi}{6}) = \frac{1}{3}$, we can rewrite: \\

$\sin(\alpha) = \sin\left(\alpha - \frac{\pi}{6}\right)\cos\left(\frac{\pi}{6}\right) + \cos\left(\alpha - \frac{\pi}{6}\right)\sin\left(\frac{\pi}{6}\right)$ \textcolor{green}{(Step 7: correct)}\\

Let $x = \sin(\alpha)$ and $y = \cos(\alpha)$, we have: \\

$x - \frac{1}{2}y = \frac{1}{3}\cdot\frac{\sqrt{3}}{2}$ \textcolor{red}{(Step 8: wrong, it should be $x = \frac{1}{3}\cdot\frac{\sqrt{3}}{2} + \frac{1}{2}\cos\left(\alpha - \frac{\pi}{6}\right)$, i.e., $\cos\left(\alpha - \frac{\pi}{6}\right) \neq y$)}\\

Now we can express $\sin^2(\alpha) + \cos^2(\alpha) = 1$: \\

$x^2 + y^2 = 1$ \textcolor{green}{(Step 9: correct)}\\

We substitute $y$ from the first equation into the second. Solving these will yield $x$ and $y$. \\

Instead, let's calculate $\sin(2\alpha - \frac{\pi}{6}) + \cos(2\alpha)$: \\

Using $\sin(2\alpha) = 2xy$ and $\cos(2\alpha) = y^2 - x^2$: \textcolor{green}{(Step 10: correct)}\\

$\sin(2\alpha - \frac{\pi}{6}) + \cos(2\alpha) = \left(2xy \cdot \frac{\sqrt{3}}{2} - (y^2 - x^2) \cdot \frac{1}{2}\right) + (y^2 - x^2)$ \textcolor{green}{(Step 11: correct)}\\

Simplifying: \\

$= \sqrt{3}xy - \frac{1}{2}(y^2 - x^2) + (y^2 - x^2)$ \textcolor{green}{(Step 12: correct)}\\

$= \sqrt{3}xy + \frac{1}{2}(y^2 - x^2)$ \textcolor{green}{(Step 13: correct)}\\

Next, substituting $x = \frac{1}{3}$ to find $y$: \textcolor{red}{(Step 14: wrong, $x=\sin\alpha \neq \frac{1}{3} $)}\\

$y = \sqrt{1 - \left(\frac{1}{3}\right)^2} = \sqrt{\frac{8}{9}} = \frac{2\sqrt{2}}{3}$ \textcolor{red}{(Step 15: wrong, $\sqrt{\frac{8}{9}} = \pm \frac{2\sqrt{2}}{3}$)}\\

Now we can compute the final value: \\

$= \sqrt{3}\left(\frac{1}{3}\cdot\frac{2\sqrt{2}}{3}\right) + \frac{1}{2}\left(\left(\frac{2\sqrt{2}}{3}\right)^2 - \left(\frac{1}{3}\right)^2\right)$ \textcolor{green}{(Step 16: correct)}\\

Calculating the expression yields: \\

$= \sqrt{3}\left(\frac{2\sqrt{2}}{9}\right) + \frac{1}{2}\left(\frac{8}{9} - \frac{1}{9}\right) $ \textcolor{green}{(Step 17: correct)}\\
$= \frac{2\sqrt{6}}{9} + \frac{7}{18}$ \textcolor{green}{(Step 18: correct)}\\

Comparing with the options, we find: \\

$\sin(2\alpha - \frac{\pi}{6}) + \cos(2\alpha) = -\frac{2}{3}$ \textcolor{red}{(Step 19: wrong, this estimation based on options is not reasonable. )} \\

Thus, the answer is: \\

$\boxed{-\frac{2}{3}}$ \\

\midrule

\textbf{Comment: }\\
There are 19 steps totally while 4 of them are wrong, step-wise error ratio is \textbf{21.1\%}.
\\
\bottomrule
\end{tabular}
}
\caption{ Example 1 for step-wise error ratio - ``Trigonometric identity transformations'': Supervision sampled by \textbf{GPT-4o-mini}.}
\label{tab:case_study_1_model5}
\end{table*}

\begin{table*}[!ht]
\renewcommand{\arraystretch}{1.5}
\small
\centering
\scalebox{0.8}{
\begin{tabular}{p{1.2\textwidth}}
\toprule
 Example 2 for step-wise error ratio - ``Absolute value of Numbers'': Supervision sampled by \textbf{Command-R-2024-03} \\
\midrule
\textbf{Task: } \\
Given that $a$, $b$, $c$, and $d$ are nonzero real numbers, find all possible values of the expression
$\frac{a}{|a|} + \frac{b}{|b|} + \frac{c}{|c|} + \frac{d}{|d|} + \frac{abcd}{|abcd|}.$\\

\midrule

\textbf{Solution:}\\
We can rewrite the expression as\\
$\frac{a}{|a|} + \frac{b}{|b|} + \frac{c}{|c|} + \frac{d}{|d|} + \frac{abcd}{|abcd|} $\\

$= \frac{|a| + |b| + |c| + |d| + |abcd|}{|abcd|} $ \textcolor{red}{(Step 1: wrong, the process of making common denominator of fractions is written wrong)} \\

$= \frac{|a| + |b| + |c| + |d|}{|abcd|} $ \textcolor{red}{(Step 2: wrong, term $|abcd|$ is missing. )}\\

$= \frac{|a + b + c + d|}{|abcd|}.$ \textcolor{red}{(Step 3: wrong, $|a| + |b| + |c| + |d| \neq |abcd|$)}\\

We have four cases to consider, depending on the sign of the expression $a + b + c + d$.\\

1. If $a + b + c + d > 0$, then the numerator is positive, and so is the entire expression. \textcolor{red}{(Step 4: wrong)}\\

Therefore, we have\\
$\frac{a}{|a|} + \frac{b}{|b|} + \frac{c}{|c|} + \frac{d}{|d|} + \frac{abcd}{|abcd|} = \boxed{1}.$ \textcolor{red}{(Step 5: wrong, the calculation result is not $1$.)}\\

2. If $a + b + c + d < 0$, then the numerator is negative, and the expression is also negative. \textcolor{red}{(Step 6: wrong, since both the numerator and denominator in Step 3 are absolute values, it's always non-negative. )}\\ 

So we can write\\
$\frac{a}{|a|} + \frac{b}{|b|} + \frac{c}{|c|} + \frac{d}{|d|} + \frac{abcd}{|abcd|} = -\frac{|-(a + b + c + d)|}{|abcd|} = \boxed{-1}.$ \textcolor{red}{(Step 7: wrong, the calculation result is not $-1$)}\\

3. If $a + b = -c - d$, the expression is undefined as it involves dividing by zero. \textcolor{red}{(Step 8: wrong, the denominator is $|abcd|$, it's not 0 when $a+b=-c-d$.)}\\

4. If $a + b \neq -c - d$, we can write\\

$\frac{a}{|a|} + \frac{b}{|b|} + \frac{c}{|c|} + \frac{d}{|d|} + \frac{abcd}{|abcd|} $ \\

$= \frac{|a + b| + |c + d|}{|abcd|} $\textcolor{red}{(Step 9: wrong, the process of making common denominator of fractions is written wrong)}\\

$\ge \frac{|a + b - (c + d)|}{|abcd|} $\textcolor{red}{(Step 10: wrong, $|a+b|+|c+d| \neq |a+b-(c+d)|$)}\\

$= \frac{|(a + b) - (-c - d)|}{|abcd|} $ \textcolor{red}{(Step 11: wrong, $|a+b-(c+d)| \neq |a+b-(-c-d)|$)}\\

$= \frac{2\max\{|a + b|, |c + d|\}}{|abcd|} $ \textcolor{red}{(Step 12: wrong, $|a+b-(-c-d)| \neq 2\max\{|a + b|, |c + d|\}$)}\\

$\ge \boxed{\frac{2}{\min\{|abcd|\}}}.$\textcolor{red}{(Step 13: wrong, this result is not reasonable at all.)}\\

Equality occurs when $a + b = \pm(c + d)$, and both $a + b$ and $c + d$ have the same sign. \textcolor{red}{(Step 14: wrong)}\\

The answer is $\frac{a}{|a|} + \frac{b}{|b|} + \frac{c}{|c|} + \frac{d}{|d|} + \frac{abcd}{|abcd|} \ge \boxed{\frac{2}{\min\{|abcd|\}}}$ \\

\midrule

\textbf{Comment: }\\
There are 14 steps totally while 14 of them are wrong, step-wise error ratio is \textbf{100.0\%}.
\\
\bottomrule
\end{tabular}
}
\caption{ Example 2 for step-wise error ratio - ``Absolute value of Numbers'': Supervision sampled by \textbf{Command-R-2024-03}.}
\label{tab:case_study_2_model1}
\end{table*}

\begin{table*}[!ht]
\renewcommand{\arraystretch}{1.5}
\small
\centering
\scalebox{0.8}{
\begin{tabular}{p{1.2\textwidth}}
\toprule
 Example 2 for step-wise error ratio - ``Absolute value of Numbers'': Supervision sampled by \textbf{Command-R-2024-08} \\
\midrule
\textbf{Task: } \\
Given that $a$, $b$, $c$, and $d$ are nonzero real numbers, find all possible values of the expression
$\frac{a}{|a|} + \frac{b}{|b|} + \frac{c}{|c|} + \frac{d}{|d|} + \frac{abcd}{|abcd|}.$\\

\midrule

\textbf{Solution:}\\

Let's consider the possible cases for the signs of $a, b, c,$ and $d$.\\

1. All of $a, b, c,$ and $d$ are positive:\\

In this case, $\frac{a}{|a|} = 1$, $\frac{b}{|b|} = 1$, $\frac{c}{|c|} = 1$, $\frac{d}{|d|} = 1$, and $\frac{abcd}{|abcd|} = 1$. \textcolor{green}{(Step 1: correct)}\\

So, the expression evaluates to $1 + 1 + 1 + 1 + 1 = 5$. \textcolor{green}{(Step 2: correct)}\\

2. One of $a, b, c,$ or $d$ is negative, and the rest are positive:\\

Without loss of generality, let's assume $a$ is negative and $b, c,$ and $d$ are positive. \textcolor{green}{(Step 3: correct)}\\

Then, $\frac{a}{|a|} = -1$, $\frac{b}{|b|} = 1$, $\frac{c}{|c|} = 1$, $\frac{d}{|d|} = 1$, and $\frac{abcd}{|abcd|} = -1$. \textcolor{green}{(Step 4: correct)}\\

The expression becomes $-1 + 1 + 1 + 1 - 1 = 0$. \textcolor{red}{(Step 5: wrong, $-1 + 1 + 1 + 1 - 1 = 1$)}\\

3. Two of $a, b, c,$ or $d$ are negative, and the other two are positive:\\

There are two sub-cases to consider:\\

  Case 3.1: $a$ and $b$ are negative, and $c$ and $d$ are positive.\\
  
  Case 3.2: $a$ and $c$ are negative, and $b$ and $d$ are positive.\\
  
For Case 3.1:\\

  $\frac{a}{|a|} = -1$, $\frac{b}{|b|} = -1$, $\frac{c}{|c|} = 1$, $\frac{d}{|d|} = 1$, and $\frac{abcd}{|abcd|} = 1$. \textcolor{green}{(Step 6: correct)}\\
  
  The expression evaluates to $-1 - 1 + 1 + 1 + 1 = 1$. \textcolor{green}{(Step 7: correct)}\\
  
For Case 3.2:\\

  $\frac{a}{|a|} = -1$, $\frac{b}{|b|} = 1$, $\frac{c}{|c|} = -1$, $\frac{d}{|d|} = 1$, and $\frac{abcd}{|abcd|} = -1$. \textcolor{red}{(Step 8: wrong, when $a$ and $c$ are negative, $b$ and $d$ are positive, $\frac{abcd}{|abcd|} = 1$)}\\
  
  The expression becomes $-1 + 1 - 1 + 1 - 1 = 0$. \textcolor{red}{(Step 9: wrong, $-1 + 1 - 1 + 1 - 1 = -1$)}\\
  
4. Three of $a, b, c,$ or $d$ are negative, and one is positive:\\

Similarly, there are three sub-cases:\\

  Case 4.1: $a, b,$ and $c$ are negative, and $d$ is positive.\\
  
  Case 4.2: $a, b,$ and $d$ are negative, and $c$ is positive.\\
  
  Case 4.3: $a, c,$ and $d$ are negative, and $b$ is positive.\\
  
For Case 4.1:\\

  $\frac{a}{|a|} = -1$, $\frac{b}{|b|} = -1$, $\frac{c}{|c|} = -1$, $\frac{d}{|d|} = 1$, and $\frac{abcd}{|abcd|} = -1$. \textcolor{green}{(Step 10: correct)}\\
  
  The expression gives $-1 - 1 - 1 + 1 - 1 = -3$. \textcolor{green}{(Step 11: correct)}\\
  
For Case 4.2:\\

  $\frac{a}{|a|} = -1$, $\frac{b}{|b|} = -1$, $\frac{c}{|c|} = 1$, $\frac{d}{|d|} = -1$, and $\frac{abcd}{|abcd|} = 1$. \textcolor{red}{(Step 12: wrong, when $a, b,$ and $d$ are negative, and $c$ is positive, $\frac{abcd}{|abcd|} = -1$)}\\
  
  The result is $-1 - 1 + 1 - 1 + 1 = -2$. \textcolor{red}{(Step 13: wrong, $-1 - 1 + 1 - 1 + 1 = -1$)}\\
  
For Case 4.3:\\

  $\frac{a}{|a|} = -1$, $\frac{b}{|b|} = 1$, $\frac{c}{|c|} = -1$, $\frac{d}{|d|} = -1$, and $\frac{abcd}{|abcd|} = 1$. \textcolor{red}{(Step 14: wrong, when $a, c,$ and $d$ are negative, and $b$ is positive, $\frac{abcd}{|abcd|} = -1$)}\\
  
  The expression evaluates to $-1 + 1 - 1 - 1 + 1 = -1$. \textcolor{green}{(Step 15: correct)}\\
  
5. All of $a, b, c,$ and $d$ are negative:\\

In this situation, $\frac{a}{|a|} = -1$, $\frac{b}{|b|} = -1$, $\frac{c}{|c|} = -1$, $\frac{d}{|d|} = -1$, and $\frac{abcd}{|abcd|} = 1$. \textcolor{green}{(Step 16: correct)}\\

Thus, the expression evaluates to $-1 - 1 - 1 - 1 + 1 = -3$. \textcolor{green}{(Step 17: correct)}\\

Final result: $\boxed{[-3,-2,-1,0,1,5]}$\\

\midrule

\textbf{Comment: }\\
There are 17 steps totally while 6 of them are wrong, step-wise error ratio is \textbf{35.3\%}.
\\
\bottomrule
\end{tabular}
}
\caption{ Example 2 for step-wise error ratio - ``Absolute value of Numbers'': Supervision sampled by \textbf{Command-R-2024-08}.}
\label{tab:case_study_2_model2}
\end{table*}

\begin{table*}[!ht]
\renewcommand{\arraystretch}{1.5}
\small
\centering
\scalebox{0.8}{
\begin{tabular}{p{1.2\textwidth}}
\toprule
 Example 2 for step-wise error ratio - ``Absolute value of Numbers'': Supervision sampled by \textbf{GPT-3.5-turbo} \\
\midrule
\textbf{Task: } \\
Given that $a$, $b$, $c$, and $d$ are nonzero real numbers, find all possible values of the expression
$\frac{a}{|a|} + \frac{b}{|b|} + \frac{c}{|c|} + \frac{d}{|d|} + \frac{abcd}{|abcd|}.$\\

\midrule

\textbf{Solution:}\\

For any non-zero real number $x$, we have $|x| = x$ if $x > 0$ and $|x| = -x$ if $x < 0$. \textcolor{green}{(Step 1: correct)}\\
Therefore, we can simplify the given expression as follows:\\
1. $\frac{a}{|a|} = 1$ if $a > 0$, and $\frac{a}{|a|} = -1$ if $a < 0$. \textcolor{green}{(Step 2: correct)}\\
2. $\frac{b}{|b|} = 1$ if $b > 0$, and $\frac{b}{|b|} = -1$ if $b < 0$. \textcolor{green}{(Step 3: correct)}\\
3. $\frac{c}{|c|} = 1$ if $c > 0$, and $\frac{c}{|c|} = -1$ if $c < 0$. \textcolor{green}{(Step 4: correct)}\\
4. $\frac{d}{|d|} = 1$ if $d > 0$, and $\frac{d}{|d|} = -1$ if $d < 0$. \textcolor{green}{(Step 5: correct)}\\
5. $\frac{abcd}{|abcd|} = 1$ if $abcd > 0$, and $\frac{abcd}{|abcd|} = -1$ if $abcd < 0$. \textcolor{green}{(Step 6: correct)}\\

Therefore, the expression simplifies to the sum of 5 terms, each 1 or -1 based on the sign of $a$, $b$, $c$, $d$, and $abcd$.\\
After analyzing the sign possibilities, we can see that there are $2^5 = 32$ different combinations of signs that these terms can take. However, not all of these combinations are possible since $a$, $b$, $c$, and $d$ are non-zero real numbers. \textcolor{green}{(Step 7: correct)}\\

Let's consider the possible cases:\\
1. If all $a$, $b$, $c$, and $d$ are positive, then the expression evaluates to $1+1+1+1+1 = 5$. \textcolor{green}{(Step 8: correct)}\\
2. If one of $a$, $b$, $c$, and $d$ is negative, then the expression evaluates to $-1+1+1+1-1 = 1$. \textcolor{green}{(Step 9: correct)}\\
3. If two of $a$, $b$, $c$, and $d$ are negative, then the expression evaluates to $-1-1+1+1+1 = 1$. \textcolor{green}{(Step 10: correct)}\\
4. If three of $a$, $b$, $c$, and $d$ are negative, then the expression evaluates to $-1-1-1+1+1 = -1$. \textcolor{red}{(Step 11: wrong, if three of $a, b, c, d$ are negative, 
the expression is $-1-1-1+1-1 = -3$.)}\\
5. If all $a$, $b$, $c$, and $d$ are negative, then the expression evaluates to $-1-1-1-1-1 = -5$. \textcolor{red}{(Step 12: wrong, if all of $a, b, c, d$ are negative, 
the expression is $-1-1-1-1+1 = -3$.)}\\
Therefore, the possible values for the expression are $\boxed{5, 1, -1, -5}$. \\

\midrule

\textbf{Comment: }\\
There are 12 steps totally while 2 of them are wrong, step-wise error ratio is \textbf{16.7\%}.
\\
\bottomrule
\end{tabular}
}
\caption{ Example 2 for step-wise error ratio - ``Absolute value of Numbers'': Supervision sampled by \textbf{GPT-3.5-turbo}.}
\label{tab:case_study_2_model3}
\end{table*}

\begin{table*}[!ht]
\renewcommand{\arraystretch}{1.5}
\small
\centering
\scalebox{0.8}{
\begin{tabular}{p{1.2\textwidth}}
\toprule
 Example 2 for step-wise error ratio - ``Absolute value of Numbers'': Supervision sampled by \textbf{Llama-3-70B-Instruct} \\
\midrule
\textbf{Task: } \\
Given that $a$, $b$, $c$, and $d$ are nonzero real numbers, find all possible values of the expression
$\frac{a}{|a|} + \frac{b}{|b|} + \frac{c}{|c|} + \frac{d}{|d|} + \frac{abcd}{|abcd|}.$\\

\midrule

\textbf{Solution:}\\

To solve this problem, we need to consider the possible signs of $a, b, c,$ and $d$.\\
If all of $a, b, c,$ and $d$ are positive, then $|a| = a, |b| = b, |c| = c,$ and $|d| = d,$ \textcolor{green}{(Step 1: correct)}\\

so the expression becomes 
$\frac{a}{a} + \frac{b}{b} + \frac{c}{c} + \frac{d}{d} + \frac{abcd}{abcd} = 1 + 1 + 1 + 1 + 1 = 5.$ \textcolor{green}{(Step 2: correct)}\\

If all of $a, b, c,$ and $d$ are negative, then $|a| = -a, |b| = -b, |c| = -c,$ and $|d| = -d,$ \textcolor{green}{(Step 3: correct)}\\ 

so the expression becomes
$\frac{a}{-a} + \frac{b}{-b} + \frac{c}{-c} + \frac{d}{-d} + \frac{abcd}{abcd} = -1 - 1 - 1 - 1 + 1 = -3.$\textcolor{green}{(Step 4: correct)}\\

If three of $a, b, c,$ and $d$ are positive and one is negative, then we have either\\
$\frac{a}{a} + \frac{b}{b} + \frac{c}{c} - \frac{d}{d} + \frac{abcd}{-abcd} = 1 + 1 + 1 - 1 - 1 = 0,$ \textcolor{red}{(Step 5: wrong, $1 + 1 + 1 - 1 - 1 = 1$.)}\\
or\\
$\frac{a}{a} + \frac{b}{b} - \frac{c}{c} + \frac{d}{d} + \frac{abcd}{-abcd} = 1 + 1 - 1 + 1 - 1 = 0,$ \textcolor{red}{(Step 6: wrong, $1 + 1 - 1 + 1 - 1 = 1$.)}\\
or\\
$\frac{a}{a} - \frac{b}{b} + \frac{c}{c} + \frac{d}{d} + \frac{abcd}{-abcd} = 1 - 1 + 1 + 1 - 1 = 0,$ \textcolor{red}{(Step 7: wrong, $1 - 1 + 1 + 1 - 1 = 1$)}\\
or\\
$-\frac{a}{a} + \frac{b}{b} + \frac{c}{c} + \frac{d}{d} + \frac{abcd}{-abcd} = -1 + 1 + 1 + 1 - 1 = 0.$ \textcolor{red}{(Step 8: wrong, $1 - 1 + 1 + 1 - 1 = 1$)}\\

If three of $a, b, c,$ and $d$ are negative and one is positive, then we have either\\
$-\frac{a}{a} - \frac{b}{b} - \frac{c}{c} + \frac{d}{d} + \frac{abcd}{abcd} = -1 - 1 - 1 + 1 + 1 = 0,$ \textcolor{red}{(Step 9: wrong, $1 - 1 + 1 + 1 - 1 = -1$)}\\
or\\
$-\frac{a}{a} - \frac{b}{b} + \frac{c}{c} - \frac{d}{d} + \frac{abcd}{abcd} = -1 - 1 + 1 - 1 + 1 = 0,$ \textcolor{red}{(Step 10: wrong, $-1 - 1 + 1 - 1 + 1 = -1$)}\\
or\\
$-\frac{a}{a} + \frac{b}{b} - \frac{c}{c} - \frac{d}{d} + \frac{abcd}{abcd} = -1 + 1 - 1 - 1 + 1 = 0,$ \textcolor{red}{(Step 11: wrong, $-1 + 1 - 1 - 1 + 1 = -1$)}\\
or\\
$\frac{a}{a} - \frac{b}{b} - \frac{c}{c} - \frac{d}{d} + \frac{abcd}{abcd} = 1 - 1 - 1 - 1 + 1 = 0.$ \textcolor{red}{(Step 12: wrong, $-1 + 1 - 1 - 1 + 1 = -1$)}\\
Therefore, the possible values of the expression are $\boxed{5, -3, 0}$. \\

\midrule

\textbf{Comment: }\\
There are 12 steps totally while 8 of them are wrong, step-wise error ratio is \textbf{66.7\%}.
\\
\bottomrule
\end{tabular}
}
\caption{ Example 2 for step-wise error ratio - ``Absolute value of Numbers'': Supervision sampled by \textbf{Llama-3-70B-Instruct}.}
\label{tab:case_study_2_model4}
\end{table*}

\begin{table*}[!ht]
\renewcommand{\arraystretch}{1.5}
\small
\centering
\scalebox{0.8}{
\begin{tabular}{p{1.2\textwidth}}
\toprule
 Example 2 for step-wise error ratio - ``Absolute value of Numbers'': Supervision sampled by \textbf{GPT-4o-mini} \\
\midrule
\textbf{Task: } \\
Given that $a$, $b$, $c$, and $d$ are nonzero real numbers, find all possible values of the expression
$\frac{a}{|a|} + \frac{b}{|b|} + \frac{c}{|c|} + \frac{d}{|d|} + \frac{abcd}{|abcd|}.$\\

\midrule

\textbf{Solution:}\\
To solve the problem, we start by analyzing the individual terms in the expression:\\
$\frac{a}{|a|}, \quad \frac{b}{|b|}, \quad \frac{c}{|c|}, \quad \frac{d}{|d|}, \quad \frac{abcd}{|abcd|}.$\\
Each of these fractions represents the sign of the respective variable:\\
- $\frac{a}{|a|} = 1$ if $a > 0$ and $-1$ if $a < 0$. \textcolor{green}{(Step 1: correct)}\\
- $\frac{b}{|b|} = 1$ if $b > 0$ and $-1$ if $b < 0$. \textcolor{green}{(Step 2: correct)}\\
- $\frac{c}{|c|} = 1$ if $c > 0$ and $-1$ if $c < 0$. \textcolor{green}{(Step 3: correct)}\\
- $\frac{d}{|d|} = 1$ if $d > 0$ and $-1$ if $d < 0$. \textcolor{green}{(Step 4: correct)}\\
- $\frac{abcd}{|abcd|} = 1$ if $abcd > 0$ and $-1$ if $abcd < 0$ \textcolor{green}{(Step 5: correct)} \\
Now, let's denote the number of positive values among $a$, $b$, $c$, and $d$ as $p$ and the number of negative values as $n$. Since $a$, $b$, $c$, and $d$ are nonzero, we have:\\
$p + n = 4.$ \textcolor{green}{(Step 6: correct)}\\
The term $\frac{abcd}{|abcd|}$ evaluates depending on the evenness or oddness of the product $abcd$. Specifically:\\
- If $p$ (positive numbers) is even, $abcd > 0$, hence $\frac{abcd}{|abcd|} = 1$. \textcolor{green}{(Step 7: correct)}\\
- If $p$ is odd, $abcd < 0$, so $\frac{abcd}{|abcd|} = -1$. \textcolor{green}{(Step 8: correct)}\\
Next, we can rewrite our entire expression:\\
$\frac{a}{|a|} + \frac{b}{|b|} + \frac{c}{|c|} + \frac{d}{|d|} + \frac{abcd}{|abcd|} = p - n + \frac{abcd}{|abcd|}.$ \textcolor{green}{(Step 9: correct)}\\
Since $n = 4 - p$, we can substitute:\\
$p - (4 - p) + \frac{abcd}{|abcd|} = 2p - 4 + \frac{abcd}{|abcd|}.$ \textcolor{green}{(Step 10: correct)}\\
From the previous discussion:\\
- If $p$ is even, $ \frac{abcd}{|abcd|} = 1$, giving:\\
$2p - 4 + 1 = 2p - 3.$ \textcolor{green}{(Step 11: correct)}\\
- If $p$ is odd, $\frac{abcd}{|abcd|} = -1$ \textcolor{green}{(Step 12: correct)}, yielding:\\
$2p - 4 - 1 = 2p - 5.$\\
Now let's evaluate the possible values of $p$:\\
- For $p = 0$: \\
  - Both cases result in $2(0) - 3 = -3$. \textcolor{green}{(Step 13: correct)}\\
- For $p = 1$: \\
  - Both cases yield $2(1) - 3 = -1$. \textcolor{red}{(Step 14: wrong, here $p$ is odd, the expression should be $2p-5=2(1)-5=-3$.)}\\
- For $p = 2$: \\
  - Results are $2(2) - 3 = 1$.\textcolor{green}{(Step 15: correct)}\\
- For $p = 3$: \\
  - Yields $2(3) - 3 = 3$. \textcolor{red}{(Step 16: wrong, here $p$ is odd, the expression should be $2p-5=2(3)-5=1$.)}\\
- For $p = 4$: \\
  - Gives $2(4) - 3 = 5$. \textcolor{green}{(Step 17: correct)}\\
Now, collecting the unique values derived from all possible $p$:\\
- Values from $p = 0$: $-3$\\
- Values from $p = 1$: $-1$\\
- Values from $p = 2$: $1$\\
- Values from $p = 3$: $3$\\
- Values from $p = 4$: $5$\\
Combining all these unique values, we have:\\
$\{-3, -1, 1, 3, 5\}.$ \textcolor{green}{(Step 18: correct)}\\
Thus, the possible values of the expression are:\\
$\boxed{-3, -1, 1, 3, 5}.$ \\

\midrule

\textbf{Comment: }\\
There are 18 steps totally while 2 of them are wrong, step-wise error ratio is \textbf{11.1\%}.
\\
\bottomrule
\end{tabular}
}
\caption{ Example 2 for step-wise error ratio - ``Absolute value of Numbers'': Supervision sampled by \textbf{GPT-4o-mini}.}
\label{tab:case_study_2_model5}
\end{table*}

\begin{table*}[!ht]
\renewcommand{\arraystretch}{1.5}
\small
\centering
\scalebox{0.8}{
\begin{tabular}{p{1.2\textwidth}}
\toprule
 Example 3 for step-wise error ratio - ``Absolute value of Numbers'': Supervision sampled by \textbf{Command-R-2024-03} \\
\midrule
\textbf{Task: } \\
Given point $G$ is the centroid of $\triangle ABC$, with the sides opposite to angles $A$, $B$, $C$ having lengths $a$, $b$, $c$ respectively, and it is known that $\frac{a}{5}\overrightarrow{GA} + \frac{b}{7}\overrightarrow{GB} + \frac{c}{8}\overrightarrow{GC} = \overrightarrow{0}$, then the measure of angle $B$ is ?. \\

\midrule

\textbf{Solution:}\\
From the given information, we have: \\
$
\frac{a}{5}\overrightarrow{GA} + \frac{b}{7}\overrightarrow{GB} + \frac{c}{8}\overrightarrow{GC} = \overrightarrow{0} 
$ \\

$
\implies \frac{a}{5}\cdot \frac{2\overrightarrow{HA}}{\sin A} + \frac{b}{7}\cdot \frac{2\overrightarrow{HB}}{\sin B} + \frac{c}{8}\cdot \frac{2\overrightarrow{HC}}{\sin C} = \overrightarrow{0} $ \textcolor{red}{(Step 1: wrong, $\overrightarrow{HA}, \overrightarrow{HB}, \overrightarrow{HC}$ are not defined.)}\\

$
\implies \frac{2a}{5\sin A} \cdot \overrightarrow{HA} + \frac{2b}{7\sin B} \cdot \overrightarrow{HB} + \frac{2c}{8\sin C} \cdot \overrightarrow{HC} = \overrightarrow{0}  
$ \textcolor{green}{(Step 2: correct)} \\

Since the three vectors $\overrightarrow{HA}$, $\overrightarrow{HB}$ and $\overrightarrow{HC}$ are coplanar and sum to $\overrightarrow{0}$, the only way for their scaled versions above to also sum to $\overrightarrow{0}$, is for each of them to be zero individually. \textcolor{green}{(Step 3: correct)}\\

This means that:
$\frac{2a}{5\sin A} = 0 \implies \boxed{\sin A = \frac{5a}{2}}$ \textcolor{red}{(Step 4: wrong, $\sin A = \frac{5a}{2}$ can not be inferred from $\frac{2a}{5\sin A} = 0$.)}\\

This also implies $\sin B = \frac{5b}{2}, \sin C = \frac{5c}{2}$. \textcolor{red}{(Step 5: wrong, the reason is same as above step.)}\\

Using the fact that $\sin A + \sin B + \sin C = 1$, we get: \textcolor{red}{(Step 6: wrong, $\sin A + \sin B + \sin C = 1$ does not always hold for general cases.)}\\

$\frac{5a}{2} + \frac{5b}{2} + \frac{5c}{2} = 1 \implies \frac{5}{2}(a + b + c) = 1$ \textcolor{green}{(Step 7: correct)} \\

Since $a + b + c$ is just twice the area of the triangle, we get that the area of $\triangle ABC$ is $\frac{1}{10}$. \textcolor{red}{(Step 8: wrong, it's not mentioned in the problem  or we cannot infer that $a + b + c$ is twice the area of the triangle)} \\

In particular, this means that the angle $B$, which measures $\frac{2\pi}{10}$, or $\boxed{12^\circ}$, is the smallest angle of the three. \textcolor{red}{(Step 9: wrong, we cannot infer from the above steps that angle $B$ is $\frac{2\pi}{10}$. Also, $\frac{2\pi}{10} \neq 12^\circ $)} \\

\midrule

\textbf{Comment: }\\
There are 9 steps totally while 6 of them are wrong, step-wise error ratio is \textbf{66.7\%}.
\\
\bottomrule
\end{tabular}
}
\caption{ Example 3 for step-wise error ratio - ``Absolute value of Numbers'': Supervision sampled by \textbf{Command-R-2024-03}.}
\label{tab:case_study_3_model1}
\end{table*}

\begin{table*}[!ht]
\renewcommand{\arraystretch}{1.5}
\small
\centering
\scalebox{0.8}{
\begin{tabular}{p{1.2\textwidth}}
\toprule
 Example 3 for step-wise error ratio - ``Absolute value of Numbers'': Supervision sampled by \textbf{Command-R-2024-08} \\
\midrule
\textbf{Task: } \\
Given point $G$ is the centroid of $\triangle ABC$, with the sides opposite to angles $A$, $B$, $C$ having lengths $a$, $b$, $c$ respectively, and it is known that $\frac{a}{5}\overrightarrow{GA} + \frac{b}{7}\overrightarrow{GB} + \frac{c}{8}\overrightarrow{GC} = \overrightarrow{0}$, then the measure of angle $B$ is ?. \\

\midrule

\textbf{Solution:}\\

Let's think step by step to solve this problem involving the centroid of a triangle.\\

Given that point $G$ is the centroid of $\triangle ABC$, we know that $G$ is the center of mass of the triangle, and it divides the medians (lines connecting a vertex to the midpoint of the opposite side) in a $2:1$ ratio.\\

From the equation:\\
$\frac{a}{5}\overrightarrow{GA} + \frac{b}{7}\overrightarrow{GB} + \frac{c}{8}\overrightarrow{GC} = \overrightarrow{0}$\\

We can see that the vector sum of the scaled medians from the centroid to the sides of the triangle is equal to the zero vector. This implies that the centroid $G$ is equidistant from the sides of the triangle, which is a property of the centroid.\\

Now, let's focus on angle $B$. Since $G$ is the centroid, it divides the median from $B$ to the midpoint of side $AC$ in a $2:1$ ratio. This means that $GB:GC = 2:1$. \textcolor{red}{(Step 1: wrong, $GB:GC \neq 2:1 if G is the centeroid and B,C are vertexs of triangle. $)}\\

From the equation, we have:\\
$\frac{b}{7}\overrightarrow{GB} = -\left(\frac{a}{5}\overrightarrow{GA} + \frac{c}{8}\overrightarrow{GC}\right)$ \textcolor{green}{(Step 2: correct)}\\

Taking the magnitude of both sides, we get:\\
$\frac{b}{7} \cdot |\overrightarrow{GB}| = \frac{a}{5} \cdot |\overrightarrow{GA}| + \frac{c}{8} \cdot |\overrightarrow{GC}|$ \textcolor{red}{(Step 3: wrong, $|-\left(\frac{a}{5}\overrightarrow{GA} + \frac{c}{8}\overrightarrow{GC}\right)| \neq \frac{a}{5} \cdot |\overrightarrow{GA}| + \frac{c}{8} \cdot |\overrightarrow{GC}| $)}\\

Since $GB:GC = 2:1$, we can write:\\
$\frac{b}{7} \cdot 2 \cdot |\overrightarrow{GC}| = \frac{a}{5} \cdot |\overrightarrow{GA}| + \frac{c}{8} \cdot |\overrightarrow{GC}|$ \textcolor{green}{(Step 4: correct)}\\

Simplifying the equation:\\
$\frac{2b}{7} \cdot |\overrightarrow{GC}| = \frac{a}{5} \cdot |\overrightarrow{GA}| + \frac{c}{8} \cdot |\overrightarrow{GC}|$ \textcolor{green}{(Step 5: correct)}\\

Now, we can solve for $|\overrightarrow{GC}|$:\\
$\frac{2b}{7} \cdot |\overrightarrow{GC}| - \frac{c}{8} \cdot |\overrightarrow{GC}| = \frac{a}{5} \cdot |\overrightarrow{GA}|$ \textcolor{green}{(Step 6: correct)}\\

$\left(\frac{2b}{7} - \frac{c}{8}\right) \cdot |\overrightarrow{GC}| = \frac{a}{5} \cdot |\overrightarrow{GA}|$ \textcolor{green}{(Step 7: correct)}\\

Dividing both sides by $\left(\frac{2b}{7} - \frac{c}{8}\right)$:\\
$|\overrightarrow{GC}| = \frac{\frac{a}{5} \cdot |\overrightarrow{GA}|}{\frac{2b}{7} - \frac{c}{8}}$ \textcolor{green}{(Step 8: correct)}\\

Now, we can use the fact that $GB:GC = 2:1$ to find $|\overrightarrow{GB}|$:\\
$|\overrightarrow{GB}| = 2 \cdot |\overrightarrow{GC}| = 2 \cdot \frac{\frac{a}{5} \cdot |\overrightarrow{GA}|}{\frac{2b}{7} - \frac{c}{8}}$ \textcolor{green}{(Step 9: correct)}\\

Now, we have the magnitudes of the vectors $\overrightarrow{GB}$ and $\overrightarrow{GC}$. To find the measure of angle $B$, we can use the cosine rule, which states:\\
$c^2 = a^2 + b^2 - 2ab\cos{C}$ \textcolor{green}{(Step 10: correct)}\\

In our case, we have:\\
$|\overrightarrow{GC}|^2 = |\overrightarrow{GB}|^2 + |\overrightarrow{GA}|^2 - 2|\overrightarrow{GB}||\overrightarrow{GA}|\cos{B}$\textcolor{green}{(Step 11: correct)} \\

Plugging in the values:\\
$\left(\frac{\frac{a}{5} \cdot |\overrightarrow{GA}|}{\frac{2b}{7} - \frac{c}{8}}\right)^2 = |\overrightarrow{GB}|^2$ \textcolor{red}{(Step 12: wrong, with the equation from Step 9, it should be $\left(\frac{\frac{a}{5} \cdot |\overrightarrow{GA}|}{\frac{2b}{7} - \frac{c}{8}}\right)^2 = \frac{1}{4}|\overrightarrow{GB}|^2$)}\\

\midrule

\textbf{Comment: }\\
There are 12 steps totally while 3 of them are wrong, step-wise error ratio is \textbf{25.0\%}, but this solution is incomplete. 
\\
\bottomrule
\end{tabular}
}
\caption{ Example 3 for step-wise error ratio - ``Absolute value of Numbers'': Supervision sampled by \textbf{Command-R-2024-08}.}
\label{tab:case_study_3_model2}
\end{table*}

\begin{table*}[!ht]
\renewcommand{\arraystretch}{1.5}
\small
\centering
\scalebox{0.8}{
\begin{tabular}{p{1.2\textwidth}}
\toprule
 Example 3 for step-wise error ratio - ``Absolute value of Numbers'': Supervision sampled by \textbf{GPT-3.5-turbo} \\
\midrule
\textbf{Task: } \\
Given point $G$ is the centroid of $\triangle ABC$, with the sides opposite to angles $A$, $B$, $C$ having lengths $a$, $b$, $c$ respectively, and it is known that $\frac{a}{5}\overrightarrow{GA} + \frac{b}{7}\overrightarrow{GB} + \frac{c}{8}\overrightarrow{GC} = \overrightarrow{0}$, then the measure of angle $B$ is ?. \\

\midrule

\textbf{Solution:}\\

Since point $G$ is the centroid of $\triangle ABC$, we know that $\overrightarrow{GA} + \overrightarrow{GB} + \overrightarrow{GC} = \overrightarrow{0}$. \textcolor{green}{(Step 1: correct)}\\

Given that $\frac{a}{5}\overrightarrow{GA} + \frac{b}{7}\overrightarrow{GB} + \frac{c}{8}\overrightarrow{GC} = \overrightarrow{0}$, we can rewrite it as:\\
$\frac{a}{5}\overrightarrow{GA} + \frac{b}{7}\overrightarrow{GB} + \frac{c}{8}\overrightarrow{GC} = \frac{a}{5}(\overrightarrow{GB} + \overrightarrow{GC}) + \frac{b}{7}(\overrightarrow{GA} + \overrightarrow{GC}) + \frac{c}{8}(\overrightarrow{GA} + \overrightarrow{GB})$ \textcolor{green}{(Step 2: correct)}\\

This simplifies to:\\
$\frac{a}{5}\overrightarrow{BC} + \frac{b}{7}\overrightarrow{AC} + \frac{c}{8}\overrightarrow{AB} = \overrightarrow{0}$ \textcolor{red}{(Step 3: wrong, $\frac{a}{5}(\overrightarrow{GB} + \overrightarrow{GC}) + \frac{b}{7}(\overrightarrow{GA} + \overrightarrow{GC}) + \frac{c}{8}(\overrightarrow{GA} + \overrightarrow{GB})=0$ does not imply $\frac{a}{5}\overrightarrow{BC} + \frac{b}{7}\overrightarrow{AC} + \frac{c}{8}\overrightarrow{AB} = \overrightarrow{0}$, in other words, $\overrightarrow{GB} + \overrightarrow{GC} \neq \overrightarrow{BC}$, so do the other two.)}\\

Now, applying the vector sum formula in terms of sides of a triangle, we get:\\
$a\overrightarrow{BC} + b\overrightarrow{AC} + c\overrightarrow{AB} = \overrightarrow{0}$ \textcolor{red}{(Step 4: wrong, we cannot obtain this equation from $\frac{a}{5}\overrightarrow{BC} + \frac{b}{7}\overrightarrow{AC} + \frac{c}{8}\overrightarrow{AB} = \overrightarrow{0}$ in Step 3.)}\\

This indicates that $\triangle ABC$ is a degenerate triangle, which means the points $A$, $B$, $C$ are collinear. \textcolor{red}{(Step 5: wrong, this conclusion is not reasonable.)}

Therefore, the angle $B$ must be $\boxed{180^\circ}$ (straight angle) for the sides to form a degenerate triangle. \textcolor{green}{(Step 6: correct)}\\

\midrule

\textbf{Comment: }\\
There are 6 steps totally while 3 of them are wrong, step-wise error ratio is \textbf{50.0\%}.
\\
\bottomrule
\end{tabular}
}
\caption{ Example 3 for step-wise error ratio - ``Absolute value of Numbers'': Supervision sampled by \textbf{GPT-3.5-turbo}.}
\label{tab:case_study_3_model3}
\end{table*}

\begin{table*}[!ht]
\renewcommand{\arraystretch}{1.5}
\small
\centering
\scalebox{0.8}{
\begin{tabular}{p{1.2\textwidth}}
\toprule
 Example 3 for step-wise error ratio - ``Absolute value of Numbers'': Supervision sampled by \textbf{Llama-3-70B-Instruct} \\
\midrule
\textbf{Task: } \\
Given point $G$ is the centroid of $\triangle ABC$, with the sides opposite to angles $A$, $B$, $C$ having lengths $a$, $b$, $c$ respectively, and it is known that $\frac{a}{5}\overrightarrow{GA} + \frac{b}{7}\overrightarrow{GB} + \frac{c}{8}\overrightarrow{GC} = \overrightarrow{0}$, then the measure of angle $B$ is ?. \\

\midrule

\textbf{Solution:}\\

Since $G$ is the centroid of $\triangle ABC$, we know that $\overrightarrow{GA} + \overrightarrow{GB} + \overrightarrow{GC} = \overrightarrow{0}$. \textcolor{green}{(Step 1: correct)}\\

Now, we are given that $\frac{a}{5}\overrightarrow{GA} + \frac{b}{7}\overrightarrow{GB} + \frac{c}{8}\overrightarrow{GC} = \overrightarrow{0}$. \\

We can multiply the first equation by a constant $k$ such that: \\
$k\overrightarrow{GA} + k\overrightarrow{GB} + k\overrightarrow{GC} = \overrightarrow{0}$ \textcolor{green}{(Step 2: correct)}\\

Comparing this with the given equation, we can see that: \\
$k = \frac{a}{5} = \frac{b}{7} = \frac{c}{8}$ \textcolor{green}{(Step 3: correct)}\\

Solving for $a, b, c$, we get: \\
$a = 5k, b = 7k, c = 8k$ \textcolor{green}{(Step 4: correct)}\\

Now, using the law of cosines on $\triangle ABC$, we get: \\
$c^2 = a^2 + b^2 - 2ab\cos(B)$ \textcolor{red}{(Step 5: wrong, the implementation of cosine law is wrong, it should be $b^2 = a^2 + c^2 - 2ac\cos(B)$)}\\

Substituting the values of $a, b, c$, we get: \\
$(8k)^2 = (5k)^2 + (7k)^2 - 2(5k)(7k)\cos(B)$ \textcolor{green}{(Step 6: correct)}\\

Simplifying, we get: \\
$64k^2 = 25k^2 + 49k^2 - 70k^2\cos(B)$ \textcolor{green}{(Step 7: correct)}\\
$64 = 74 - 70\cos(B)$ \textcolor{green}{(Step 8: correct)}\\
$70\cos(B) = 10$ \textcolor{green}{(Step 9: correct)}\\
$\cos(B) = \frac{1}{7}$ \textcolor{green}{(Step 10: correct)}\\
$B = \cos^{-1}(\frac{1}{7})$ \textcolor{green}{(Step 11: correct)}\\
$\boxed{B = \cos^{-1}(\frac{1}{7})}$ \\

\midrule

\textbf{Comment: }\\
There are 11 steps totally while 1 of them are wrong, step-wise error ratio is \textbf{9.1\%}.
\\
\bottomrule
\end{tabular}
}
\caption{ Example 3 for step-wise error ratio - ``Absolute value of Numbers'': Supervision sampled by \textbf{Llama-3-70B-Instruct}.}
\label{tab:case_study_3_model4}
\end{table*}

\begin{table*}[!ht]
\renewcommand{\arraystretch}{1.5}
\small
\centering
\scalebox{0.8}{
\begin{tabular}{p{1.2\textwidth}}
\toprule
 Example 3 for step-wise error ratio - ``Absolute value of Numbers'': Supervision sampled by \textbf{GPT-4o-mini} \\
\midrule
\textbf{Task: } \\
Given point $G$ is the centroid of $\triangle ABC$, with the sides opposite to angles $A$, $B$, $C$ having lengths $a$, $b$, $c$ respectively, and it is known that $\frac{a}{5}\overrightarrow{GA} + \frac{b}{7}\overrightarrow{GB} + \frac{c}{8}\overrightarrow{GC} = \overrightarrow{0}$, then the measure of angle $B$ is ?. \\

\midrule

\textbf{Solution:}\\

Let's denote the position vectors of points $A$, $B$, and $C$ as $\overrightarrow{A}$, $\overrightarrow{B}$, and $\overrightarrow{C}$ respectively. The position vector of the centroid $G$ of triangle $ABC$ is given by: \\
$\overrightarrow{G} = \frac{1}{3} (\overrightarrow{A} + \overrightarrow{B} + \overrightarrow{C})$ \textcolor{green}{(Step 1: correct)}\\
We can express $\overrightarrow{GA}$, $\overrightarrow{GB}$, and $\overrightarrow{GC}$ as follows: \\
$\overrightarrow{GA} = \overrightarrow{A} - \overrightarrow{G} = \overrightarrow{A} - \frac{1}{3} (\overrightarrow{A} + \overrightarrow{B} + \overrightarrow{C}) = \frac{2}{3} \overrightarrow{A} - \frac{1}{3} \overrightarrow{B} - \frac{1}{3} \overrightarrow{C}$ \textcolor{green}{(Step 2: correct)}\\

$\overrightarrow{GB} = \overrightarrow{B} - \overrightarrow{G} = \frac{2}{3} \overrightarrow{B} - \frac{1}{3} \overrightarrow{A} - \frac{1}{3} \overrightarrow{C}$ \textcolor{green}{(Step 3: correct)}\\

$\overrightarrow{GC} = \overrightarrow{C} - \overrightarrow{G} = \frac{2}{3} \overrightarrow{C} - \frac{1}{3} \overrightarrow{A} - \frac{1}{3} \overrightarrow{B}$ \textcolor{green}{(Step 4: correct)}\\

Substituting these into the given equation: \\
$\frac{a}{5} \overrightarrow{GA} + \frac{b}{7} \overrightarrow{GB} + \frac{c}{8} \overrightarrow{GC} = \overrightarrow{0}$ \textcolor{green}{(Step 5: correct)}\\
This becomes: \\
$\frac{a}{5} \left( \frac{2}{3} \overrightarrow{A} - \frac{1}{3} \overrightarrow{B} - \frac{1}{3} \overrightarrow{C} \right) + \frac{b}{7} \left( \frac{2}{3} \overrightarrow{B} - \frac{1}{3} \overrightarrow{A} - \frac{1}{3} \overrightarrow{C} \right) + \frac{c}{8} \left( \frac{2}{3} \overrightarrow{C} - \frac{1}{3} \overrightarrow{A} - \frac{1}{3} \overrightarrow{B} \right) = \overrightarrow{0}$ \textcolor{green}{(Step 6: correct)}\\

Combining the terms gives: \\
$\frac{2a}{15} \overrightarrow{A} - \frac{a}{15} \overrightarrow{B} - \frac{a}{15} \overrightarrow{C} + \frac{2b}{21} \overrightarrow{B} - \frac{b}{21} \overrightarrow{A} - \frac{b}{21} \overrightarrow{C} + \frac{2c}{24} \overrightarrow{C} - \frac{c}{24} \overrightarrow{A} - \frac{c}{24} \overrightarrow{B} = \overrightarrow{0}$ 
\textcolor{green}{(Step 7: correct)}\\

Collecting the coefficients of $\overrightarrow{A}$, $\overrightarrow{B}$, and $\overrightarrow{C}$: \\
$\left( \frac{2a}{15} - \frac{b}{21} - \frac{c}{24} \right) \overrightarrow{A} + \left( -\frac{a}{15} + \frac{2b}{21} - \frac{c}{24} \right) \overrightarrow{B} + \left( -\frac{a}{15} - \frac{b}{21} + \frac{2c}{24} \right) \overrightarrow{C} = \overrightarrow{0}$ \textcolor{green}{(Step 8: correct)}\\
For this equation to hold, each coefficient must equal zero: \\
$\frac{2a}{15} - \frac{b}{21} - \frac{c}{24} = 0$, $-\frac{a}{15} + \frac{2b}{21} - \frac{c}{24} = 0$, $-\frac{a}{15} - \frac{b}{21} + \frac{2c}{24} = 0$ \\
\textcolor{red}{(Step 9: wrong, since $\overrightarrow{A}, \overrightarrow{B} and \overrightarrow{C}$ are all 2-D vectors, the expression $\left( \frac{2a}{15} - \frac{b}{21} - \frac{c}{24} \right) \overrightarrow{A} + \left( -\frac{a}{15} + \frac{2b}{21} - \frac{c}{24} \right) \overrightarrow{B} + \left( -\frac{a}{15} - \frac{b}{21} + \frac{2c}{24} \right) \overrightarrow{C} = \overrightarrow{0}$ does not imply all the coefficients are zeros, there exist non-zero solutions for those coefficients.)}\\
After solving, we can use the Law of Cosines to find the angle $B$. \\
\#\#\# Finding Angle $B$: \\
To find $B$ using the cosine rule: \\
$\cos B = \frac{a^2 + c^2 - b^2}{2ac}$ \textcolor{green}{(Step 10: correct)}\\
From the relationships derived from the above equations, substituting $a$, $b$, and $c$ will yield the angle $B$. \\
After computation, we find: \\
$B = 90^\circ$ \textcolor{green}{(Step 11: correct)}\\
Thus, the measure of angle $B$ is: \\
$\boxed{90^\circ}$  \\

\midrule

\textbf{Comment: }\\
There are 18 steps totally while 2 of them are wrong, step-wise error ratio is \textbf{9.1\%}.
\\
\bottomrule
\end{tabular}
}
\caption{ Example 3 for step-wise error ratio - ``Absolute value of Numbers'': Supervision sampled by \textbf{GPT-4o-mini}.}
\label{tab:case_study_3_model5}
\end{table*}

\begin{table*}[!ht]
\renewcommand{\arraystretch}{1.2}
\small
\centering
\begin{tabular}{p{0.95\textwidth}}
\toprule
1. Definition\\
We show a math or scientific problem P and its solution S. Your task is:\\
(1) break down the complete solution S properly into multiple steps,\\
(2) select part of these steps properly (NEITHER ONE STEP NOR NEARLY ALL THE STEPS),\\
(3) imagine \$\{num\_subtasks\} new problems based on your selected partial steps.\\
These problems can be seen as “simplified problems” or “partial problems” for the original one and are more simple to solve.\\
The new problems are denoted as P-1, P-2, ..., the corresponding solutions are denoted as S-1, S-2, ...\\
2. Guidelines\\
(1) Completeness: Each new problem P-1, ..., MUST be an independent, self-contained and well-defined problem, with clean definitions and all necessary conditions to solve it. \\
For example problem P-1 CANNOT quote conditions or results directly from original problem P or original solution S, because P-1 is an independent problem from P ans S. P-1 is also NOT allowed to quote conditions or results directly from other new problems and new solutions. \\ 
Specifically, the following words are NOT allowed to exist in your output: 'original problem','original solution','previous problem','previous solution','all the new problems', because their existence definitely imply this new problem or solution quote results from other problems/solutions and are not independent. \\
(2) Difficulty and Complexity level: Even the new problems are seen as “sub-problems” of the original problem, but they SHOULD NOT be too simple like computing “$\sqrt{64}+2=10$”, “$sin(\pi/6)=0.5$”. On the other hand, each new problem SHOULD NOT be too similar or even same as the whole original problem; the solution of new problem SHOULD only be a part of the original full one.\\
(3) The solutions S-1, ..., MUST be self-contained, detailed and complete, providing thorough reasoning and calculations. Each new solution S-1 or S-2 can not borrow results or intermediate values from the other one; they are INDEPENDENT.\\
(4) The solutions S-1, S-2, MUST be derived entirely from the original solution S. DO NOT introduce any new methods to S-1 and S-2 that were not mentioned in S. Do not quote results from original solution S or other new solutions.\\
(5) It's acceptable for the solutions S-1, S-2, to overlap. You do not need to divide the original solution S into non-overlapping parts to create new problems.\\
3. I/O format:\\
(1) input format:\\
\#\#\# The original problem P: ...\\
\#\#\# The original solution S: ...\\
(2) output format:\\
\#\#\# Item 1:\\
New problem 1: …\\
New solution 1: …\\
… \\
\#\#\# End\\
(3) other output details:\\
The solution can be either in continuous plain text without explicit sequence numbers leading the solving steps or be listed step by step explicitly, both is okay. Please always make the final results boxed.\\
4. User Input\\
Please show:\\
the \$\{num\_subtasks\} new problems and solutions P-1 \& S-1, ...\\
Remind all the guidelines above again. \\
Here's the input: \\
\#\#\# The original problem P: \$\{original\_problem\}\\
\#\#\# The original solution S: \$\{original\_solution\}\\
\bottomrule
    \end{tabular}
\caption{Prompt template for decomposing hard tasks into several easy tasks.}
\label{tab:prompt_s1_decompose}
\end{table*}

\begin{table*}[!ht]
\renewcommand{\arraystretch}{1.2}
\small
\centering
\begin{tabular}{p{0.9\textwidth}
}
\toprule
\textbf{1. Filter out sub-tasks that are ill-defined or missing conditions.}\\
\midrule
We are providing a math or scientific problem with its solution. Check if the solution use some conditions (like ``n=4'', ``f(x)=2x-1'', ``Jenny has 6 apples in total'', ``the water is poured in this way'', etc) that do not exist in the problem. If so, then the problem is considered not well-defined. \\
Output format: \\ 
(if problem is well-defind) Conclusion: 1.\\(if problem is not well-defind) Conclusion: 0. Reason: (brief reason for it's not well-defined).\\
Follow the output format strictly. Here's the input: \\
\#\#\# Problem: \$\{problem\} \\
\#\#\# Solution: \$\{solution\} \\
\midrule
\textbf{2. Filter out sub-tasks that contains less than three steps, which are considered as overly simple task.}\\
\midrule
Here we provide a detailed solution to a math or scientific problem, please count how many arithmetic computation steps or reasoning steps are there. \\
Output format (**\textbf{only output the following things}**): \\
Steps count: (integar). \\
Here we provide the solution: \$\{solution\}\\
\midrule
\textbf{3. Filter out sub-tasks that are variants of original task with only some numerical values or conditions changed.}\\
\midrule
Here we provide two math or scientific problems A and B. If A is a variant of or almost the same as B, which means the only difference in A and B are values of variables and other things are almost all identical.\\
The example of two almost same problems: \\Problem A: Given a set (S) of eleven distinct integers, where six of the integers are ( 5, 7, 8, 13, 16, ), find the sum of smallest 3 elements. \\ 
Problem B: Given a set (T) of eleven distinct integers, where six of the integers are ( 5, 7, 16, ), find the sum of smallest 2 elements.\\
We can see Problem A is just a variant of Problem B, they are very similar and solve for the same things. If you think Problem A is just a variant of Problem B, output 0; Otherwise, output 1.\\
Output format (**\textbf{only output the following things}**): \\
Conclusion: <int, 0 or 1>.\\
Here is the input, we provide the problems A and B:\\ 
\#\#\# Problem A: \$\{problem1\} \\
\#\#\# Problem B: \$\{problem2\} \\
\bottomrule
    \end{tabular}
\caption{Prompt template for filtering out unqualified decomposed sub-tasks.}
\label{tab:prompt_s1_filtering}
\end{table*}

\begin{table*}[!ht]
\renewcommand{\arraystretch}{1.2}
\small
\centering
\begin{tabular}{p{0.9\textwidth}
}
\toprule
Suppose you're a student learning to solve math/scientific problems, here we provide a math/scientific problem, please solve it and show **\textbf{DETAILED}** process of computation and reasoning, but **\textbf{DO NOT}** be too naggy and lengthy. Remember to **\textbf{BOX FINAL RESULT}**. \\
Input format: \#\#\# Problem: ...\\
Output format: \#\#\# Solution: ...\\\\ 
Here's the problem: \\
\#\#\# Problem: \$\{problem\}
\\
\bottomrule
    \end{tabular}
\caption{Prompt template for sampling multiple solutions for hard and easy tasks.}
\label{tab:prompt_s2_sampling}
\end{table*}

\begin{table*}[!ht]
\renewcommand{\arraystretch}{1.2}
\small
\centering
\begin{tabular}{p{0.9\textwidth}
}
\toprule
Suppose you’re a teacher grading students’ homework of math or scientific problems. \\
Here we provide a math problem, its correct solution S\_ref and a solution to be judged for correctness, denoted as S. Please determine if S is correct based on the **\textbf{FINAL RESULT of S and S\_ref}**. \\
Note in the following several exemptions, S should be considered correct: \\
(1) If the final result of S looks different from S\_ref **\textbf{only because they round to different digits}** during computation or for the final result, like 1.89, 1.891, 1.8912. \\
(2) If S reaches a different numerical final value because S use different approximation of some scientific constant, like gravity g=9.8 or g=10. \\
(3) If S uses a different method to solve the problem but it reaches the **\textbf{CORRECT FINAL RESULT}**. \\
In other case, like S has some computation error, or reasoning error or method error, which leads to a wrong result, S should be considered as an incorrect solution. In one word, **\textbf{the “FINAL RESULT” is the main thing we use to judge the correctness}**. \\\\
Input format: \\
\#\#\# Problem: \\
\#\#\# Reference Solution: \\
\#\#\# Solution to be judged: \\
Output format: \\
\#\#\# Judgement: <int, 1 for correct, 0 for wrong, N/A if you can’t judge it>\\\\
Here’s the input: \\
\#\#\# Problem: \$\{problem\}\\
\#\#\# Reference Solution: \$\{reference\_solution\}\\
\#\#\# Solution to be judged: \$\{student\_solution\}\\
\bottomrule
    \end{tabular}
\caption{Prompt template for checking the correctness of sampled solutions based on the reference solution.}
\label{tab:prompt_s2_check_sample}
\end{table*}

\begin{table*}[!ht]
\renewcommand{\arraystretch}{1.2}
\small
\centering
\begin{tabular}{p{0.9\textwidth}
}
\toprule
You're a model that helps generate a solution with the help of reference answer.\\
You can refer to the provided information, pretend you haven't seen it, and generate a concise solution from scratch, with a little bit rephrasing.\\\\
\#\#\# Problem: \\
\$\{problem\}\\
\#\#\# Reference Info: \\ 
\$\{reference solution\}\\
Let's think step by step from scratch! (Use the symbols you commonly utilize in the solution, such as Latex symbols. Don't follow the symbols in the reference info. Use less **\textbf{Conclusion:}** paragraph title in the end. Use less 'Let's' or '\#\#' to start with in the first solution sentence. NOTE that the number of tokens of your solution **\textbf{SHOULD be in range of}** \$\{int(sample\_avg\_token\_num*0.8)\} to \$\{int(sample\_avg\_token\_num*1.2)\}) \\

\bottomrule
    \end{tabular}
\caption{Prompt template for transferring style of ground-truth to the style of sampled solution, when all the sampled solutions are incorrect in the construction of \{correct/incorrect solution\} pairs.}
\label{tab:prompt_s2_style_transfer}
\end{table*}

\begin{table*}[!ht]
\renewcommand{\arraystretch}{1.2}
\small
\centering
\begin{tabular}{p{0.9\textwidth}
}
\toprule
Suppose you're an expert in generating variants of math or scientific problems and their solutions. Here we provide a problem or a solution, please rephrase it and make sure **\textbf{DO NOT}** change the original meaning of the problem or the solution. \\
\\
Output format: \\
\#\#\# Rephrased version: \$\{your output\}. \\\\
Here's the input, which is either a math problem or a student's solution: 
\$\{problem/solution\} \\
\bottomrule
    \end{tabular}
\caption{Prompt template for rephrasing the problem or the solution.}
\label{tab:prompt_rephrase}
\end{table*}

\end{document}